\documentclass[10pt,twocolumn,letterpaper]{article}

\usepackage{cvpr}              
\usepackage{booktabs}
\usepackage{makecell}
\usepackage{colortbl}
\usepackage{xcolor}
\usepackage[most]{tcolorbox}

\definecolor{ex1figure_red}{RGB}{200,29,49}
\definecolor{ex1figure_blue}{RGB}{46,84,161}

\definecolor{ex2figure_red}{RGB}{203,41,60}

\definecolor{ex2figure_blue}{RGB}{47,85,151}

\definecolor{cvprblue}{rgb}{0.21,0.49,0.74}
\usepackage[pagebackref,breaklinks,colorlinks,allcolors=cvprblue]{hyperref}

\newenvironment{FontExtractPrompt}[1][]{
\begin{tcolorbox}[
    title={Textual Style Description Collection Prompt},
    fonttitle=\bfseries\footnotesize,
    fontupper=\scriptsize,
    colback=gray!10,
    colframe=black!50,
    coltitle=black,
    sharp corners,
    boxsep=3pt,
    left=4pt,
    right=4pt,
    top=4pt,
    bottom=4pt,
    enhanced,
    #1
]
}{\end{tcolorbox}}

\def\paperID{} 
\def\confName{CVPR}
\def\confYear{2026}

\title{Beyond Patches: Global-aware Autoregressive Model for \\Multimodal Few-Shot Font Generation}

\author{
Haonan Cai$^{1,2}$\footnotemark[1] \quad
Yuxuan Luo$^{1}$\footnotemark[1] \quad
Zhouhui Lian$^{1}$\footnotemark[2] \\
$^{1}$Wangxuan Institute of Computer Technology, Peking University \\
$^{2}$School of Electronics Engineering and Computer Science, Peking University \\
}

\begin{document}
\maketitle

\footnotetext[1]{*: Equal contribution.}
\footnotetext[2]{\textdagger: Corresponding author: lianzhouhui@pku.edu.cn}

\begin{abstract}
Manual font design is an intricate process that transforms a stylistic visual concept into a coherent glyph set. This challenge persists in automated Few-shot Font Generation (FFG), where models struggle to preserve both structural integrity and stylistic fidelity from limited references. While autoregressive (AR) models have demonstrated impressive generative capabilities, their application to FFG is constrained by conventional patch-level tokenization, which neglects global dependencies crucial for coherent font synthesis. Moreover, existing FFG methods remain within the image-to-image paradigm, relying solely on visual references and overlooking the role of language in conveying stylistic intent during font design. To address these limitations, we propose GAR-Font, a novel AR framework for multimodal few-shot font generation. GAR-Font introduces a global-aware tokenizer that effectively captures both local structures and global stylistic patterns, a multimodal style encoder offering flexible style control through a lightweight language-style adapter without requiring intensive multimodal pretraining, and a post-refinement pipeline that further enhances structural fidelity and style coherence. Extensive experiments show that GAR-Font outperforms existing FFG methods, excelling in maintaining global style faithfulness and achieving higher-quality results with textual stylistic guidance. Project Page: \url{https://xtryer-s.github.io/projects_pages/GAR_Font}
\end{abstract}    
\begin{figure}
    \centering
    \includegraphics[width=\linewidth]{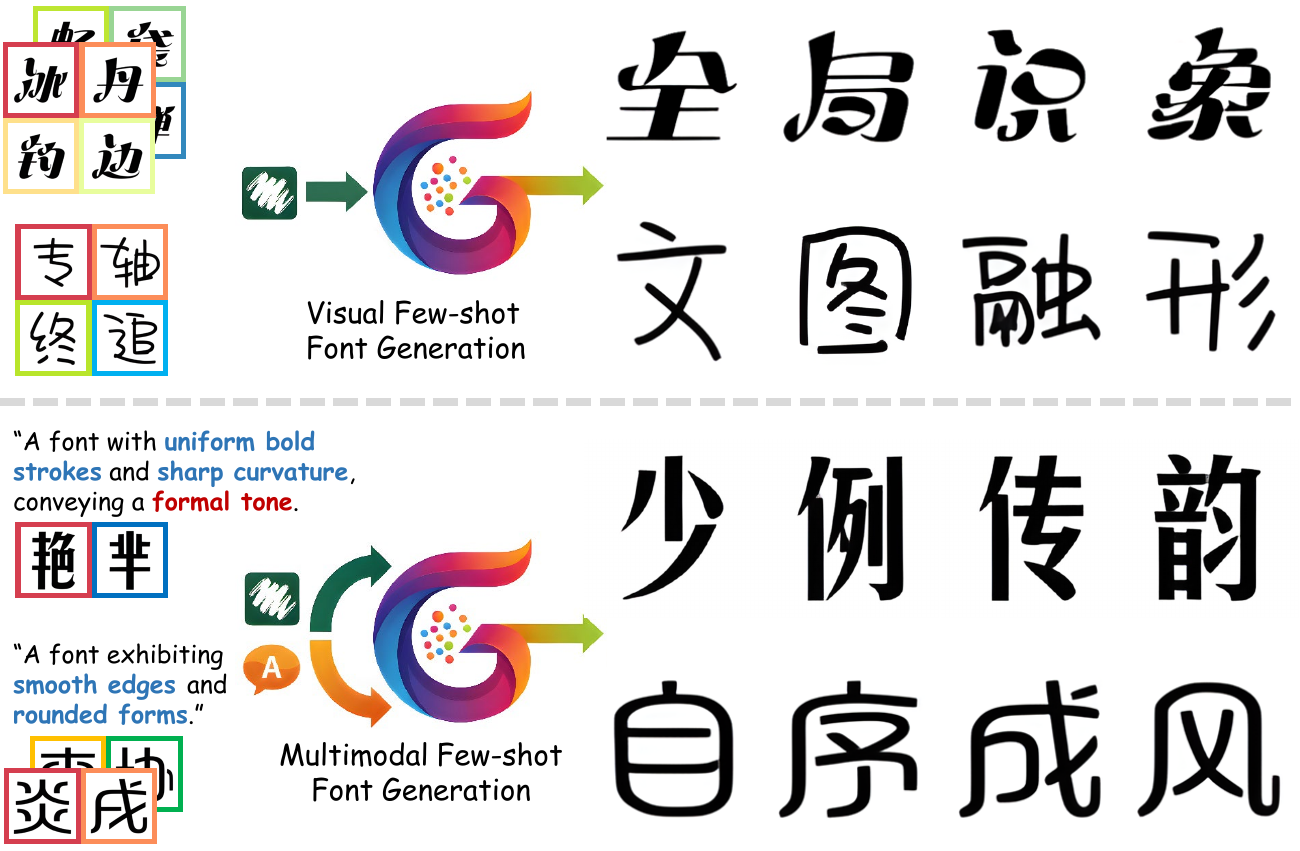}
    \caption{GAR-Font results under visual and multimodal few-shot settings. The generated poem reflects the key contributions of our model: global-aware tokenization for style fidelity, multimodal style encoding for text-image control, reduced reference requirements, and an autoregressive design that enables controllable high-quality font synthesis.}
    \label{fig: Teaser}
\end{figure}
\section{Introduction}

High-quality fonts are central to visual communication, yet their manual creation is costly: practitioners must translate their design concepts into a full, consistent glyph set. This challenge is compounded for logographic systems like Chinese and Japanese, which contain tens of thousands of characters and complex stroke geometries.

Few-shot Font Generation (FFG) seeks to automate this process by generating an entire font library from only a handful of reference examples. However, it presents several technical difficulties. First, it demands precise structural fidelity, where strokes, radicals, and local geometry must be correct for every glyph. Second, it requires capturing a holistic and faithful style representation, generalizing from a handful of examples across diverse characters.

A core obstacle for FFG is the lack of an effective visual global representation. GAN-based methods~\cite{LF_Font, CG_GAN} often exhibit noticeable style discrepancies from the reference fonts, and struggle with stroke-level accuracy. Diffusion models~\cite{Diff_Font, Font_Diffuser, HFH_Font} improve structural correctness and local fidelity but do not guarantee a coherent global style. Existing sequence approaches such as VQ-Font~\cite{VQ_Font} and IF-Font~\cite{IF_Font} rely on local patch or block tokenization, which fragments global cues and leads to noticeable discrepancies from the reference styles. Recent autoregressive (AR) image generation methods~\cite{Titok, SpectralAR} further reveal the superiority of globally contextualized 1D tokens over 2D patched tokenization, highlighting a promising direction for modeling global stylistic patterns in font synthesis. 

Meanwhile, existing FFG methods remain single-modal, using only visual modality for style control. In contrast, linguistic descriptions convey global conceptual design intents beyond visual appearance, providing complementary representations that help overcome the limitations of vision-only models when style must be inferred from a few examples.

These insights motivate GAR-Font, an expressive and controllable FFG framework that integrates visual and linguistic information. Leveraging the AR modeling capacity, GAR-Font introduces three major technical contributions:

\begin{itemize}
\item We propose \textbf{a global-aware tokenizer (G-Tok)} that fuses local features with global perception, capturing both fine-grained stroke details and font-level style patterns.
\item We design \textbf{an AR generator with multimodal style encoder}, first pretrained on visual inputs, and then augmented with a lightweight language-style adapter.
\item We introduce \textbf{a post-refinement pipeline} that improves structural fidelity and stylistic coherence, producing high-quality glyphs from limited references.
\end{itemize}

To validate the effectiveness of GAR-Font, we conduct extensive experiments across multiple settings. In the standard vision-only scenario, it surpasses other competitive FFG methods, demonstrating superior structural and stylistic fidelity. In the multimodal setting, the augmented multimodal style encoder further improves generation quality. Using 4 reference images plus one textual description, GAR-Font matches the quantitative performance of 8 images while providing flexible control. Ablation studies further confirm that G-Tok produces stable and coherent representations that preserve holistic structure and reference style fidelity—key factors for high-quality font synthesis. GAR-Font moves beyond patch-level modeling by learning global-aware representations and enabling text-driven multimodal FFG, establishing a new AR-based solution for automatic and controllable font generation.
\section{Related Work}
\subsection{Autoregressive Image Generation}
Autoregressive (AR) models have shown powerful generation ability across modalities~\cite{Flamingo, Videollava, llavamed, Unifiedio2}, and their extension to visual generation~\cite{ImageTransformer, llamagen, Chameleon, Emu, Emu3, januspro, simpleAR} has demonstrated promising potential. Typically, they follow a two-stage paradigm: a tokenizer encodes images into discrete representations (e.g., VQ-VAEs~\cite{VQ_VAE, VQ_VAE2}, RQ-VAE~\cite{RQ_VAE_RQ_FORMER}, VQ-GAN~\cite{VQ_GAN, VIT_VQGAN}), and a transformer decoder~\cite{MaskGIT, star_scale_wise, VAR, RAR, SpectralAR} that sequentially predicts these tokens for image reconstruction.

Most AR models use 2D patch-wise tokenization, capturing local details but lacking holistic awareness~\cite{MVTQ, EfficientVQGAN, ARPG_LocalTokens, ImageGPT}. To improve global coherence, recent works explore complementary strategies: global-context modeling~\cite{Hita} introduces holistic queries for structural reasoning; frequency-domain methods~\cite{SpectralAR, NFIG} encode coarse-to-fine spectral context; 1D tokenizers~\cite{Titok, FlexTok, InstellaT2I} enhance efficiency but lose adaptability; semantic tokenization~\cite{ImageFolder, TexTok} leverages language priors for meaningful visual codes. Structured visual like glyphs, however, demands tokenizers that unify local stroke detail with global aesthetics. Our GAR-Font addresses this challenge through a \textbf{global-aware tokenizer} that integrates CNN-based locality with Transformer-based global reasoning, effectively unifying fine-grained stroke detail and overall stylistic coherence.

\subsection{Few-shot Font Generation}
Few-shot Font Generation (FFG) can be broadly divided into vector-based and image-based approaches. Vector FFG methods have progressed in sequential modeling and efficient representations~\cite{FontRNN, Deepvecfont, DeepVecFontV2, DualVector, Vecfusion, Vecfontsdf}. Yet complex ideographic generation and unseen glyph generalization remain challenging~\cite{EasyFont, CVFont}. Image-based methods, in contrast, learn 2D pixel priors for robust adaptation, evolving from early image-to-image translation~\cite{zi2zi, word2word, zigan, HGAN} to more recent VQ and diffusion formulations~\cite{Font_Diffuser, HFH_Font, FSTDIFF, VQ_Font}. These models typically employ content–style disentanglement to separately capture structure and style~\cite{zhang2018separating, EMD, AGIS-Net, DG_Font, MX_Font, DS-Font, FineGrainedFFG, fewshotFFG, Xmpfont, Diff_Font}, enhanced by mechanisms like contrastive learning for global coherence~\cite{SCFont} and cross-attention for local transfer~\cite{Fs_Font}.  

The challenge is most acute for glyph-rich, ideographic languages such as Chinese and Korean, where thousands of characters demand fine structure and faithful global style~\cite{LegacyLearning, SCFont, GenerateExpert, HFH_Font}. In this regime, \textbf{image-based AR models that operate FFG on globally contextualized visual tokens} show particular promise: by modeling spatial dependencies directly in the 2D domain, they better capture the intricate geometry and visual regularities of glyphs.

\begin{figure}[t!]
    \centering
    \includegraphics[width=\linewidth]{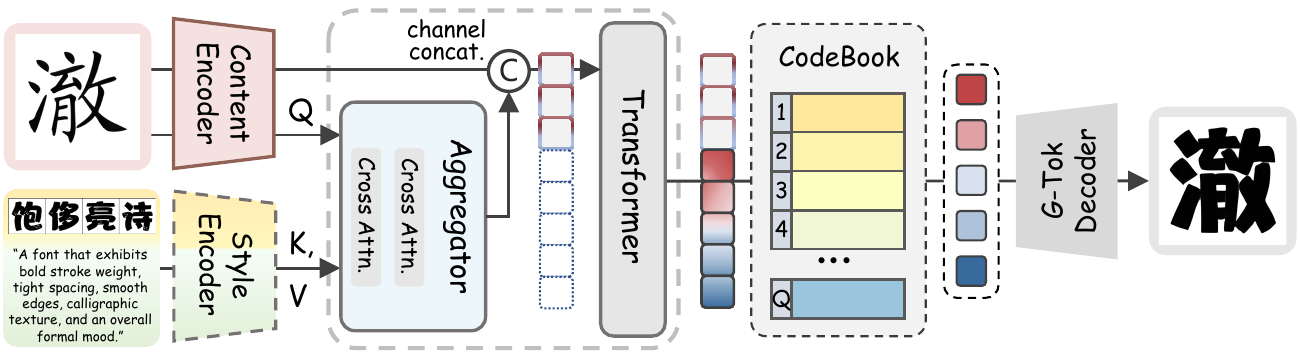}
    \caption{The overall architecture of GAR-Font. It comprises a global-aware tokenizer (G-Tok), and an AR generator, equipped with a multimodal style encoder.}
    \label{fig: Model}
\end{figure}

\begin{figure*}[!t]
\centering
\includegraphics[width=0.9\linewidth]{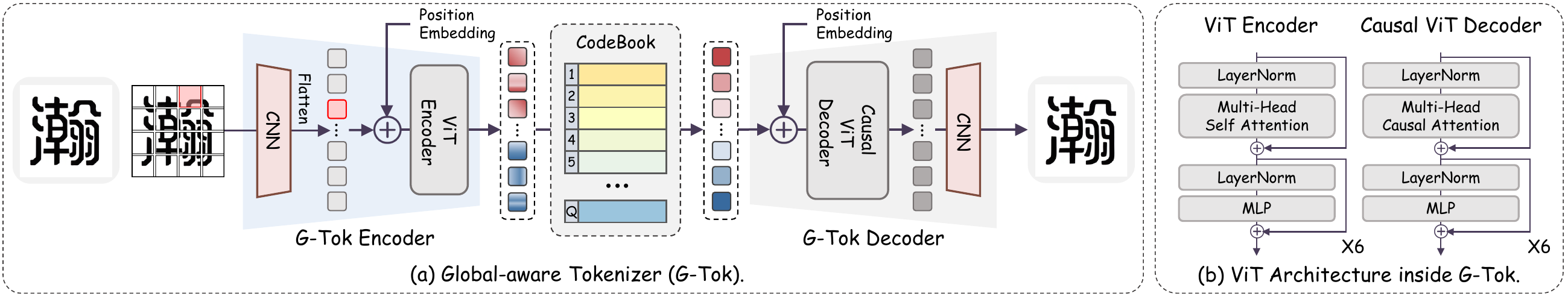}
\caption{(a) Overview of the G-Tok architecture, which adopts a hybrid CNN–ViT design. (b) Details of the global ViT encoder and causal ViT decoder. With self- and causal-attention, G-Tok captures global dependencies, enabling coherent and high-quality font synthesis.}
\label{fig: G-tok}
\end{figure*}

\subsection{Multimodal Alignment for Style Control}
Multimodal alignment for style control unifies visual and textual modalities for coherent, fine-grained style manipulation. Foundational models~\cite{Emu3, januspro, BAGEL, Seed-X, Qwen-Image, Hunyuan-Image} achieve this through costly multimodal pretraining. For fonts, however, only textual inputs describing style are incorporated, and the independent encoding of content and style in FFG frameworks~\cite{VQ_Font, Fs_Font, LF_Font} naturally forms a structured visual style space that serves as a strong prior for multimodal integration. Recent works pursue efficient alignment via parameter-efficient modules~\cite{FonTS, FontAdapter, LoRA, AdaCLIP} or resampler-based strategies aligning vision and language spaces~\cite{Flamingo, PaLM2-VAdapter, CalliReader, TextAdapter}. Motivated by these works, GAR-Font extends the style encoder in FFG models to a \textbf{multimodal style encoder} that consists of a visual style encoder and a lightweight pluggable \textbf{language-style adapter} that aligns textual descriptions with visual style embeddings, enabling fine-grained, flexible style control without the need for large-scale multimodal pretraining.

\section{Method}
The architecture of GAR-Font is illustrated in Figure~\ref{fig: Model}. Our framework comprises three core components:

\noindent\textbf{A global-aware tokenizer (G-Tok)} discretizes glyphs into tokens, capturing intricate structure and global visual style.

\noindent\textbf{An AR generator with multimodal style encoder} first learns from visual inputs to establish a stable style space, and then employs a lightweight language-style adapter for multimodal style control.

\noindent\textbf{A post-refinement} enhances structural fidelity and style coherence for high-quality few-shot font generation.

Given a content glyph, a few style references, and optional text, GAR-Font encodes and aggregates content and style, autoregressively generates codebook representations and decodes them softly into high-quality glyphs, which are subsequently improved by a post-refining stage. The details of each module adopted in the proposed GAR-Font will be presented in the following subsections.

\subsection{Global-aware Tokenizer}
The Global-aware Tokenizer (G-Tok), shown in Figure~\ref{fig: G-tok}, encodes glyphs into tokens using a hybrid CNN–ViT encoder, a vector quantizer, and a causal hybrid decoder. By fusing local convolutional features with Transformer-based global context, it captures fine-grained structure and overall style, enabling coherent, high-quality font synthesis.

\paragraph{Hybrid encoding.} Given a glyph image $\mathbf{I} \in \mathbb{R}^{H \times W \times 3}$, a CNN encoder $E_{\text{CNN}}$ extracts local stroke features, which are flattened and aggregated through a ViT encoder $E_{\text{ViT}}$:
\begin{equation}
    \mathbf{T} = E_{\text{ViT}}\big(\mathrm{Proj}(E_{\text{CNN}}(\mathbf{I})) + \mathbf{P}_{\text{2D}}\big) \in \mathbb{R}^{N\times d},
\end{equation}
where $\mathbf{P}_{\text{2D}}$ refers to the 2D sinusoidal position embeddings~\cite{ViT}, $N$ and $d$ denote token count and embedding dimension. The CNN backbone preserves spatial locality, capturing fine stroke geometry, while the ViT aggregates tokens globally for style fidelity. This hybrid design captures both structure fidelity and long-range style dependencies.

\paragraph{Vector Quantization.} Following standard VQ-VAE~\cite{VQ_VAE} and VQ-GAN~\cite{VQ_GAN}, we discretize latent tokens by mapping them to the nearest entries in a learnable table. We apply the commitment and embedding regularization with an entropy term to stabilize training and preserve diversity.

\paragraph{Causal decoding and optimization.} A causal ViT–CNN decoder reconstructs glyph $\hat{\mathbf{I}}$ from quantized tokens, modeling sequential dependencies while convolutional layers refine local details. Figure~\ref{fig: G-tok}(b) illustrates this structure.

The tokenizer is optimized end-to-end with weighted reconstruction, perceptual, and vector quantization losses:
\begin{equation}
    \mathcal{L}_{\text{tok}} =
\lambda_{\text{rec}}\mathcal{L}_{\text{rec}} +
\lambda_{\text{per}}\mathcal{L}_{\text{per}} +
\lambda_{\text{vq}}\mathcal{L}_{\text{vq}},
\end{equation}
where $\mathcal{L}_{\text{rec}}=\|\mathbf{I}-\hat{\mathbf{I}}\|_1$ denotes the L1 reconstruction loss, $\mathcal{L}_{\text{per}}=\|\Phi(\mathbf{I})-\Phi(\hat{\mathbf{I}})\|_2^2$ represents the perceptual loss, and $\mathcal{L}_{\text{vq}}$ is the standard vector quantization loss. 

\subsection{AR Generator with Multimodal Style Encoder}
Using G-Tok’s global representations, the AR generator performs conditional sequential prediction. To leverage prior FFG insights (Content-style Aggregator) while avoiding the cost of joint image–text training, GAR-Font adopts a decoupled learning paradigm. The generator, aggregator, and style encoder are first trained on visual inputs to learn a stable representation. Based on this foundation, the visual style encoder is extended into a multimodal one through a lightweight, plug-in language adapter that aligns textual cues with learned visual styles, enabling flexible text-guided generation without disturbing the visual prior.

\subsubsection{Pretraining on Visual Modality}
\begin{figure}[t!]
    \centering
    \includegraphics[width=0.96\linewidth]{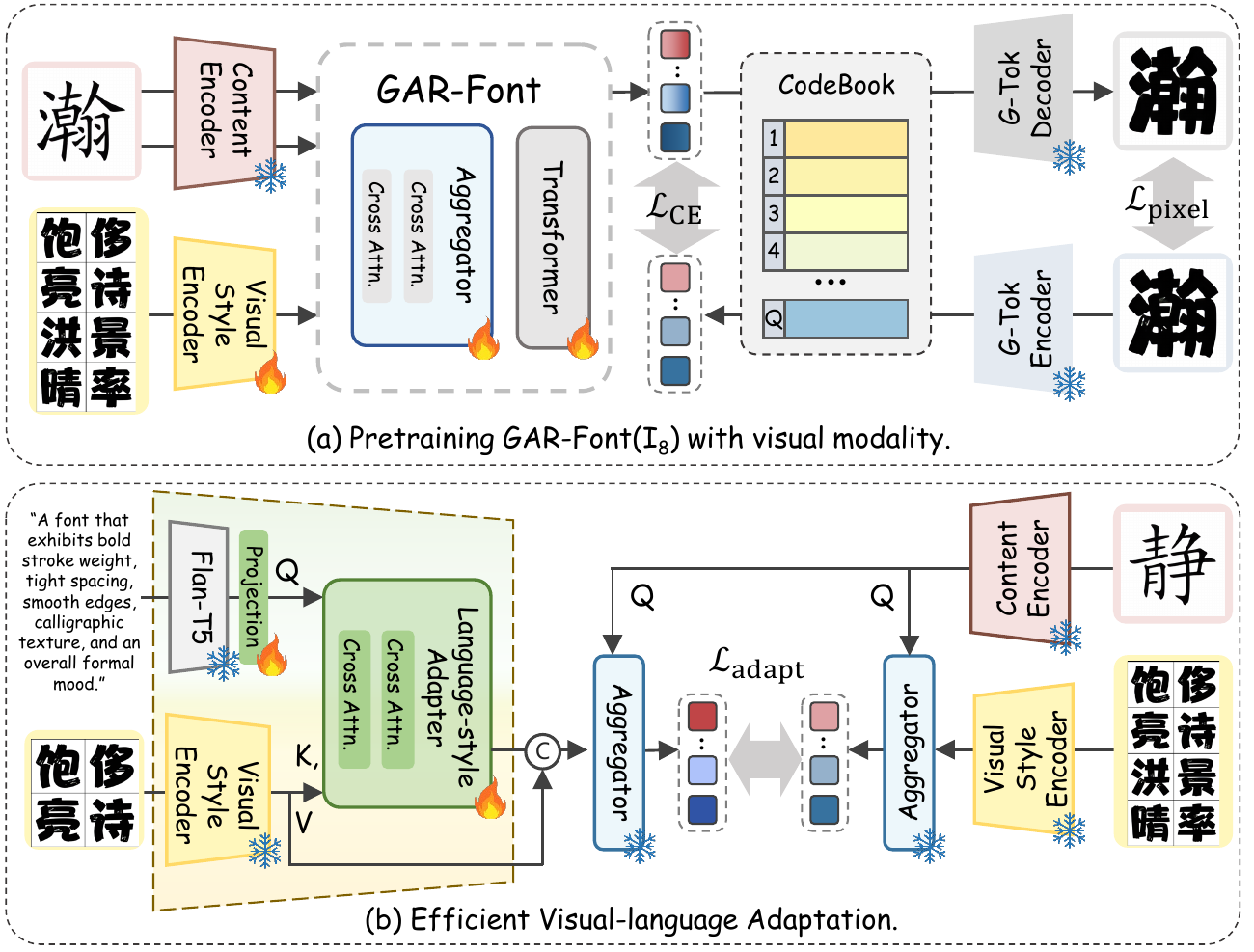}

    \caption{GAR-Font adopts a two-stage training: (a) Visual Pretraining builds a stable content–style space via token and pixel losses; (b) Vision–language Adaptation aligns text embeddings with style features with a lightweight adapter for text-guided generation while preserving visual priors.}
    \label{fig: Training}

\end{figure}

Visualized in Figure~\ref{fig: Training}(a), to build a stable style–content representation, the AR generator is first trained on visual conditions, following the successful practices of prior FFG methods. This ensures robust modeling of both content structure and stylistic variations across glyphs, providing a solid foundation for subsequent multimodal adaptation.

\paragraph{Encoders and Content–style Aggregator.} 
As shown in Figure~\ref{fig: Training}(a), we adopt both CNN architectures on the content encoder and the visual style encoder. The Content-style Aggregator follows the previous design~\cite{Fs_Font, LF_Font}. Given a content glyph and $N_s$ style references, the encoders respectively extract content features $\mathbf{F}_c$ and style features$\{\mathbf{F}_{vis_j}\}^{N_s}_{j=1}$. These features are then fused where content queries attend to fine-grained style cues:
\begin{equation}
\tilde{\mathbf{T}}_{\text{vis}} = \mathrm{Aggregator}\big(\mathbf{F}_c, \{\mathbf{F}_{\text{vis}_j}\}_{j=1}^{N_s}\big).
\end{equation}
The aggregated visual representation $\tilde{\mathbf{T}}_{\text{vis}}$ is concatenated with $\mathbf{F}_c$ to form the conditioning input $\mathbf{T}$.


\paragraph{Autoregressive modeling and soft decoding.} Conditioned on $\mathbf{T}$, the Transformer generator autoregressively predicts the next token, producing logits $\mathbf{L}$ over the G-Tok codebook $\mathcal{C}$. Instead of discrete hard decoding, we apply a soft projection that maps $\mathbf{L}$ onto the codebook 
$\tilde{\mathbf{Z}} = \mathrm{Softmax}(\mathbf{L}) \cdot \mathcal{C}$.
This continuous mapping not only preserves gradient flow during training, enabling pixel-level supervision, but can also improves stroke continuity and glyph fidelity by leveraging the full expressive capacity of the codebook, yielding smoother and more accurate glyphs compared with discrete hard decoding.

\paragraph{Training objectives.} During this visual-pretraining stage, the aggregator, generator, and style encoder are jointly optimized with a combined token and pixel-level loss:
\begin{equation}
\mathcal{L}_{\text{AR}} = \lambda_{\text{CE}}\mathcal{L}_{\text{CE}} + \lambda_{\text{pixel}}\mathcal{L}_{\text{pixel}},
\end{equation}
where $\mathcal{L}_{\text{CE}}$ is the cross-entropy loss over the target token indices $\mathbf{s}$, and $\mathcal{L}_{\text{pixel}}$ measures the L1 reconstruction error between the generated glyph $\hat{\mathbf{I}}_f$ and ground truth $\mathbf{I}_f$.

\subsubsection{Multimodal Style Encoder Adaptation}
Visual pretraining forms a stable style–content space but lacks high-level conceptual control. To address this, we introduce a lightweight, plug-in language adapter that aligns textual design cues with visual style representations, as shown in Figure~\ref{fig: Training}(b). Through this adapter, the style encoder is extended into a multimodal form, enabling flexible text-guided modulation without disrupting the visual prior.


\paragraph{Vision-language Adapter.} The adapter bridges textual descriptions and visual style features through iterative cross-attention. Specifically, a subset of visual style features $\{\mathbf{F}_{vis_j}\}^{k}_{j=1}$ is first extracted from $k<N_s$ reference glyphs by the visual style encoder, while a textual font description embedding is derived from a pretrained Flan-T5 encoder. The text embedding is projected into the visual feature space and iteratively refined by attending to style features. The aligned textual–visual token is then spatially expanded into $\mathbf{F}_t$ and concatenated with the visual features:

\begin{equation}
\tilde{\mathbf{F}}_{mm} = [\mathbf{F}_{vis_1}, \ldots, \mathbf{F}_{vis_k}, \mathbf{F}_t].
\end{equation}
\paragraph{Training objectives.} 
The multimodal style encoder may produce features $\tilde{\mathbf{F}}_{mm}$ of different token lengths than the visual-only encoder due to varying numbers of visual references. Thus, we supervise training on their aggregated objectives. Specifically, $\tilde{\mathbf{T}}_{\text{mm}}$ and $\tilde{\mathbf{T}}_{\text{vis}}$ are obtained by aggregating $k$ visual plus one textual reference, and all $N_s$ visual references, respectively. The adapter is trained by minimizing the $\ell_2$ distance between these representations:
\begin{equation}
\mathcal{L}_{\text{adapt}} = \|\tilde{\mathbf{T}}_{\text{mm}} - \tilde{\mathbf{T}}_{\text{vis}}\|_{2}^{2}.
\end{equation}

This objective encourages the multimodal encoder to internalize stylistic intent consistent with the full-visual setting, facilitating effective text–style substitution and reducing reliance on visual inputs.

\subsection{Post-refinement}
GAR-Font adopts a two-stage post-refinement to improve few-shot style generalization and structural accuracy: novel font adaptation (NFA) and structural enhancement (SE).

\paragraph{Novel font adaptation (NFA) for few-shot generalization.} The pretrained generator learns general font patterns but exhibits minor inconsistencies on unseen styles. NFA mitigates this by performing a lightweight adaptation using a few reference glyphs, updating the LoRA layers of the Transformer generator with a mixed token–pixel loss:
\begin{equation}
\mathcal{L}_{\text{NFA}}=\lambda_{\text{CE}}\mathcal{L}_{\text{CE}}+\lambda_{\text{pixel}}\mathcal{L}_{\text{pixel}},
\end{equation}
where $\mathcal{L}_{\text{CE}}$ is the token cross-entropy and $\mathcal{L}_{\text{pixel}}=\|\mathbf{I}-\hat{\mathbf{I}}\|_1$ encourages pixel-level accuracy. This yields stable few-shot adaptation and better preservation of unseen styles.

\paragraph{Structural enhancement (SE) for precise glyph reconstruction.}
While NFA enhances style fidelity, minor structural distortions may remain. To enhance structural clarity and readability, SE further refines glyphs using a group-relative optimization based on GRPO~\cite{DeepseekMath_GRPO}. The generator is treated as a policy $\pi_\theta$ that outputs token sequences $\mathbf{s}$; each decoded glyph receives a composite reward:
\begin{equation}
r=\lambda_{\text{ocr}}\,r_{\text{ocr}}+\lambda_{\text{style}}\,r_{\text{style}},
\end{equation}

where $r_{\text{ocr}}$ is obtained from a pretrained OCR model as

\begin{equation}
r_{\text{ocr}}=
\begin{cases}
p_{\text{ocr}}, & \text{if } \hat{y}=y,\\[3pt]
0, & \text{otherwise.}
\end{cases}
\end{equation}

Here $p_{\text{ocr}}$ is the recognition confidence, and $r_{\text{style}}$ measures the style consistency via a pretrained discriminator.


Rewards are normalized within each sampled group to compute advantages $A^{(k)} = \frac{r^{(k)} - \mu(r)}{\sigma(r)}$. SE updates only the LoRA layers by maximizing advantage-weighted likelihood with KL regularization to a frozen reference policy:

\begin{equation}
\mathcal{L}_{\text{SE}} \!=\! -\mathbb{E}_{\mathbf{s}\sim\pi_\theta}\big[A(\mathbf{s})\log\pi_\theta(\mathbf{s})\big]+\beta\,\mathrm{KL}(\pi_\theta\|\pi_{\text{ref}}),
\end{equation}
where $A(\mathbf{s})$ denotes the group-normalized advantage and $\pi_{\text{ref}}$ is the frozen reference policy used for stability.  

\begin{table*}[t!]
\centering
\caption{
Quantitative results on vision-only FFG. 
\textcolor{red!30}{\rule[0pt]{6pt}{6pt}} / 
\textcolor{blue!30}{\rule[0pt]{6pt}{6pt}} 
denote the best results on the \textit{Large} / \textit{Small} datasets, respectively. 
GAR-Font($I_8$) shows competitive metrics at pretraining and achieves top reconstruction and perceptual performance on the Large UFSC split after NFA-$8$+SE, demonstrating improved structural fidelity and perceptual quality.
}

\setlength{\tabcolsep}{3pt}
\renewcommand{\arraystretch}{1.00}
\resizebox{0.90\textwidth}{!}{
\begin{tabular}{l | c | c c c c c c | c c c c c c }
\toprule
Method & Train & \multicolumn{6}{c|}{Unseen Fonts Seen Characters (UFSC)} & \multicolumn{6}{c}{Unseen Fonts Unseen Characters (UFUC)} \\
\cmidrule(lr){3-8} \cmidrule(lr){9-14}
 &  Set & RMSE$\downarrow$ & SSIM$\uparrow$ & LPIPS$\downarrow$ & FID$\downarrow$ & Acc(C)$\uparrow$ & Acc(S)$\uparrow$
 & RMSE$\downarrow$ & SSIM$\uparrow$ & LPIPS$\downarrow$ & FID$\downarrow$ & Acc(C)$\uparrow$ & Acc(S)$\uparrow$ \\
\hline
LF-Font & \textit{S} & 0.3984 & 0.3276 & 0.2641 & 32.7464 & 0.4624 & 0.0148 & 0.3983 & 0.3299 & 0.2620 & 32.7028 & 0.4889 & 0.0160 \\
         & \textit{L} & 0.3988 & 0.3318 & 0.2450 & 21.1028 & 0.8989 & 0.0024 & 0.3986 & 0.3333 & 0.2451 & 21.8141 & 0.9082 & 0.0028 \\
\hline
VQ-Font & \textit{S} & \cellcolor{blue!20}\textbf{0.2727} & \cellcolor{blue!20}\textbf{0.5642} & 0.1830 & 35.2472 & 0.8763 & 0.0016 & \cellcolor{blue!20}\textbf{0.2744} & \cellcolor{blue!20}\textbf{0.5616} & 0.1822 & 36.7914 & 0.8882 & 0.0016 \\
         & \textit{L} & 0.2734 & 0.5633 & 0.1749 & 19.3103 & 0.8549 & 0.0014 & 0.2741 & 0.5627 & 0.1746 & 19.971 & 0.8434 & 0.0015 \\
\hline
DG-Font & \textit{S} & 0.3208 & 0.4991 & 0.1281 & 13.8392 & 0.9706 & 0.0764 & 0.3173 & 0.5074 & 0.1270 & 14.8249 & 0.9706 & 0.0797 \\
         & \textit{L} & 0.3117 & 0.5193 & 0.1235 & 17.1646 & 0.9172 & 0.1089 & 0.3074 & 0.5289 & 0.1214 & 17.2638 & 0.9174 & 0.1120 \\
\hline
CF-Font & \textit{S} & 0.3110 & 0.526 & 0.1301 & 18.6961 & 0.8542 & 0.0725 & 0.3077 & 0.5333 & 0.1282 & 19.309 & 0.8687 & 0.0747 \\
         & \textit{L} & 0.2993 & 0.5418 & 0.1155 & 13.354 & 0.8931 & 0.1549 & 0.2967 & 0.5474 & 0.1144 & 14.0878 & 0.8947 & 0.1570 \\
\hline
IF-Font & \textit{S} & 0.4076 & 0.3220 & 0.1713 & 14.2393 & 0.9804 & 0.0246 & 0.4063 & 0.3257 & 0.1724 & 14.8211 & 0.9750 & 0.0211 \\
         & \textit{L} & 0.3969 & 0.3374 & 0.1480 & 11.6470 & 0.9387 & 0.1148 & 0.3949 & 0.3433 & 0.1476 & 11.8445 & 0.9354 & 0.1153 \\
\hline
Diff-Font & \textit{S} & 0.3688 & 0.3903 & 0.1851 & 10.5722 & 0.4419 & 0.0515 & - & - & - & - & - & - \\
           & \textit{L} & 0.3651 & 0.3949 & 0.1791 & 9.7051 & 0.4029 & 0.1208 & - & - & - & - & - & - \\
\hline
Font-Diffuser & \textit{S} & 0.3010 & 0.4994 & 0.1728 & 26.2647 & \cellcolor{blue!20}\textbf{0.9994} & 0.0212 & 0.2999 & 0.5023 & 0.1720 & 26.9122 & \cellcolor{blue!20}\textbf{0.9999} & 0.0226 \\
               & \textit{L} & 0.2645 & 0.5813 & 0.1419 & 21.4246 & \cellcolor{red!20}\textbf{0.9979} & 0.0527 & 0.2631 & 0.5849 & 0.1407 & 21.9637 & \cellcolor{red!20}\textbf{0.9980} & 0.0578 \\
\Xhline{1.2pt}
GAR-Font($I_{8}$) & \textit{S} & 0.3080 & 0.5052 & 0.1313 & 7.9484 & 0.9408 & 0.0802 & 0.3142 & 0.4932 & 0.1421 & 8.4841 & 0.8993 & 0.0796 \\
 & \textit{L} & 0.2772 & 0.5799 & 0.1112 & 7.7155 & 0.9146 & 0.1928 & 0.2784 & 0.5787 & 0.1121 & 8.0349 & 0.8912 & 0.1892 \\
\hline
GAR-Font & \textit{S} & 0.2979 & 0.5418 & 0.1177 & \cellcolor{blue!20}\textbf{6.6909} & 0.9195 & \cellcolor{blue!20}\textbf{0.1128} 
& 0.3002 & 0.5354 & 0.1195 & \cellcolor{blue!20}\textbf{6.4693} & 0.9191 & \cellcolor{blue!20}\textbf{0.1112} \\
($I_{8}$, +NFA-$8$)  & \textit{L} & 0.2600 & 0.6158 & 0.0979 & \cellcolor{red!20}\textbf{6.5634} & 0.9210 & 0.3313 
& 0.2603 & 0.6160 & 0.0983 & \cellcolor{red!20}\textbf{6.5842} & 0.8921 & 0.3518 \\
\hline
GAR-Font & \textit{S} & 0.2909 & 0.5619 & \cellcolor{blue!20}\textbf{0.1111} & 8.4951 & 0.9817 & 0.1101 
& 0.2935 & 0.5553 & \cellcolor{blue!20}\textbf{0.1129} & 8.3504 & 0.9804 & 0.1025 \\
($I_{8}$, +NFA-$8$+SE)  & \textit{L} & \cellcolor{red!20}\textbf{0.2503} & \cellcolor{red!20}\textbf{0.6411} & \cellcolor{red!20}\textbf{0.0885} & 8.9851 & 0.9795 & \cellcolor{red!20}\textbf{0.3518}
& \cellcolor{red!20}\textbf{0.2540} & \cellcolor{red!20}\textbf{0.6356} & \cellcolor{red!20}\textbf{0.0903} & 8.6602 & 0.9670 & \cellcolor{red!20}\textbf{0.3735} \\
\bottomrule
\end{tabular}
}
\label{tab:table1}
\end{table*}
\begin{figure*}
    \centering
    \includegraphics[width=0.90\linewidth]{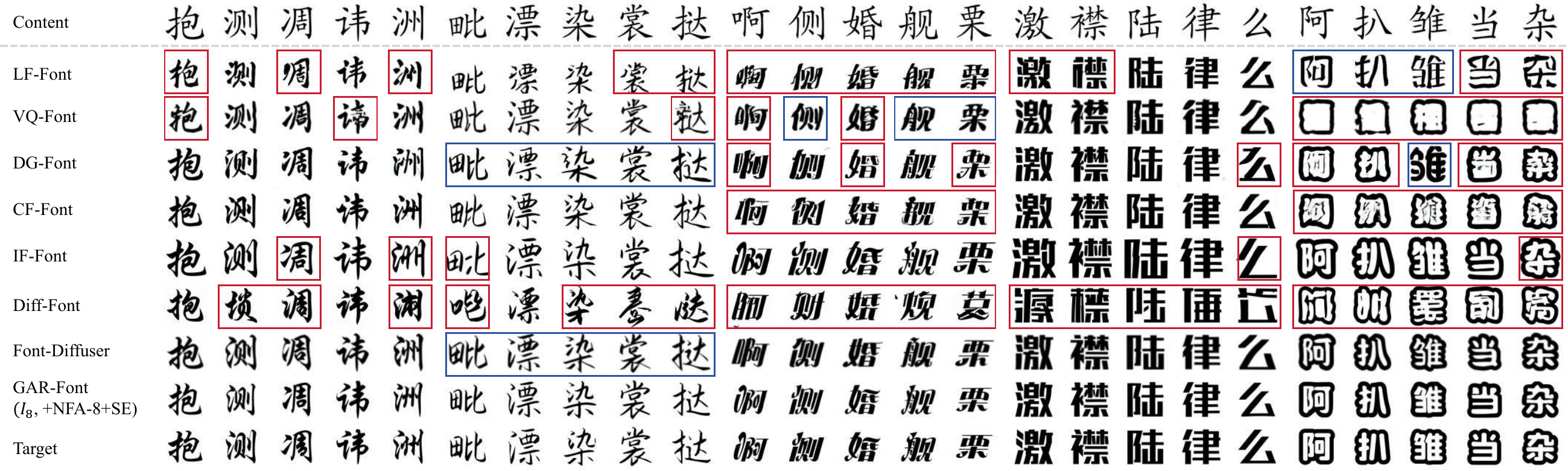}

    \caption{Qualitative results on vision-only FFG (UFSC, \textit{Large} dataset). \textcolor{ex1figure_red}{\fbox{\rule{0pt}{4pt}\rule{4pt}{0pt}}} / \textcolor{ex1figure_blue}{\fbox{\rule{0pt}{4pt}\rule{4pt}{0pt}}} indicate structural errors and style mismatches. GAR-Font($I_8$, +NFA-$8$+SE) produces the most faithful glyphs with superior structure and style alignment.} 
    \label{fig: Experiment1}

\end{figure*}

\section{Experiments}
\subsection{Datasets and Evaluation Metrics}
We evaluate GAR-Font and prior FFG methods on two datasets: a small-scale set containing 440 font styles (\textit{S}) and a large-scale set with 3,040 styles (\textit{L}). The small-scale dataset is a strict subset of the large-scale one, allowing fair cross-scale comparison. In each dataset, we randomly select 40 as unseen test fonts, while the remaining 400 (for \textit{S}) or 3,000 (for \textit{L}) are used for training.

Both datasets are constructed based on the official GB2312 character set, including 6,763 Chinese characters. Among them, 6,251 characters are randomly chosen for training, and the remaining 512 are held out as unseen characters. During evaluation, we consider two settings:

\noindent(1) \textit{Unseen Fonts Seen Characters (UFSC)}, where 512 seen characters are rendered with the 40 unseen fonts; and

\noindent(2) \textit{Unseen Fonts Unseen Characters (UFUC)}, where 512 unseen characters are rendered with the same unseen fonts.

We use RMSE$\downarrow$, SSIM$\uparrow$, LPIPS$\downarrow$, and FID$\downarrow$ to measure pixel- and perception-level similarity between generated and ground-truth glyphs. Following~\cite{LF_Font, HFH_Font}, we train a content classifier over 6,763 characters (99.71\% accuracy) and a style classifier over 3,040 fonts (92.72\% accuracy) to compute content accuracy (Acc(C)$\uparrow$) and style accuracy (Acc(S)$\uparrow$). All glyphs are resized to $64 \times 64$ for evaluation.

\subsection{Implementation Details}
The G-Tok tokenizer discretizes each glyph into 64 tokens using a 2,048-entry, dimension-8 codebook, trained for 200k iterations (batch = 16, lr = $1{\times}10^{-4}$). The vision-only AR generator uses the AdamW optimizer ($\beta_1 = 0.9$, $\beta_2 = 0.95$), conditioned on one Kaiti content. GAR-Font($I_8$) is trained with $N_s=8$ style glyphs for 600k/1M iterations on the small/large sets (batch = 32, lr = $1{\times}10^{-4}$).

For multimodal style encoder adaptation, the Language–style Adapter is trained for 40k iterations (batch size = 128, lr = $1{\times}10^{-4}$) with the visual-only features ($N_s=8$) as ground truth. By adjusting multimodal encoder's visual style reference numbers $k=2/4$, we build multimodal variants: GAR-Font($M_2$) and GAR-Font($M_4$).

In post-refinement, NFA is conducted on $8$ target font glyphs (denoted as NFA-$8$) for $10$ epochs (lr = $2{\times}10^{-5}$). SE is based on GRPO, where each group generates $4$ samples and each character is trained with $8$ glyphs from different fonts (epochs = $10$, batch size = $32$, learning rate = $5\mathrm {e}{-6}$). 

\subsection{Comparison on Few-shot Font Generation}
We conduct an experiment for \textbf{vision-only FFG}, where all models generate glyphs from only reference images. GAR-Font($I_{8}$) is compared with seven open-source methods: GAN/VAE-based LF-Font~\cite{LF_Font}, VQ-Font~\cite{VQ_Font}, DG-Font~\cite{DG_Font}, CF-Font~\cite{CF_Font}; diffusion-based Diff-Font~\cite{Diff_Font} and Font-Diffuser~\cite{Font_Diffuser}; and the AR-based IF-Font~\cite{IF_Font}. All methods are trained and evaluated on the \textit{S} and \textit{L} datasets using their official resolutions and hyperparameters. GAR-Font($I_8$) is evaluated both before and after post-refinement.

Table~\ref{tab:table1} shows that GAR-Font($I_{8}$) consistently outperforms existing vision-only FFG approaches. Even at the pretrained stage, it achieves competitive RMSE$\downarrow$ and SSIM$\uparrow$ on both \textit{UFSC}/\textit{UFUC}, and obtains the best FID$\downarrow$ scores on the large (\textit{L}) and small (\textit{S}) datasets. With NFA-$8$ and SE, GAR-Font further improves in both style fidelity and structural accuracy, achieving RMSE$\downarrow$ ($0.2503/0.2540$) and SSIM$\uparrow$ ($0.6411/0.6356$) on \textit{UFSC}/\textit{UFUC}, surpassing existing methods by a large margin.

Figure~\ref{fig: Experiment1} presents comparisons under the \textit{UFSC} setting. GAR-Font($I_8$, +NFA-$8$+SE) generates the most precise and coherent glyphs. By contrast, LF-Font, VQ-Font, DG-Font, CF-Font, and Diff-Font often exhibit stroke distortion or structural collapse on complex styles. IF-Font frequently produces incomplete glyphs with overly thick strokes. Font-Diffuser maintains reasonable structure and style but often shows slight style shifts in the results.

\begin{table*}[ht!]
\centering
\caption{
Quantitative results on multimodal FFG. GAR-Font($M_2$) and GAR-Font($M_4$) integrate textual style guidance with 2 or 4 visual references, outperforming vision-only baselines across all major metrics. Results demonstrate that language complements visual references, enhancing fine-grained style representation while reducing reliance on handcrafted visuals.}

\setlength{\tabcolsep}{3pt}
\renewcommand{\arraystretch}{1.02}
\resizebox{0.82\textwidth}{!}{
\begin{tabular}{l | c c c c c c | c c c c c c }
\toprule
Method & \multicolumn{6}{c|}{Unseen Fonts Seen Characters (UFSC)} & \multicolumn{6}{c}{Unseen Fonts Unseen Characters (UFUC)} \\
\cmidrule(lr){2-7} \cmidrule(lr){8-13}
 & RMSE$\downarrow$ & SSIM$\uparrow$ & LPIPS$\downarrow$ & FID$\downarrow$ & Acc(C)$\uparrow$ & Acc(S)$\uparrow$
 & RMSE$\downarrow$ & SSIM$\uparrow$ & LPIPS$\downarrow$ & FID$\downarrow$ & Acc(C)$\uparrow$ & Acc(S)$\uparrow$ \\
\hline
$n_\text{ref} = 2$& 0.2816 & 0.5695 & 0.1158 & 7.3553 & 0.9184 & 0.1535 & 0.2825 & 0.5694 & 0.1162 & 7.5500 & 0.8930 & 0.1619 \\

$n_\text{ref} = 4$& 0.2807 & 0.5735 & 0.1138 & 7.3781 & 0.9206 & 0.1741 & 0.2813 & 0.5735 & 0.1146 & \textbf{7.4880} & 0.8958 & 0.1818 \\
$n_\text{ref} = 8$& 0.2772 & 0.5799 & 0.1112 & 7.7155 & 0.9146 & \textbf{0.1928} & 0.2784 & 0.5787 & 0.1121 & 8.0349 & 0.8912 & \textbf{0.1892} \\
\Xhline{1.2pt}
GAR-Font($M_2$)  & 0.2811 & 0.5724 & 0.1136 & \textbf{7.3145} & \textbf{0.9296} & 0.1203
                 & 0.2817 & 0.5731 & 0.1143 & 7.5306 & \textbf{0.9068} & 0.1289 \\

GAR-Font($M_4$)   & \textbf{0.2764} & \textbf{0.5825} & \textbf{0.1098} & 7.4915 & 0.9260 & 0.1688 
                 & \textbf{0.2776} & \textbf{0.5816} & \textbf{0.1107} & 7.6607 & 0.9039 & 0.1744 \\
\bottomrule
\end{tabular}
}
\label{tab:adapter_experiment}

\end{table*}

\begin{figure*}[!t]
    \centering
    \includegraphics[width=0.86\linewidth]{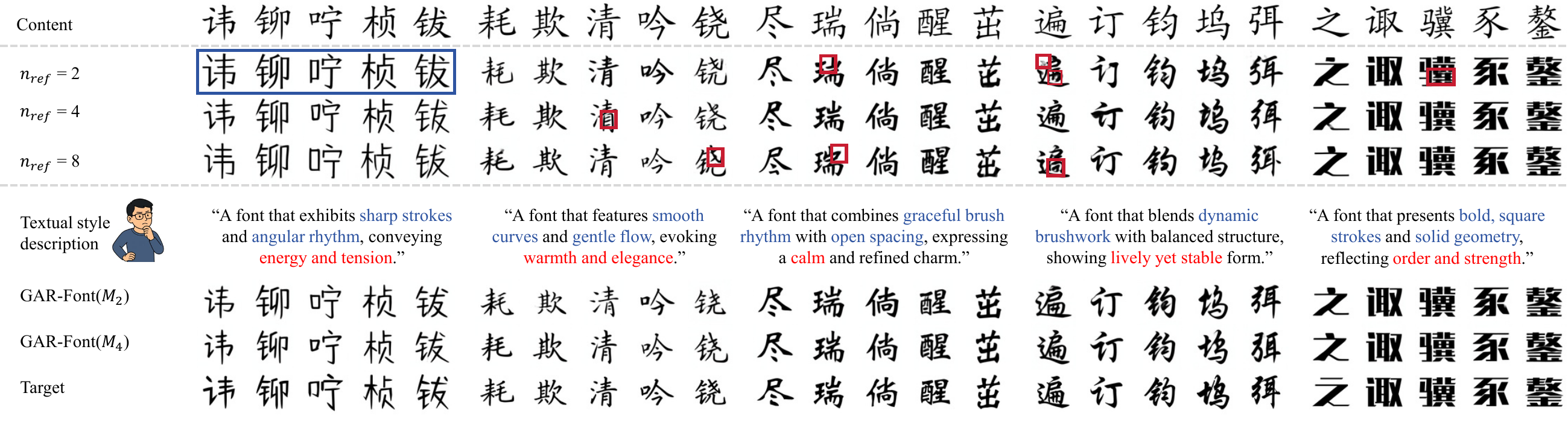}

    \caption{Qualitative results on multimodal FFG(UFSC, \textit{Large} dataset). \textcolor{ex2figure_red}{\fbox{\rule{0pt}{4pt}\rule{4pt}{0pt}}} denotes local slight structural mistakes and \textcolor{ex2figure_blue}{\fbox{\rule{0pt}{4pt}\rule{4pt}{0pt}}} marks samples with stylistic regression. With textual style guidance, GAR-Font($M_2$) and GAR-Font($M_4$) yield more structurally accurate and stylistically coherent generations, demonstrating enhanced stable fine-grained style control compared to vision-only baselines (\textit{first three rows}).} 
    \label{fig: Experiment2}

\end{figure*}

\subsection{Efficient Vision-Language Adaptation}
We further evaluate \textbf{multimodal FFG} to assess whether textual style descriptions can improve over vision-only references. Using GAR-Font($I_8$) on the large (\textit{L}) dataset, we vary the number of visual references $n_\text{ref}\in{2,4,8}$ as baselines. We then introduce multimodal variants GAR-Font($M_2$) and GAR-Font($M_4$), which pair 2 or 4 visual references with one textual style description. 
For experimental evaluation, due to the lack of such corpora, Qwen2.5-VL~\cite{Qwen25VL} generates descriptions as a proxy of human-authored design intent; details are provided in the supplementary material.

As shown in Table~\ref{tab:adapter_experiment}, both multimodal models outperform their vision-only counterparts with the same number of visual references across all major metrics. GAR-Font($M_4$) even surpasses the 8-reference visual model, achieving lower RMSE$\downarrow$/LPIPS$\downarrow$ and higher SSIM$\uparrow$ on the \textit{UFSC}, along with a better FID$\downarrow$ ($7.4915$ vs. $7.7155$). Similar gains appear on \textit{UFUC}, confirming that language provides complementary style cues and reduces reliance on numerous visual references. A mild decrease in ACC(S) is observed, likely because textual guidance yields smoother and more diverse styles beyond the classifier’s limit.

Qualitative examples in Figure~\ref{fig: Experiment2} show that GAR-Font($M_2$) and GAR-Font($M_4$) preserve styles more reliably than vision-only models, particularly when $n_\text{ref}=2$, where the visual baseline noticeably drifts toward generic shapes. This highlights the effectiveness of language in reinforcing style fidelity under low-reference conditions.

\subsection{Ablation Studies}
\subsubsection{On G-Tok's Hybrid Architecture}
To validate G-Tok's hybrid CNN-ViT design, we start from a CNN-based tokenizer~\cite{llamagen} and progressively insert ViT blocks at different depths to introduce global awareness.

We conduct two complementary analyses on \textit{UFUC} test:
\begin{enumerate}
\item \textbf{Linear Probing} evaluates the discriminative quality of frozen G-Tok representations for style and content prediction using a single-layer linear classifier. Features are extracted and flattened from the frozen tokenizer encoder, and high classification accuracy indicates that the encoder effectively captures and discriminates structural and stylistic information.
\item \textbf{Reconstruction Robustness} recovers glyphs corrupted by localized Gaussian noise ($\sigma = 0.2$, affecting 20\% of the glyph area). High performance indicates that the tokenizer effectively preserves global structure and stylistic cues despite local perturbations.
\end{enumerate}
Table~\ref{tab:ablation_linear_robust_combined} shows that the pure ViT excels in linear probing but suffers in reconstruction, while the CNN baseline ensures stable recreation with weaker probing results. The hybrid G-Tok progressively integrates global attention, with the CNN-ViT-6 variant achieving the best, improving both classification accuracy and reconstruction fidelity across all metrics. Visualizations are in the supplementary material.

\begin{table}[!t]
\centering
\caption{
Ablation of G-Tok's hybrid CNN-ViT architecture on UFUC(\textit{Small} dataset). \textbf{Bold} and \underline{underline} indicate the best and second-best results. Progressive integration of attention improves both discriminative representation and reconstruction fidelity.
}

\renewcommand{\arraystretch}{1.02}
\resizebox{0.46\textwidth}{!}{
\begin{tabular}{l | c c | c c c c }
\toprule
Method & \multicolumn{2}{c|}{Linear Probing} & \multicolumn{4}{c}{Reconstruction Robustness} \\
\cmidrule(lr){2-3} \cmidrule(lr){4-7}
 & Acc(S)$\uparrow$ & Acc(C)$\uparrow$ & RMSE$\downarrow$ & SSIM$\uparrow$ & LPIPS$\downarrow$ & FID$\downarrow$ \\
\hline
CNN       & 0.5515 & 0.3879 & 0.1167 & 0.8535 & 0.0423 & 28.4279 \\
ViT-6       & \textbf{0.6907} & \textbf{0.5334} & 0.1636 & 0.7333 & 0.0933 & 98.4270 \\
CNN-ViT-2 & 0.5229 & 0.3772 & 0.1114 & 0.8552 & \underline{0.0420} & 30.0034 \\
CNN-ViT-4 & 0.5585 & 0.3667 & \underline{0.1114} & \underline{0.8563} & 0.0430 & \underline{24.7323} \\
CNN-ViT-6 & \underline{0.6277} &\underline{0.4897} & \textbf{0.1088} & \textbf{0.8594} & \textbf{0.0412} & \textbf{22.1577}\\
\bottomrule
\end{tabular}
}
\label{tab:ablation_linear_robust_combined}

\end{table}

\begin{table}[!t]
\centering
\caption{Ablation of G-Tok’s architecture on UFUC(\textit{Small} dataset). Incorporating self-attention (CNN+Non-Causal ViT) improves global modeling while causal attention further strengthens sequential modeling, yielding the best overall performance.}

\setlength{\tabcolsep}{3pt}
\renewcommand{\arraystretch}{1.02}
\resizebox{0.46\textwidth}{!}{
\begin{tabular}{l | c c c c c c }
\toprule
Method & RMSE$\downarrow$ & SSIM$\uparrow$ & LPIPS$\downarrow$ & FID$\downarrow$ & Acc(C)$\uparrow$ & Acc(S)$\uparrow$ \\
\hline
CNN & 0.3447 & 0.4350 & 0.1728 & 10.5239 & 0.6722 & 0.0221 \\
CNN+Non-Causal ViT & 0.3271 & 0.4745 & 0.1562 & 8.7504 & 0.8019 & 0.0436 \\
CNN+Causal ViT & \textbf{0.3142} & \textbf{0.4932} & \textbf{0.1421} & \textbf{8.4841} & \textbf{0.8993} & \textbf{0.0796} \\
\bottomrule
\end{tabular}
}
\label{tab:ablation_gtok_generator}

\end{table}

\begin{table*}[t!]
\centering
\caption{
Ablation of decoding strategy and pixel-level supervision on UFSC and UFUC(\textit{Small} dataset).
Soft decoding outperforms hard decoding across all metrics, and pixel supervision further enhances reconstruction fidelity and recognition accuracy.
}

\setlength{\tabcolsep}{3pt}
\renewcommand{\arraystretch}{0.98}
\resizebox{0.82\textwidth}{!}{
\begin{tabular}{l |c| c c c c c c | c c c c c c }
\toprule
Training Loss & Decoding &\multicolumn{6}{c|}{Unseen Fonts Seen Characters (UFSC)} & \multicolumn{6}{c}{Unseen Fonts Unseen Characters (UFUC)} \\
\cmidrule(lr){3-8} \cmidrule(lr){9-14}
& Strategy & RMSE$\downarrow$ & SSIM$\uparrow$ & LPIPS$\downarrow$ & FID$\downarrow$ & Acc(C)$\uparrow$ & Acc(S)$\uparrow$ & 
RMSE$\downarrow$ & SSIM$\uparrow$ & LPIPS$\downarrow$ & FID$\downarrow$ & Acc(C)$\uparrow$ & Acc(S)$\uparrow$ \\
\hline
w/o pixel loss & hard & 0.3235 & 0.4679 & 0.1517 & 10.3181 & 0.8647 & 0.0377 & 0.3322 & 0.4510 & 0.1602 & 11.2142 & 0.7465 & 0.0396 \\

         & soft & 0.3231 & 0.4745 & 0.1502 & 9.6696 & 0.9171 & 0.0426 & 0.3313 & 0.4583 & 0.1589 & 10.3813 & 0.8104 & 0.0430 \\
\hline
w/ pixel loss & hard & 0.3083 & 0.4991 & 0.1329 & 8.3024 & 0.9032 & 0.0771 & 0.3157 & 0.4862 & 0.1448 & 8.9043 & 0.8404 & 0.0680 \\

 & soft & \textbf{0.3080} & \textbf{0.5052} & \textbf{0.1313} & \textbf{7.9484} & \textbf{0.9408} & \textbf{0.0802} & \textbf{0.3142} & \textbf{0.4932} & \textbf{0.1421} & \textbf{8.4841} & \textbf{0.8993} & \textbf{0.0796} \\

\bottomrule
\end{tabular}
}
\label{tab:ablation_decode_strategy}

\end{table*}
\begin{table*}[t!]
\centering
\caption{
Quantitative comparison of joint training (GAR-Font($VL_k$)) versus the decoupled adapter scheme (GAR-Font($M_k$)) on unseen fonts. The decoupled Language-style Adapter surpasses joint-trained multimodal encoders, validating more effective vision–language alignment and improved reconstruction and perceptual metrics.
}

\setlength{\tabcolsep}{3pt}
\renewcommand{\arraystretch}{0.98}
\resizebox{0.82\textwidth}{!}{
\begin{tabular}{l | c c c c c c | c c c c c c }
\toprule
Method & \multicolumn{6}{c|}{Unseen Fonts Seen Characters (UFSC)} & \multicolumn{6}{c}{Unseen Fonts Unseen Characters (UFUC)} \\
\cmidrule(lr){2-7} \cmidrule(lr){8-13}
 & RMSE$\downarrow$ & SSIM$\uparrow$ & LPIPS$\downarrow$ & FID$\downarrow$ & Acc(C)$\uparrow$ & Acc(S)$\uparrow$
 & RMSE$\downarrow$ & SSIM$\uparrow$ & LPIPS$\downarrow$ & FID$\downarrow$ & Acc(C)$\uparrow$ & Acc(S)$\uparrow$ \\
\hline
GAR-Font($VL_2$)  & 0.3070 & 0.5124 & 0.1458 & 10.7566 & 0.8552 & 0.1104
                 & 0.3100 & 0.5087 & 0.1473 & 11.0381 & 0.7979 & 0.1209 \\

GAR-Font($VL_4$)  & 0.2983 & 0.5378 & 0.1279 & 7.9272 & 0.8970 & 0.1165
                 & 0.3006 & 0.5351 & 0.1290 & 8.1126 & 0.8665 & 0.1216 \\

GAR-Font($M_2$)  & 0.2811 & 0.5724 & 0.1136 & \textbf{7.3145} & \textbf{0.9296} & 0.1203
                 & 0.2817 & 0.5731 & 0.1143 & \textbf{7.5306} & \textbf{0.9068} & 0.1289 \\

GAR-Font($M_4$)  & \textbf{0.2764} & \textbf{0.5825} & \textbf{0.1098} & 7.4915 & 0.9260 & \textbf{0.1688} 
                 & \textbf{0.2776} & \textbf{0.5816} & \textbf{0.1107} & 7.6607 & 0.9039 & \textbf{0.1744} \\
\bottomrule
\end{tabular}
}
\label{tab:generator_with_adapter_experiment}

\end{table*}

\subsubsection{On G-Tok's Global and Causal Modeling}
We further ablate G-Tok's ViT architecture to assess its global and causal modeling. We pretrain three visual AR variants that differ in G-Tok and examine them on \textit{UFUC}: 
\begin{enumerate}
\item \textbf{CNN:} a baseline CNN tokenizer without global context;
\item \textbf{CNN + Non-causal ViT:} a hybrid CNN-ViT tokenizer with self-attention that only supports global interaction;
\item \textbf{CNN + Causal ViT:} full G-Tok, hybrid CNN-ViT tokenizer with causal self-attention for sequential modeling.
\end{enumerate}
Table~\ref{tab:ablation_gtok_generator} shows that incorporating self-attention ViT improves performance over the CNN baseline, while the causal ViT further enhances sequential modeling, achieving the best results across all metrics. Detailed \textit{UFSC} results are provided in the supplementary material.

\subsubsection{On AR Generator's Soft-decoding}
To validate the effectiveness of pixel-level supervision and the soft decoding strategy, we conduct ablation experiments on our vision-only generator.
The results in Table~\ref{tab:ablation_decode_strategy} demonstrate that soft decoding achieves superior performance over all metrics, particularly when combined with pixel-level supervision. 
This combination yields the lowest RMSE$\downarrow$, LPIPS$\downarrow$, and FID$\downarrow$ while achieving the highest SSIM$\uparrow$ and classifier accuracies on \textit{UFSC} and \textit{UFUC}, indicating improved pixel-level fidelity and faithful style preservation.

\subsubsection{On Multimodal Style Encoder's Adaptation}
GAR-Font adopts a decoupled training paradigm: a visual encoder is pretrained, followed by multimodal control via a lightweight adapter. To validate this design, we compare it with jointly training a multimodal style encoder under the same configuration on the large (\textit{L}) dataset.

We evaluate two joint training variants, \textbf{GAR-Font($VL_2$)} and \textbf{GAR-Font($VL_4$)}, against the decoupled multimodal variants \textbf{GAR-Font($M_2$)} and \textbf{GAR-Font($M_4$)}. As Table~\ref{tab:generator_with_adapter_experiment} shows, decoupled variants consistently outperform joint-trained counterparts under the same number of visual references. For example, GAR-Font($M_4$) attains the best RMSE$\downarrow$, LPIPS$\downarrow$, and SSIM$\uparrow$ on \textit{UFSC} / \textit{UFUC}, while GAR-Font($VL_4$) performs worse. GAR-Font($M_2$) also achieves the best FID$\downarrow$ on \textit{UFSC} ($7.3145$). These results validate the decoupled vision–language training paradigm, where a pretrained visual encoder and a lightweight adapter achieve efficient and robust text–visual alignment, improving generalization in multimodal FFG.

\subsection{Limitations and Future Work}
Although GAR-Font surpasses existing FFG methods in visual fidelity, several limitations remain. First, the Multimodal Style Encoder currently relies on late adaptation via a language-style adapter; exploring earlier text-image fusion may enable finer stylistic control and reduce reliance on visual references. Second, our experiments are conducted at a resolution of $64 \times 64$, which may limit direct applicability to high-DPI scenarios; extending the framework to higher resolutions may require G-Tok to handle longer token sequences. Finally, we aim to expand controllability beyond style and content, adding attributes like stroke thickness, character width, and slant for more flexible font generation.

\section{Conclusion}
In this paper, we present GAR-Font, a global-aware autoregressive framework for few-shot font generation that combines a hybrid global tokenizer, an autoregressive generator with a multimodal style encoder, and a post-refinement pipeline. GAR-Font achieves superior structural and stylistic fidelity, outperforming prior vision-only baselines, while demonstrating that language-guided adaptation can rival or exceed heavy visual conditioning with improved flexibility.

\section*{Acknowledgments}
This work was supported by National Natural Science Foundation of China (Grant No.: 62372015), Leading Projects in Key Research Fields of Language Funded by the National Language Commission, Center For Chinese Font Design and Research, Key Laboratory of Intelligent Press Media Technology, and National Engineering Research Center of New Electronic Publishing Technologies.
\clearpage
\setcounter{page}{1}
\setcounter{section}{0}
\setcounter{figure}{0}
\setcounter{table}{0}
\maketitlesupplementary
\renewcommand\thesection{\Alph{section}}
\renewcommand{\thetable}{S\arabic{table}}    
\renewcommand{\thefigure}{S\arabic{figure}}  

\section{Overview}

This supplementary material provides additional details on dataset construction, implementation, and extended experiments supporting the main paper. The sections are organized as follows:
\begin{itemize}
    \item \textbf{\cref{supplesec:sup_models}}: Model configurations of G-Tok, the autoregressive generator, and the multimodal style encoder.
    \item \textbf{\cref{supplesec:sup_data_dist}}: Data curation and partitioning, covering both training and evaluation protocols.
    \item \textbf{\cref{supplesec:sup_quant_ext}}: Extended quantitative results, including scaling analyses of our autoregressive generator.
    \item \textbf{\cref{supplesec:sup_Ablate}}: Detailed and visualized ablations of key design choices in GAR-Font.
    \item \textbf{\cref{supplesec:sup_vis_results}}: Qualitative results for visual-only and multimodal FFG, cross-language, and higher-resolution generation.
    \item \textbf{\cref{supplesec:sup_Failure_cases}}: Analysis of GAR-Font failure cases in dense-stroke and complex font styles.
\end{itemize}

\section{Model Configuration}
\label{supplesec:sup_models}
\cref{tab:model-structure} details the architectural specifications of GAR-Font. The framework relies on three core components: (1) The \textbf{Global-aware Tokenizer (G-Tok)} , which employs a hybrid CNN-ViT to discretize glyphs into a compact codebook; (2) The \textbf{Autoregressive Generator}, which serves as the synthesis backbone, using a Transformer decoder to predict tokens conditioned on aggregated content and style features; and (3) The \textbf{Multimodal Style Encoder}, which utilizes a lightweight adapter to align textual embeddings with visual style features for text-driven control.
\begin{table}[h]
\centering
\small
\setlength{\tabcolsep}{2.5pt} 
\renewcommand{\arraystretch}{0.91}
\begin{tabular}{p{0.8\columnwidth} c}
\toprule
\textbf{Key Components} & \textbf{Params (M)} \\
\midrule
\textbf{G-Tok} & 79.59 \\
CNN Encoder & 28.56 \\
ViT Encoder (layers = 6) & 4.73 \\
Codebook (size = 2048, dim = 8) & 0.02 \\
ViT Decoder (layers = 6) & 4.73 \\
CNN Decoder & 41.42 \\
\midrule
\textbf{AR-Generator} &  346.23 \\
Content Encoder & 28.56 \\
Visual Style Encoder & 2.78 \\
Content-style Aggregator (layers = 3) & 0.79 \\
Transformer Decoder (layers = 24) & 314.10 \\
\midrule
\textbf{Multimodal Style Encoder} & 8.04 \\
Projection & 0.52 \\
Visual Style Encoder & 2.78 \\
Language-Style Adapter (layers = 6) & 4.74 \\
\bottomrule
\end{tabular}

\caption{Key GAR-Font components and parameter counts.}

\label{tab:model-structure}
\end{table}
\begin{figure}[t]
  \centering
  \includegraphics[width=0.92\linewidth]{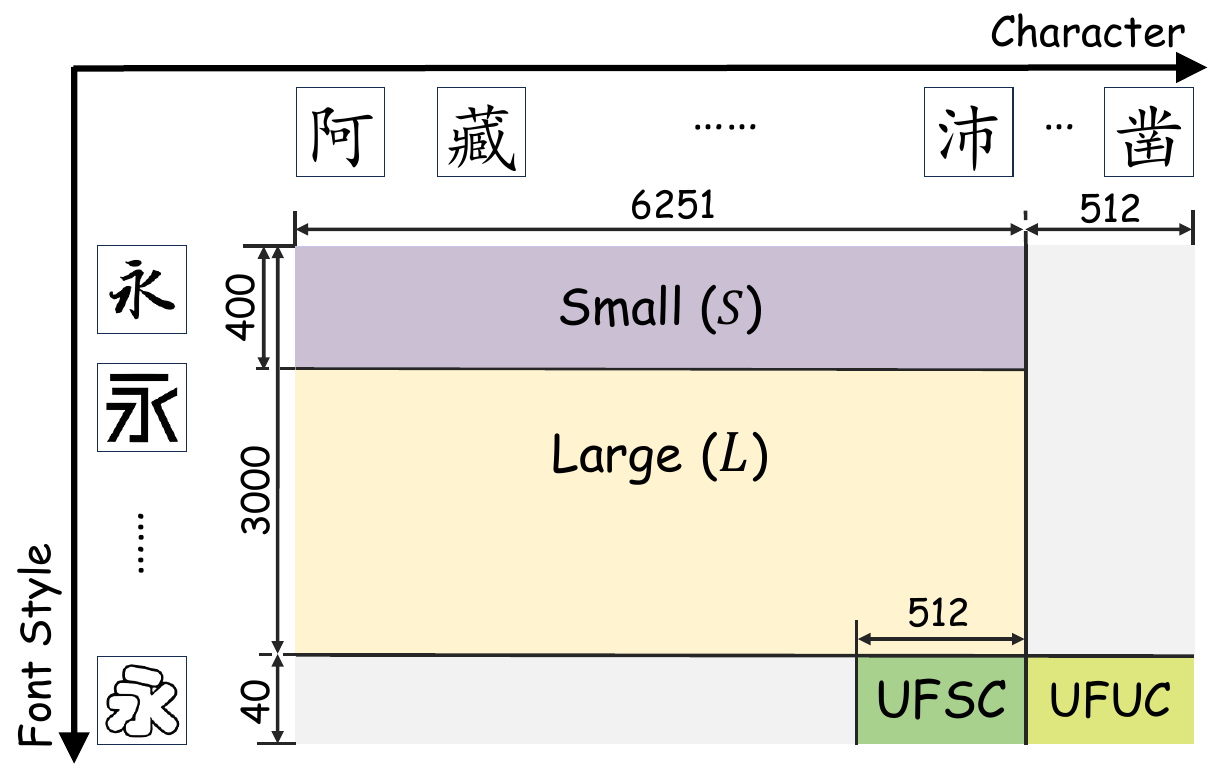} 

  \caption{\textbf{Visual illustration of the dataset partition.} The data is organized along font and character axes. Pre-training utilizes the purple and yellow regions ($S$ and $L$). Evaluation is conducted on the bottom green regions (\textit{UFSC} and \textit{UFUC}), strictly isolating unseen styles and characters.}

  \label{fig:data_viz}
\end{figure}

\section{Data Curation}
\label{supplesec:sup_data_dist}

\subsection{Data Collection and Statistics}
\label{supplesubsec:data_stats}
We construct a comprehensive font dataset derived from the official GB2312 character set. As illustrated in \cref{fig:data_viz}, the whole training and test dataset is structured as a matrix spanned by two orthogonal axes: \textit{Font Style} (vertical axis) and \textit{Character Category} (horizontal axis). 
The collected data comprises 3,040 fonts and 6,763 characters. 

For training data, along the Character Axis, we split 6,251 training characters (left column) and left 512 characters unseen (right column). The unseen characters are reserved strictly for testing to evaluate the model's capability to generate novel glyph structures. Similarly, the font library is divided into 3,000 training fonts (top rows) and 40 unseen test fonts (bottom rows).

\subsection{Pretraining Data}
\label{supplesubsec:data_pretrain}
The pretraining phase utilizes the data located in the upper-left quadrant of \cref{fig:data_viz}, defined by the intersection of training fonts and training characters. Within this quadrant, we define two configurations to investigate scaling behaviors:
\begin{itemize}[leftmargin=*]
    \item Large ($L$): The full training block consisting of all 3,000 training fonts paired with the 6,251 training characters (represented by the blue region).
    \item Small ($S$): A subset consisting of the first 400 training fonts paired with the same 6,251 characters (represented by the reddish overlay).
\end{itemize}
Training on $S$ versus $L$ allows us to assess the model's data efficiency and performance scaling with respect to the diversity of source styles.

\subsection{Textual Prompt Collection}
\label{supplesubsec:prompt}

To support multimodal few-shot font generation (FFG), we construct a consistent textual prompt set that captures font-level stylistic attributes. Since human-authored font design descriptions are not available in existing datasets, we automatically generate textual prompts to approximate human design intent. For each font, we randomly sample 40 glyph images and jointly input them into Qwen2.5-VL. The model is instructed to produce a single, unified description summarizing only the visual properties that remain consistent across the full glyph set—such as stroke weight, curvature, structural proportions, spatial rhythm, edge texture, and overall tonal characteristics. This process yields a controlled and stylistically coherent textual representation for each font.

The exact prompt used for textual description extraction is provided below:

\begin{FontExtractPrompt}
You are an experienced typographic style analyst. You are given a set of glyph images belonging to the same font. Your task is to synthesize a unified stylistic description that captures only the consistent, font-level visual attributes shared across the full glyph set.

Your output must adhere to the following specifications:

\textbf{1. Required Format}  

Provide a single paragraph that:

- begins with the phrase “A font that …”,

- contains approximately 45–50 words,

- includes only stylistic properties observable across all glyphs,

- avoids speculative or uncertain expressions.

\textbf{2. Allowed Stylistic Dimensions}  

Constrain your analysis to the following attributes:

- stroke weight (light, medium, bold, uniform, contrasting),

- curvature (straight, angular, rounded, flowing, sharp),

- structural proportions (compact, tall, wide, balanced),

- spacing and rhythm (tight, loose, even, irregular),

- edge rendering (smooth, sharp, rough, brush-like),

- overall tone or mood (elegant, modern, classical, playful, gentle, formal).

\textbf{3. Constraints}  

All statements must be visually grounded in the provided glyph set. Do not reference features specific to individual characters. The description must reflect global stylistic coherence and maintain typographic precision.

\end{FontExtractPrompt}

\subsection{Post-Refinement Data}
\label{supplesubsec:data_refine}
To further adapt the model to novel styles and enhance structural consistency, we employ specific data subsets:

\noindent Novel Font Adaptation (NFA). NFA adapts the pre-trained model to the style of the 40 unseen test fonts. For each test font, we sample 8 characters (NFA-$8$) from the 6,251 training character set to serve as style references. This process operates within the vertical column of the training characters but focuses on the unseen font rows.

\noindent Structural Enhancement (SE). SE aims to consolidate global glyph consistency. It utilizes the entire 6,763 characters (spanning both training and unseen characters) but restricts the style to a manageable subset of 400 fonts (sampled from $S$). This ensures the model sees a complete range of structural geometries during the refinement phase without the computational cost of the full font library.

\subsection{Evaluation Data}
\label{supplesecsubsec:data_eval}
Evaluation is strictly conducted on the held-out bottom rows of the matrix (\cref{fig:data_viz}), ensuring no overlap with the pre-training data. We define two rigorous settings:
\begin{itemize}[leftmargin=*]
    \item \textit{UFSC} (Unseen Fonts, Seen Characters): Represented by the light green region. This setting evaluates the model's ability to stylize known characters into novel font styles.
    \item \textit{UFUC} (Unseen Fonts, Unseen Characters): Represented by the dark green region. This is the most challenging zero-shot setting, where the model must generate glyphs that are novel in both style and structure.
\end{itemize}


\section{More Quantitative Experiments}
\label{supplesec:sup_quant_ext}
\begin{table}[t]
\centering

\caption{Quantitative evaluation on VQ-Font vs. VQ-Font (G). Replacing the original VQ-VAE with G-Tok halves the FID score and boosts content accuracy by nearly 9\%.}

\setlength{\tabcolsep}{3pt}
\renewcommand{\arraystretch}{1.1}
\small
\resizebox{\linewidth}{!}{
\begin{tabular}{lcccccc}
\toprule
\multicolumn{7}{c}{\textbf{Unseen Fonts Seen Characters (UFSC)}} \\
\midrule
Method & RMSE$\downarrow$ & SSIM$\uparrow$ & LPIPS$\downarrow$ & FID$\downarrow$ & Acc(C)$\uparrow$ & Acc(S)$\uparrow$ \\
\midrule
VQ\_Font        & 0.2727 & 0.5642 & 0.1830 & 35.2472 & 0.8763 & 0.0016 \\
VQ\_Font (G)    & \textbf{0.2725} & \textbf{0.5644} & \textbf{0.1731} & \textbf{17.3296} & \textbf{0.9646} & \textbf{0.0022} \\
\midrule\midrule
\multicolumn{7}{c}{\textbf{Unseen Fonts Unseen Characters (UFUC)}} \\
\midrule
Method & RMSE$\downarrow$ & SSIM$\uparrow$ & LPIPS$\downarrow$ & FID$\downarrow$ & Acc(C)$\uparrow$ & Acc(S)$\uparrow$ \\
\midrule
VQ\_Font        & 0.2744 & 0.5616 & 0.1822 & 36.7914 & 0.8882 & 0.0016 \\
VQ\_Font (G)    & \textbf{0.2732} & \textbf{0.5637} & \textbf{0.1731} & \textbf{17.9152} & \textbf{0.9653} &\textbf{0.0023} \\
\bottomrule
\end{tabular}
}
\label{tab:quant_eval_gtok_adaptation}
\end{table}
\begin{figure}[t]
  \centering
  \includegraphics[width=0.95\linewidth]{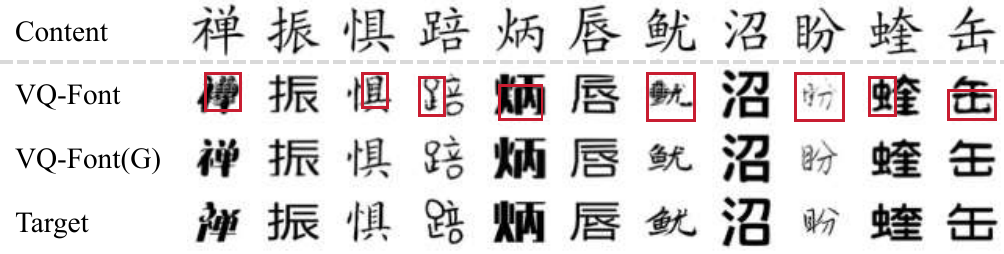} 
  \caption{
Qualitative comparison on VQ-Font vs. VQ-Font (G). 
\textcolor{ex2figure_red}{\fbox{\rule{0pt}{4pt}\rule{4pt}{0pt}}} marks structural errors in the generated glyph. 
G-Tok improves structure preservation and produces more faithful font styles.
}
  \label{fig:supple_vq_g_ufuc}
\end{figure}

\begin{table*}[t!]
\centering
\caption{Quantitative results of NFA glyph number ablation for few-shot font adaptation. Increasing from NFA-$8$ to NFA-$128$ consistently improves style faithfulness and perceptual quality. Applying SE further enhances structural fidelity.}

\setlength{\tabcolsep}{3pt}
\renewcommand{\arraystretch}{1.02}
\resizebox{0.94\textwidth}{!}{
\begin{tabular}{l | c | c c c c c c | c c c c c c }
\toprule
Method & Train & \multicolumn{6}{c|}{Unseen Fonts Seen Characters (UFSC)} & \multicolumn{6}{c}{Unseen Fonts Unseen Characters (UFUC)} \\
\cmidrule(lr){3-8} \cmidrule(lr){9-14}
 &  Set & RMSE$\downarrow$ & SSIM$\uparrow$ & LPIPS$\downarrow$ & FID$\downarrow$ & Acc(C)$\uparrow$ & Acc(S)$\uparrow$
 & RMSE$\downarrow$ & SSIM$\uparrow$ & LPIPS$\downarrow$ & FID$\downarrow$ & Acc(C)$\uparrow$ & Acc(S)$\uparrow$ \\
\hline
NFA-$8$ & \textit{S} & 0.2979 & 0.5418 & 0.1177 & 6.6909 & 0.9195 & 0.1128 
& 0.3002 & 0.5354 & 0.1195 & 6.4693 & 0.9191 & 0.1112 \\
 & \textit{L} & 0.2600 & 0.6158 & 0.0979 & 6.5634 & 0.9210 & 0.3313 
& 0.2603 & 0.6160 & 0.0983 & 6.5842 & 0.8921 & 0.3518 \\
\hline
NFA-$8$+SE & \textit{S} & 0.2909 & 0.5619 & 0.1111 & 8.4951 & 0.9817 & 0.1101 
& 0.2935 & 0.5553 & 0.1129 & 8.3504 & 0.9804 & 0.1025 \\
 & \textit{L} & 0.2503 & 0.6411 & 0.0885 & 8.9851 & 0.9795 & 0.3518
& 0.2540 & 0.6356 & 0.0903 & 8.6602 & 0.9670 & 0.3735 \\

\hline
NFA-$32$ & \textit{S} & 0.2880 & 0.5598 & 0.1095 & 5.7662 & 0.8993 & 0.1786 & 0.2849 & 0.5679 & 0.1075 & 5.8886 & 0.8962 & 0.1978 \\
       & \textit{L} & 0.2561 & 0.6238 & 0.0949 & 6.1329 & 0.9093 & 0.3707 & 0.2570 & 0.6234 & 0.0957 & 6.1493 & 0.8758 & 0.3946 \\
\hline
NFA-$32$+SE & \textit{S} & 0.2803 & 0.5821 & 0.1028 & 7.1792 & 0.9734 & 0.1754 & 0.2781 & 0.5878 & 0.1012 & 7.1134 & 0.9724 & 0.1970 \\
            & \textit{L} & 0.2455 & 0.6515 & 0.0854 & 8.2308 & 0.9775 & 0.4048 & 0.2460 & 0.6513 & 0.0862 & 7.8937 & 0.9581 & 0.4342 \\
\hline
NFA-$128$ & \textit{S} & 0.2712 & 0.5933 & 0.0992 & 5.4254 & 0.9236 & 0.3179 & 0.2836 & 0.5702 & 0.1079 & 5.6667 & 0.8817 & 0.3277 \\
        & \textit{L} & 0.2435 & 0.6507 & 0.0855 & 5.7570 & 0.9228 & 0.4457 & 0.2496 & 0.6397 & 0.0904 & 5.8078 & 0.8830 & 0.4625 \\
\hline
NFA-$128$+SE & \textit{S} & 0.2671 & 0.6089 & 0.0948 & 6.9103 & 0.9718 & 0.2833 & 0.2788 & 0.5884 & 0.1022 & 7.1309 & 0.9242 & 0.2958 \\
             & \textit{L} & 0.2398 & 0.6627 & 0.0831 & 8.3751 & 0.9776 & 0.4154 & 0.2508 & 0.6377 & 0.0908 & 7.2034 & 0.9068 & 0.4183 \\
\bottomrule
\end{tabular}
}
\label{tab:ablation_nfa_number}
\end{table*}

\subsection{Adaptation on G-Tok to Other FFG Methods}
\label{supplesubsec:quantitiative_G-Tok_plugin}
To verify the versatility of our G-Tok, we integrated it into VQ-Font by replacing its native VQ-VAE with our G-Tok while maintaining the original model architecture and configuration. The modified model, \textbf{VQ-Font (G)}, was trained on the Small dataset with G-Tok. As shown in \cref{tab:quant_eval_gtok_adaptation}, this simple replacement yields significant improvements across all metrics. Most notably, FID$\downarrow$ decreases by nearly 50\% (e.g., $35.25 \to 17.33$ on \textit{UFSC}) and Content Accuracy$\uparrow$ improves by approximately 9\% ($\sim87\% \to \sim96\%$). These substantial gains demonstrate that G-Tok's hybrid CNN-ViT architecture captures far richer structural and stylistic semantics than standard VQ-VAEs, serving as a robust plug-and-play enhancement for quantization-based FFG methods. \cref{fig:supple_vq_g_ufuc} illustrates that integrating G-Tok helps the model preserve coherent global structures for complex fonts and generate styles that better align with the target font, indicating a richer and more semantically stable global representation G-Tok than the original VQ-VAE.

\begin{table}[t]
\centering
\caption{Quantitative results of GAR-Font (M2/M4) across different description sources and formats under UFSC. M2/M4 denote inference with 2/4 reference glyphs plus text. All results outperform the corresponding non-text baselines in Table 2.}

\setlength{\tabcolsep}{3pt}
\renewcommand{\arraystretch}{1.05}
\resizebox{\linewidth}{!}{
\begin{tabular}{l | c | c c c c c c}
\toprule
 Method & Variants & RMSE$\downarrow$ & SSIM$\uparrow$ & LPIPS$\downarrow$ & FID$\downarrow$ & Acc(C)$\uparrow$ & Acc(S)$\uparrow$ \\
\midrule
\textbf{SmolVLM2-2.2B} & M2  & 0.2768 & 0.5819 & 0.1111 & 7.9185 & 0.9143 & 0.1384 \\
\textbf{(Fixed-Template)} & M4  & 0.2732 & 0.5878 & 0.1088 & 7.4247 & 0.9206 & 0.1747 \\
\midrule
\textbf{Qwen2.5-VL} & M2  & 0.2765 & 0.5808 & 0.1110 & 7.7404 & 0.9263 & 0.1329 \\
\textbf{(Free-Form)}& M4  & 0.2730 & 0.5890 & 0.1095 & 6.9813 & 0.9016 & 0.1702 \\
\bottomrule
\end{tabular}
}
\label{tab3:MultiModalBobust}
\end{table}

\subsection{Effect of the Number of NFA Glyphs}
\label{supplesubsec:nfa_number_ablation}
In Section 4.3, we adopt NFA-$8$ to maintain a strict few-shot adaptation setting. We further investigate the impact of using more adaptation glyphs by extending the setting to NFA-$32$ and NFA-$128$. 

As visualized in \cref{tab:ablation_nfa_number}, increasing the number of adaptation glyphs consistently improves the style modeling and perceptual quality. Style Accuracy (Acc(S)$\uparrow$) rises notably from NFA-$8$ to NFA-$128$ on both UFSC and UFUC, indicating that additional glyphs provide richer stylistic cues for capturing font-specific characteristics. Notably, on the \textit{Large} dataset, NFA-$128$ achieves the highest style accuracy (0.4457 on UFSC and 0.4625 on UFUC), substantially outperforming NFA-$8$. These results suggest that despite the NFA-$8$ setting adopted in the main paper already provides a strong and practical few-shot configuration, more NFA glyphs further refine the font transfer quality.

\begin{table*}[!t]
\centering
\caption{Quantitative evaluation of Multimodal FFG with full post-refinement (NFA and SE) on Unseen Fonts. All models listed are post-trained with NFA-$128$ and SE stages on the Large dataset.}

\setlength{\tabcolsep}{3pt}
\renewcommand{\arraystretch}{1.05}
\resizebox{0.95\textwidth}{!}{
\begin{tabular}{l | c c c c c c | c c c c c c }
\toprule
Method & \multicolumn{6}{c|}{Unseen Fonts Seen Characters (UFSC)} & \multicolumn{6}{c}{Unseen Fonts Unseen Characters (UFUC)} \\
\cmidrule(lr){2-7} \cmidrule(lr){8-13}
 & RMSE$\downarrow$ & SSIM$\uparrow$ & LPIPS$\downarrow$ & FID$\downarrow$ & Acc(C)$\uparrow$ & Acc(S)$\uparrow$
 & RMSE$\downarrow$ & SSIM$\uparrow$ & LPIPS$\downarrow$ & FID$\downarrow$ & Acc(C)$\uparrow$ & Acc(S)$\uparrow$ \\
\hline
$n_\text{ref}=2$ & 0.2501 & 0.6438 & 0.0888 & 10.1464 & \textbf{0.9851} & 0.3314 & 0.2701 & 0.5987 & 0.1069 & 9.4294 & 0.8720 & 0.3433 \\
$n_\text{ref}=4$ & 0.2437 & 0.6552 & 0.0848 & 9.5960  & 0.9839 & 0.3711 & 0.2692 & 0.6007 & 0.1049 & 9.0877 & 0.8828 & 0.3835 \\
$n_\text{ref}=8$ & 0.2398 & 0.6627 & 0.0831 & \textbf{8.3751}  & 0.9776 & 0.4154 & \textbf{0.2508} & \textbf{0.6377} & \textbf{0.0908} & \textbf{7.2034} & \textbf{0.9068} & 0.4183 \\
\Xhline{1.2pt}
GAR-Font($M_2$) & 0.2361 & 0.6707 & 0.0799 & 8.8867  & 0.9817 & 0.4508 & 0.2527 & 0.6352 & 0.0909 & 8.1444 & 0.9029 & 0.4247 \\
GAR-Font($M_4$) & \textbf{0.2358} & \textbf{0.6712} & \textbf{0.0796} & 8.8073  & 0.9800 & \textbf{0.4566} & 0.2524 & 0.6353 & \textbf{0.0908} & 8.0699 & 0.9023 & \textbf{0.4391} \\
\bottomrule
\end{tabular}
}
\label{tab:quant_eval_ref_mm}
\end{table*}
\begin{figure*}[!t]
    \centering
    \begin{minipage}{0.495\linewidth}
        \centering
        \includegraphics[width=\linewidth]{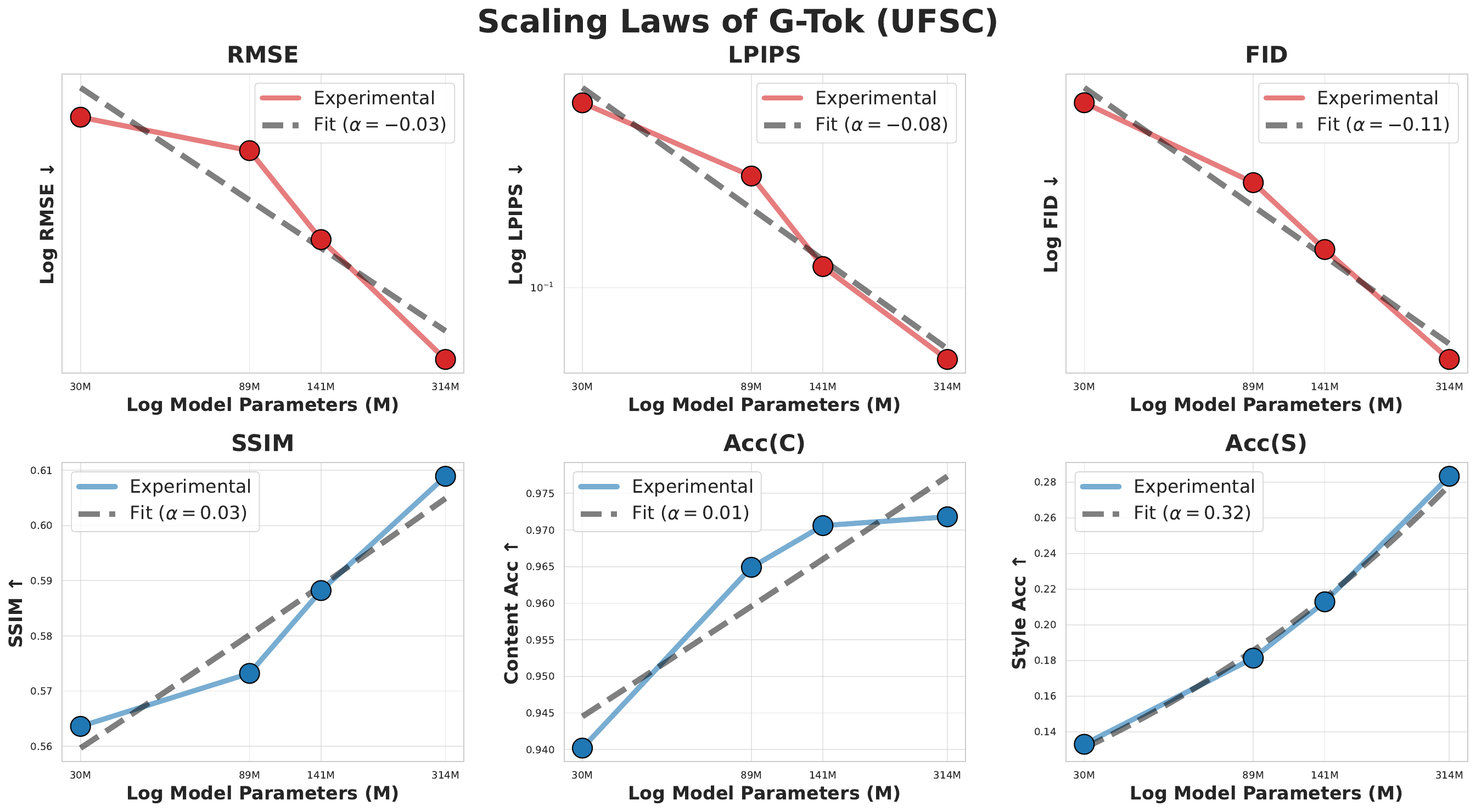}
    \end{minipage}
    \hfill
    \begin{minipage}{0.495\linewidth}
        \centering
        \includegraphics[width=\linewidth]{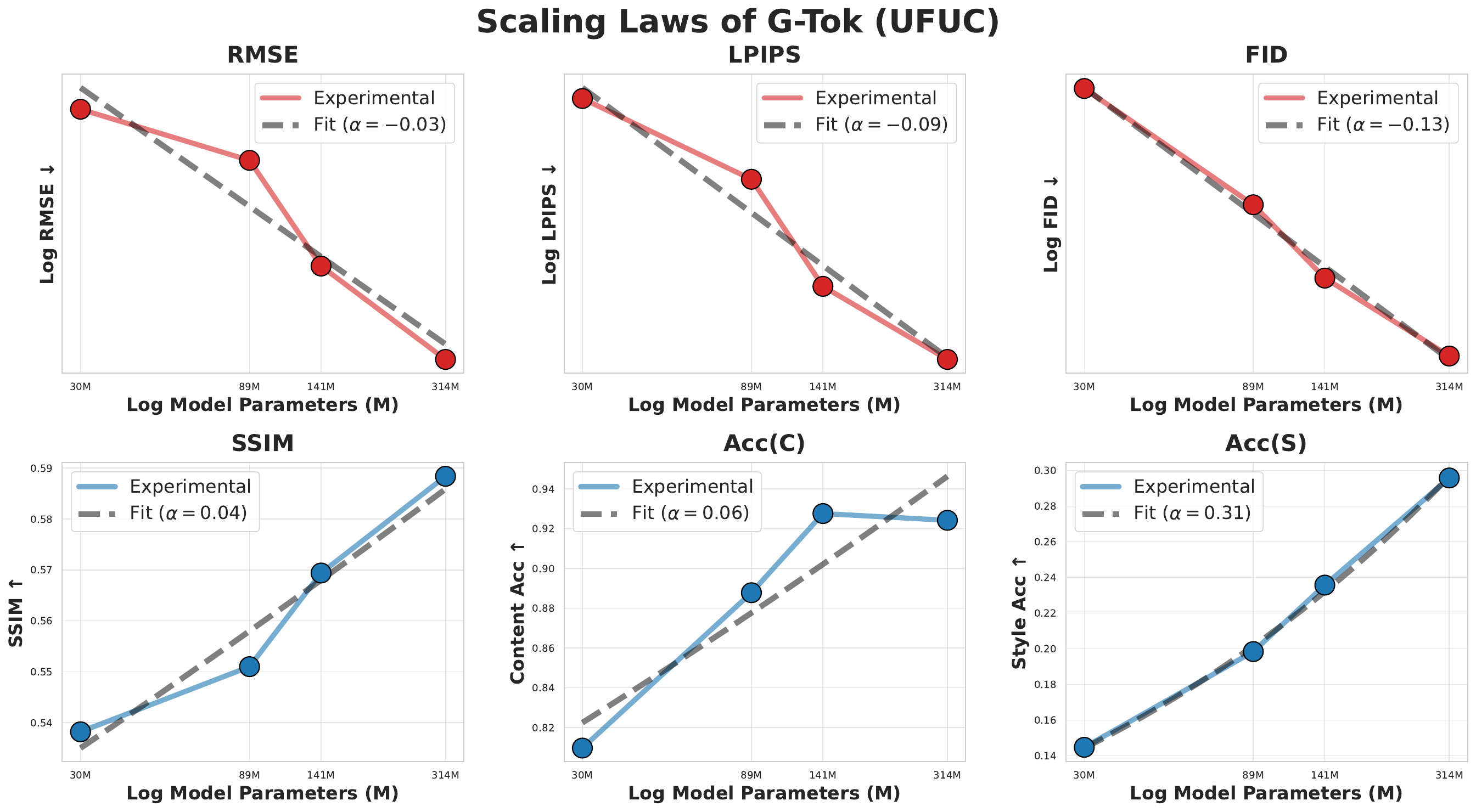}
    \end{minipage}

    \caption{Scaling laws of the GAR-Font $(I_8)$ generator with NFA-$128$ and SE refinement. The plots show performance metrics across model sizes (30M, 89M, 141M, 314M) on Unseen Fonts Seen Characters (\textit{UFSC}) and Unseen Fonts Unseen Characters (\textit{UFUC}). The dashed lines represent power-law fits, highlighting the predictable improvements in both perceptual quality and style generalization.}

    \label{fig:scaling_laws}
\end{figure*}

\begin{table*}[!t]
\centering
\caption{Quantitative evaluation of G-Tok's architecture on Unseen Fonts. All models listed are pre-trained on the Small dataset.}

\setlength{\tabcolsep}{3pt}
\renewcommand{\arraystretch}{1.05}
\resizebox{0.95\textwidth}{!}{
\begin{tabular}{l | c c c c c c | c c c c c c }
\toprule
Method & \multicolumn{6}{c|}{Unseen Fonts Seen Characters (UFSC)} & \multicolumn{6}{c}{Unseen Fonts Unseen Characters (UFUC)} \\
\cmidrule(lr){2-7} \cmidrule(lr){8-13}
 & RMSE$\downarrow$ & SSIM$\uparrow$ & LPIPS$\downarrow$ & FID$\downarrow$ & Acc(C)$\uparrow$ & Acc(S)$\uparrow$ 
 & RMSE$\downarrow$ & SSIM$\uparrow$ & LPIPS$\downarrow$ & FID$\downarrow$ & Acc(C)$\uparrow$ & Acc(S)$\uparrow$ \\
\hline
CNN & 0.3212 & 0.4836 & 0.1442 & 9.5071 & 0.9051 & 0.0235 & 0.3447 & 0.4350 & 0.1728 & 10.5239 & 0.6722 & 0.0221 \\
CNN+Non-Causal ViT & 0.3183 & 0.4919 & 0.1458 & \textbf{7.9101} & 0.9268 & 0.0402 & 0.3271 & 0.4745 & 0.1562 & 8.7504 & 0.8019 & 0.0436 \\
CNN+Causal ViT & \textbf{0.3080} & \textbf{0.5052} & \textbf{0.1313} & 7.9484 & \textbf{0.9408} & \textbf{0.0802} & \textbf{0.3142} & \textbf{0.4932} & \textbf{0.1421} & \textbf{8.4841} & \textbf{0.8993} & \textbf{0.0796} \\
\bottomrule
\end{tabular}
}
\label{tab:ablation_gtok_generator}
\end{table*}

\subsection{Robustness Across Textual Description Sources}
\label{supplesubsec:robust_text}

In Section 4.4, multimodal experiments are conducted using the generated descriptions of Qwen2.5-VL with the fixed-form template in Section C.3. To validate the robustness of our approach, we further evaluate GAR-Font using two types of descriptions: a free-form prompt for Qwen2.5-VL and the fixed-template prompt for SmolVLM2-2.2B-Instruct ~\cite{smolvlm}, each generated using only \textbf{8} reference glyphs. 

\cref{tab3:MultiModalBobust} shows the multimodal gains are consistent across description sources and formats. GAR-Font$(M_2/M_4)$ outperforms corresponding non-text baselines, demonstrating that our vision-language adaptation is robust to variations in prompt style and source model.

\subsection{Post-Refinement of Multimodal FFG}
\label{supplesubsec:mm_post_refinement}
In Section 4.4, we demonstrated the efficacy of GAR-Font$(M_2/M_4)$ on multimodal FFG, but limited to pretraining stage. To fully assess the potential of our lightweight vision-language adaptation, we extend the evaluation to the complete pipeline. We apply our NFA-$128$ and SE stages to both the vision-only baselines and our multimodal variants. All models are trained on the Large (\textit{L}) dataset.

As visualized in \cref{tab:quant_eval_ref_mm}, the inclusion of textual descriptions significantly enhances the effectiveness of the post-refinement stage. Unlike the pre-training phase where multimodal models showed a slight dip in style accuracy (Acc(S)$\uparrow$), the fully refined GAR-Font($M_2$) and GAR-Font($M_4$) exhibit a substantial lead in Acc(S)$\uparrow$ compared to their vision-only counterparts ($n_\text{ref}=2$ and $n_\text{ref}=4$). Notably, \textbf{GAR-Font($M_4$) outperforms the 8-reference vision-only baseline ($n_\text{ref}=8$)} across most key metrics, including RMSE$\downarrow$, SSIM$\uparrow$, LPIPS$\downarrow$, and Style Accuracy$\uparrow$ (0.4566 vs. 0.4154 on \textit{UFSC}).

\subsection{Scaling Laws of GAR-Font\texorpdfstring{$(I_8)$}{(I8)}}
\label{supplesubsec:scaling_law}
We evaluate the scalability of GAR-Font$(I_8)$ with NFA-$128$ and SE on the small dataset by training models from 30M to 314M parameters and measuring performance across standard quantitative metrics. Following established scaling-law formulations, we model the relationship between model size $N$ and loss metric $L$ using a power law $L(N)\propto N^{-\alpha}$, and analyze trends in log–log space, where an ideal scaling law appears linear and the slope $\alpha$ reflects scaling efficiency.

As shown in \cref{fig:scaling_laws}, the enhanced GAR-Font models closely follow these power-law predictions, exhibiting smooth, monotonic improvements across all metrics. Loss-based metrics (FID$\downarrow$, LPIPS$\downarrow$, RMSE$\downarrow$) scale linearly with negative slopes, with FID$\downarrow$ showing a pronounced gain, indicating that larger models continue to yield substantial perceptual improvements without saturation. Accuracy metrics display complementary behavior: Content Accuracy (Acc(C)$\uparrow$) saturates early due to task simplicity, whereas Style Accuracy (Acc(S)$\uparrow$) benefits most from increased capacity. This steep scaling trend highlights that NFA and SE effectively exploit larger parameter budgets to capture and generalize complex stylistic attributes, underscoring the central role of scale in high-fidelity font generation.

\section{Additional Ablative Studies}
\label{supplesec:sup_Ablate}
\subsection{On G-Tok's hybrid Architecture}
\label{supplesubsec:AblateVQG}
\begin{figure}[t]
  \centering
  \includegraphics[width=0.95\linewidth]{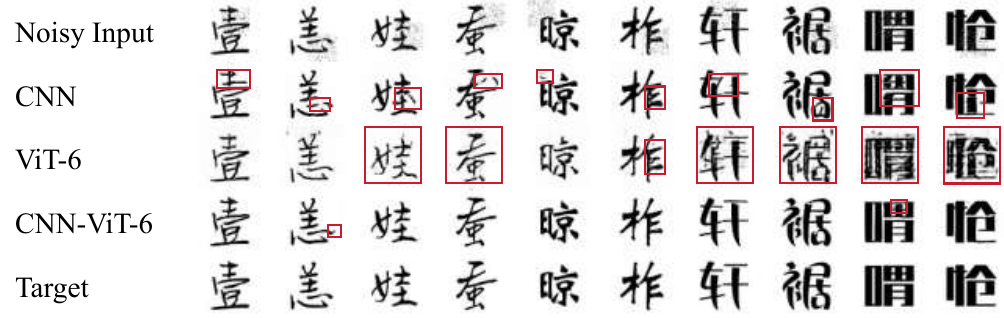} 

  \caption{Reconstruction Robustness under localized Gaussian noise ($\sigma = 0.2$, 20\% area). \textcolor{ex2figure_red}{\fbox{\rule{0pt}{4pt}\rule{4pt}{0pt}}} marks structural errors. G-Tok (CNN-ViT-6) preserves structure and style despite heavy corruption, while non-hybrid tokenizers exhibit unstable reconstructions.}

  \label{fig:supple_abl1_rec}
\end{figure}

To further illustrate the robustness of our hybrid CNN–ViT tokenizer, we provide complete visualizations of the \textbf{Reconstruction Robustness} experiment, where glyphs are corrupted with localized Gaussian noise ($\sigma = 0.2$, affecting 20\% area).  
The qualitative results in \cref{fig:supple_abl1_rec} demonstrate that G-Tok robustly recovers structural layout and stylistic traits even under severe perturbations, while non-hybrid alternatives fail to reconstruct consistent structure. 


\subsection{On G-Tok's Global and Causal Modeling}
\label{supplesubsec:AblateGTok}
We present full ablation results for the global and causal modeling components of G-Tok.  
\cref{tab:ablation_gtok_generator} reports the complete quantitative comparison on the \textit{UFSC/UFUC}. As discussed in Section 4.5.2, adding global self-attention (CNN + Non-causal ViT) significantly outperforms the CNN-only baseline, while the causal ViT further improves sequential modeling and yields the best overall performance.

\cref{fig: supple_abl2_all} provides qualitative comparisons on (\textit{UFSC}, Small) and (\textit{UFUC}, Small). The AR Generator implemented with a CNN-only tokenizer often exhibits style mismatches and inconsistent strokes. Introducing ViT modules into the tokenizer enhances its ability to perceive and capture global stylistic context, leading to more coherent font generation. The AR variant with full G-Tok (CNN + Causal ViT) achieves the most robust performance, showing visible improvements in stylistic and  structural fidelity.

\subsection{On AR Generator's Soft-Decoding}
\label{supplesubsec:AblateSoftDecoding}
\begin{figure*}[tp] 
    \centering
    
    \begin{subfigure}{0.9\linewidth} 
        \centering
        \includegraphics[width=\linewidth]{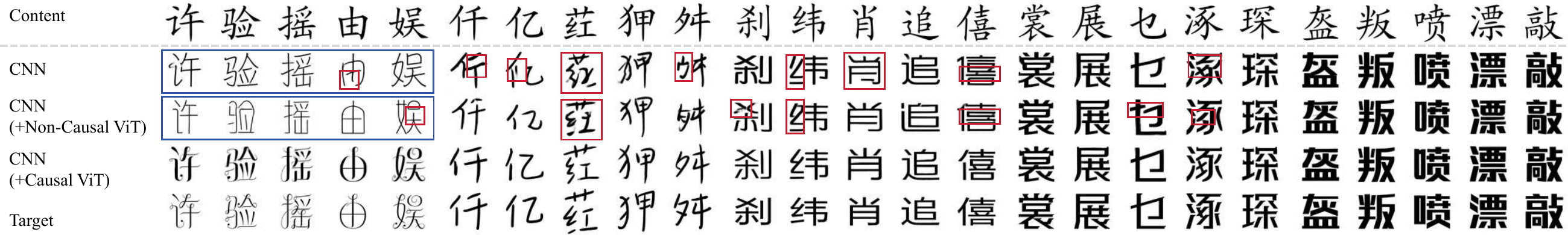}
        \caption{\textit{UFSC}, Small dataset}
        \label{fig: supple_abl2_ufsc_small}
    \end{subfigure}
    \\ 
    \begin{subfigure}{0.9\linewidth}
        \centering
        \includegraphics[width=\linewidth]{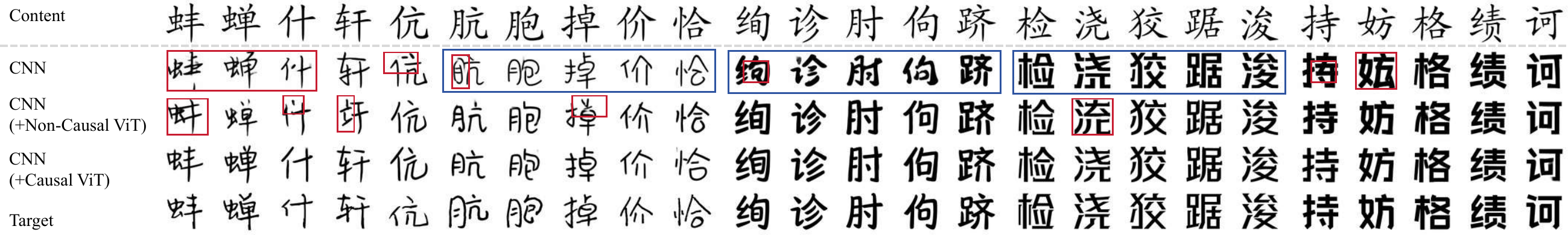}
        \caption{\textit{UFUC}, Small dataset}
        \label{fig: supple_abl2_ufuc_small}
    \end{subfigure}

    \caption{Qualitative results on G-Tok's Global and Causal Modeling under \textit{UFSC} and \textit{UFUC} protocols
        (Small dataset). 
        \textcolor{ex1figure_red}{\fbox{\rule{0pt}{4pt}\rule{4pt}{0pt}}} /
        \textcolor{ex1figure_blue}{\fbox{\rule{0pt}{4pt}\rule{4pt}{0pt}}}
        indicate structural errors and style mismatches.}
    \label{fig: supple_abl2_all}

    \begin{subfigure}{0.9\linewidth}
        \centering
        \includegraphics[width=\linewidth]{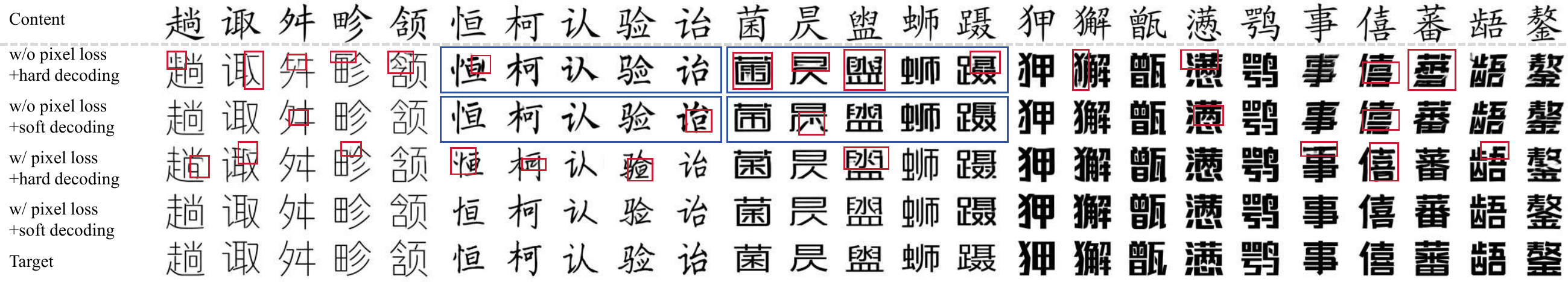}
        \caption{\textit{UFSC}, Small dataset}
        \label{fig: supple_abl3_ufsc_small}
    \end{subfigure}
    \\
    \begin{subfigure}{0.9\linewidth}
        \centering
        \includegraphics[width=\linewidth]{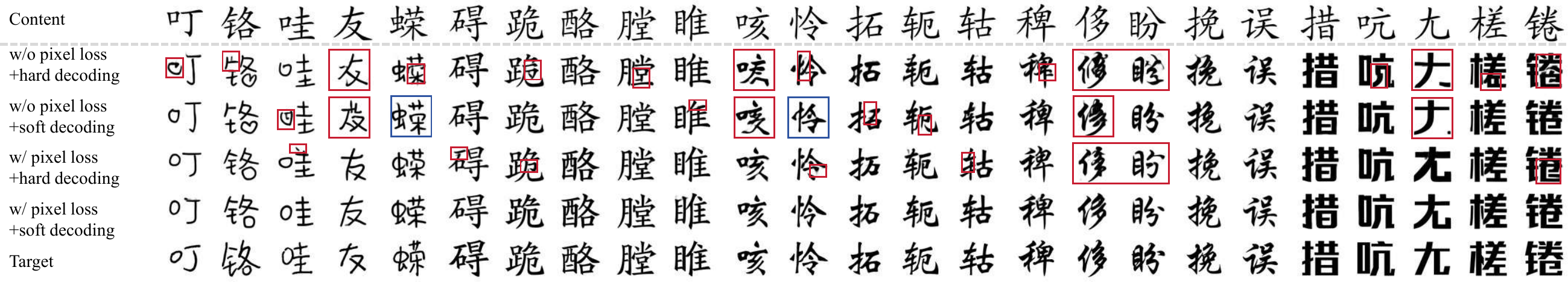}
        \caption{\textit{UFUC}, Small dataset}
        \label{fig: supple_abl3_ufuc_small}
    \end{subfigure}

    \caption{Qualitative results on AR Generator's Soft-decoding under \textit{UFSC} and \textit{UFUC} protocols
        (Small dataset).
        \textcolor{ex1figure_red}{\fbox{\rule{0pt}{4pt}\rule{4pt}{0pt}}} /
        \textcolor{ex1figure_blue}{\fbox{\rule{0pt}{4pt}\rule{4pt}{0pt}}}
        indicate structural errors and style mismatches.}
    \label{fig: supple_abl3_all}

    \begin{subfigure}{0.9\linewidth}
        \centering
        \includegraphics[width=\linewidth]{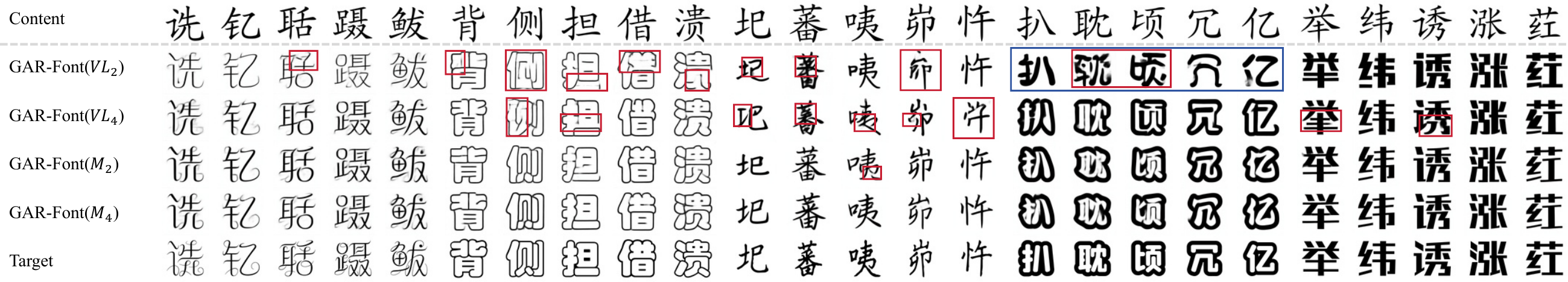}
        \caption{\textit{UFSC}, Large dataset}
        \label{fig: supple_abl4_ufsc_large}
    \end{subfigure}
    \\
    \begin{subfigure}{0.9\linewidth}
        \centering
        \includegraphics[width=\linewidth]{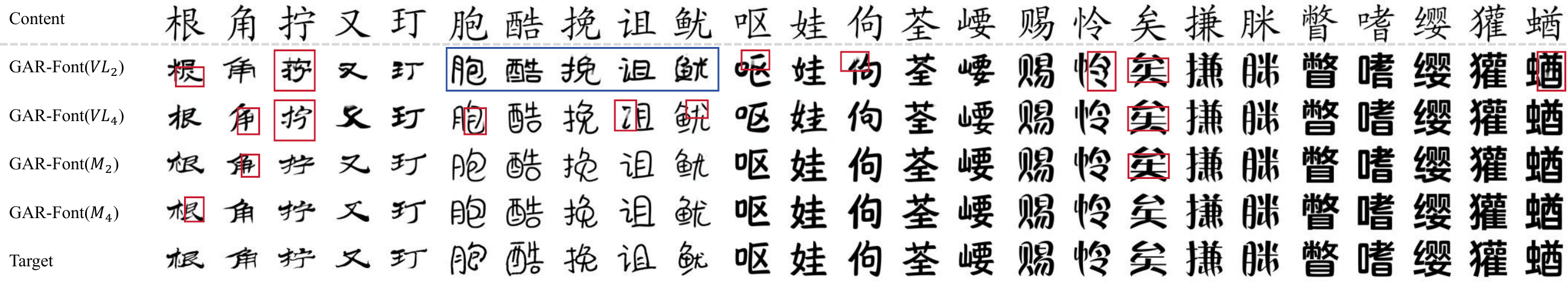}
        \caption{\textit{UFUC}, Large dataset}
        \label{fig: supple_abl4_ufuc_large}
    \end{subfigure}

    \caption{
        Qualitative results on Multimodal Style Encoder's Adaptation under \textit{UFSC} and \textit{UFUC} protocols
        (Large dataset).
        \textcolor{ex1figure_red}{\fbox{\rule{0pt}{4pt}\rule{4pt}{0pt}}} /
        \textcolor{ex1figure_blue}{\fbox{\rule{0pt}{4pt}\rule{4pt}{0pt}}}
        indicate structural errors and style mismatches.
    }
    \label{fig: supple_abl4_all}
    
\end{figure*}

We provide full visualizations to assess the impact of pixel-level supervision and the soft-decoding strategy. As shown in \cref{fig: supple_abl3_all} under both (\textit{UFSC}, Small) and (\textit{UFUC}, Small), pixel-level supervision enhances structural accuracy, while soft decoding yields smoother, more continuous strokes and reduces broken segments and visual artifacts.

\begin{figure*}[tp]
    \centering

    \begin{subfigure}{0.94\linewidth}
        \centering
        \includegraphics[width=\linewidth]{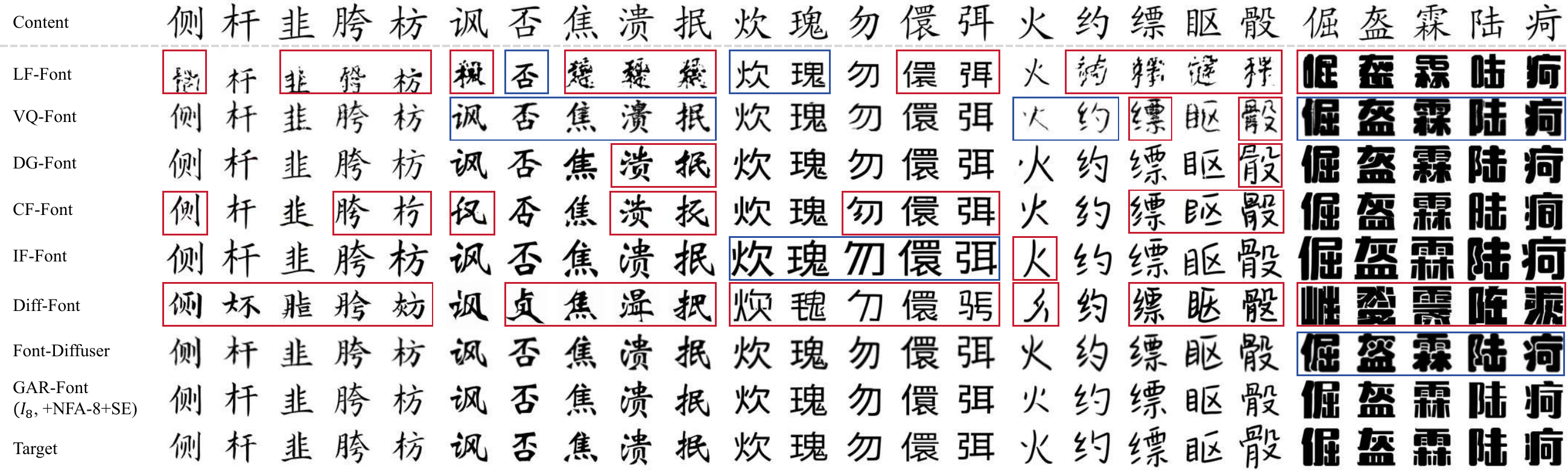}

        \caption{\textit{UFSC}, Small dataset}
        \label{fig: supple_main1_ufsc_small}

    \end{subfigure}

    \begin{subfigure}{0.94\linewidth}
        \centering
        \includegraphics[width=\linewidth]{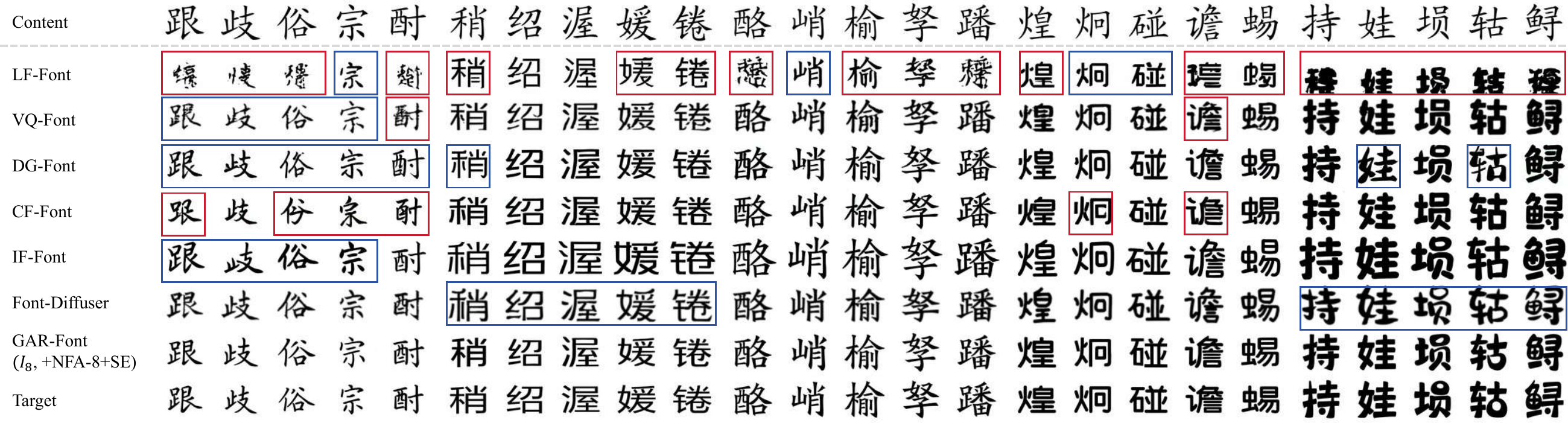}

        \caption{\textit{UFUC}, Small dataset}

        \label{fig: supple_main1_ufuc_small}
    \end{subfigure}

    \begin{subfigure}{0.94\linewidth}
        \centering
        \includegraphics[width=\linewidth]{Camera_Ready_IMGS/Vision_Only_NFA8_FFG/table_Large_UFSC_Vision_Only_FFG.pdf}

        \caption{\textit{UFSC}, Large dataset}

        \label{fig: supple_main1_ufsc_large}
    \end{subfigure}

    \begin{subfigure}{0.94\linewidth}
        \centering
        \includegraphics[width=\linewidth]{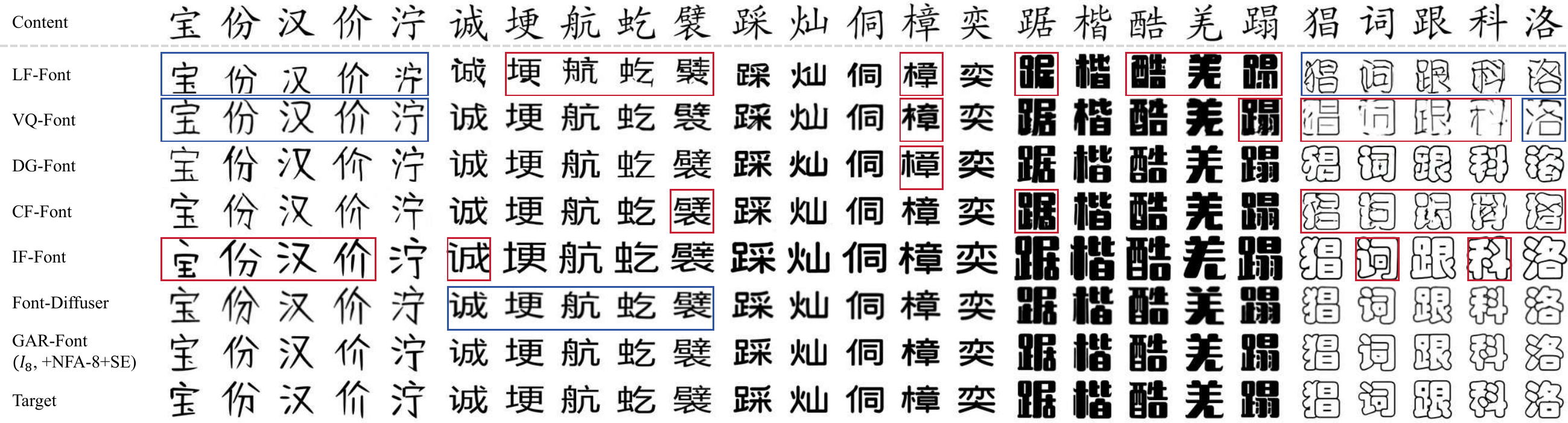}

        \caption{\textit{UFUC}, Large dataset}

        \label{fig: supple_main1_ufuc_large}
    \end{subfigure}

    \caption{
        Qualitative results on vision-only FFG across \textit{UFSC/UFUC} protocols and Small/Large datasets.
        \textcolor{ex1figure_red}{\fbox{\rule{0pt}{4pt}\rule{4pt}{0pt}}} /
        \textcolor{ex1figure_blue}{\fbox{\rule{0pt}{4pt}\rule{4pt}{0pt}}}
        indicate structural errors and style mismatches.
    }

    \label{fig: supple_main1_all}
\end{figure*}

\subsection{On Multimodal Style Encoder's Adaptation}
\label{supplesubsec:AblateMultimodalMethod}
We compare our decoupled multimodal training paradigm against joint training of the multimodal style encoder. While quantitative results are provided in Section 4.5.4, we present the full set of qualitative comparisons here.

\cref{fig: supple_abl4_all} presents visual comparisons on (\textit{UFSC/UFUC}, Large). The results reveal that GAR-Font($M_2$/$M_4$), trained with the decoupled training scheme, generate glyphs whose font styles more closely align with the target compared to the jointly trained GAR-Font($VL_2$/$VL_4$). They also demonstrate better character-structure accuracy. The decoupled training strategy enables the model to fully leverage the visual encoder’s representational capacity, thereby preserving fine-grained style features and structural priors that may be harder to retain under joint optimization.

\begin{figure*}[tp] 
    \centering

    
    \begin{subfigure}{0.94\linewidth} 
        \centering
        \includegraphics[width=\linewidth]{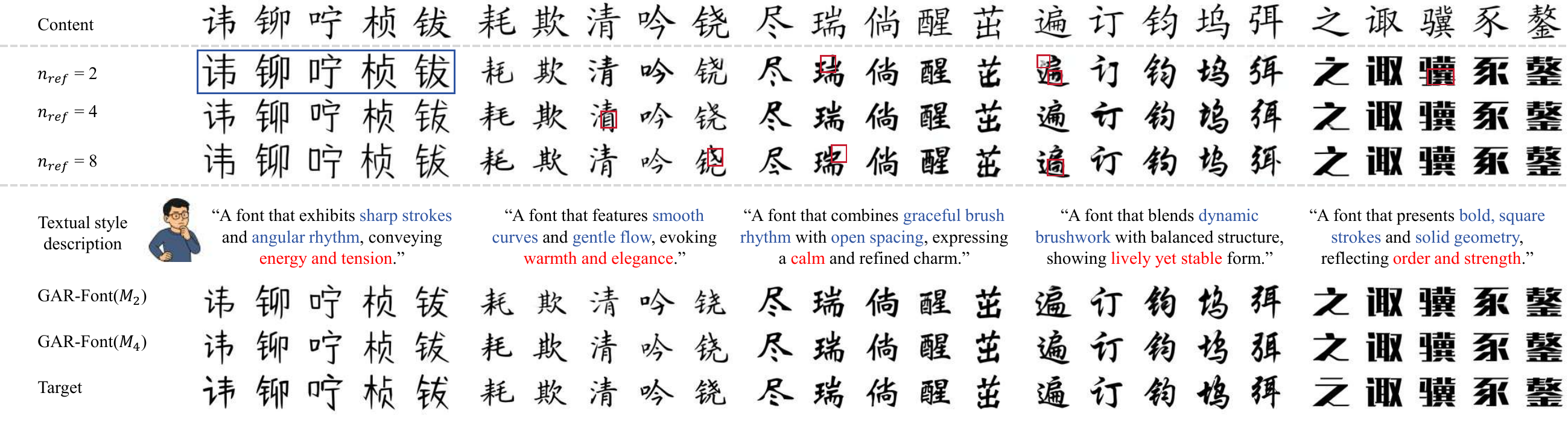}
        \caption{\textit{UFSC}, Large dataset}
        \label{fig: supple_main2_ufsc_large_pre}
    \end{subfigure}
    \\ 

    \begin{subfigure}{0.94\linewidth}
        \centering
        \includegraphics[width=\linewidth]{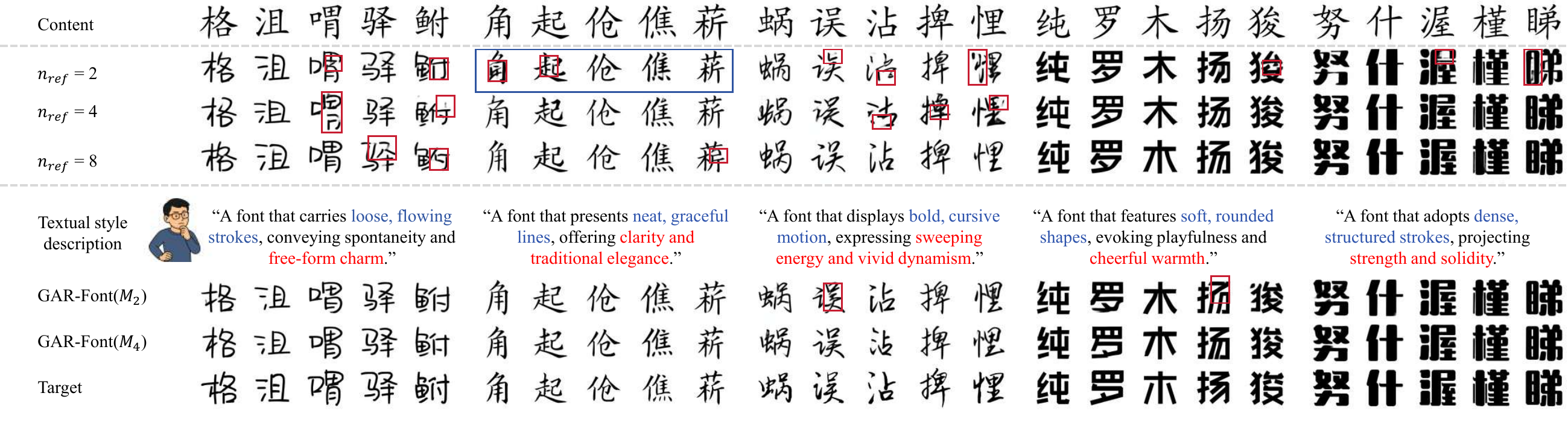}
        \caption{\textit{UFUC}, Large dataset}
        \label{fig: supple_main2_ufuc_large_pre}
    \end{subfigure}

    \caption{
        Qualitative results of Pre-Train multimodal FFG under \textit{UFSC} and \textit{UFUC} protocols
        (Large dataset). 
        \textcolor{ex2figure_red}{\fbox{\rule{0pt}{4pt}\rule{4pt}{0pt}}} denotes local slight structural mistakes, 
        and \textcolor{ex2figure_blue}{\fbox{\rule{0pt}{4pt}\rule{4pt}{0pt}}} marks stylistic drift.
    }
    \label{fig: supple_main2_large_pre_all}


    \begin{subfigure}{0.94\linewidth}
        \centering
        \includegraphics[width=\linewidth]{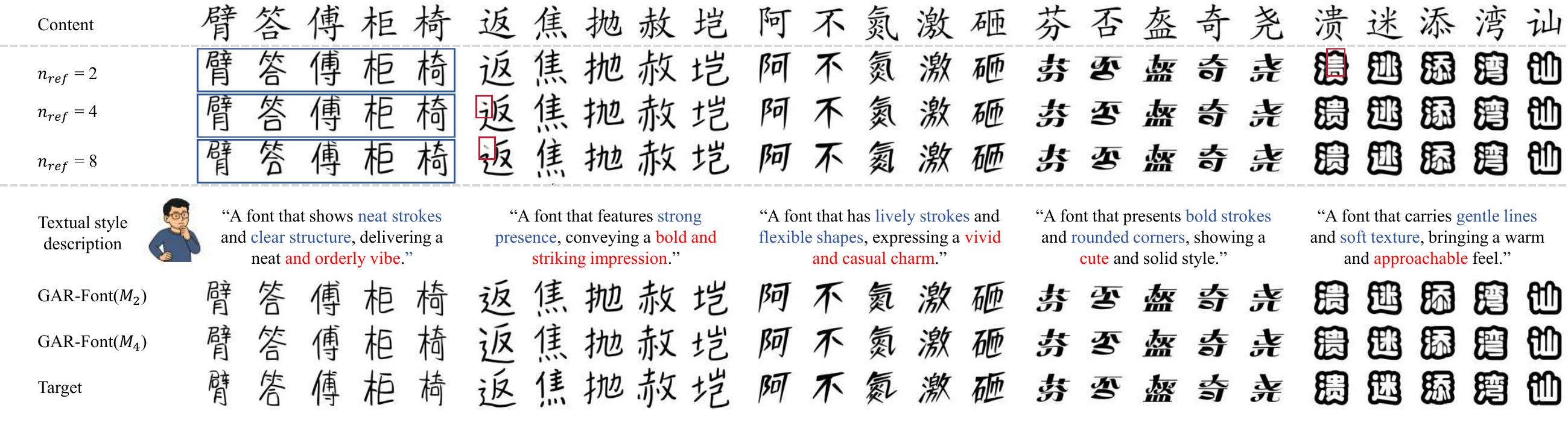}
        \caption{\textit{UFSC}, Large dataset}
        \label{fig: supple_main2_ufsc_large_post}
    \end{subfigure}

    \begin{subfigure}{0.94\linewidth}
        \centering
        \includegraphics[width=\linewidth]{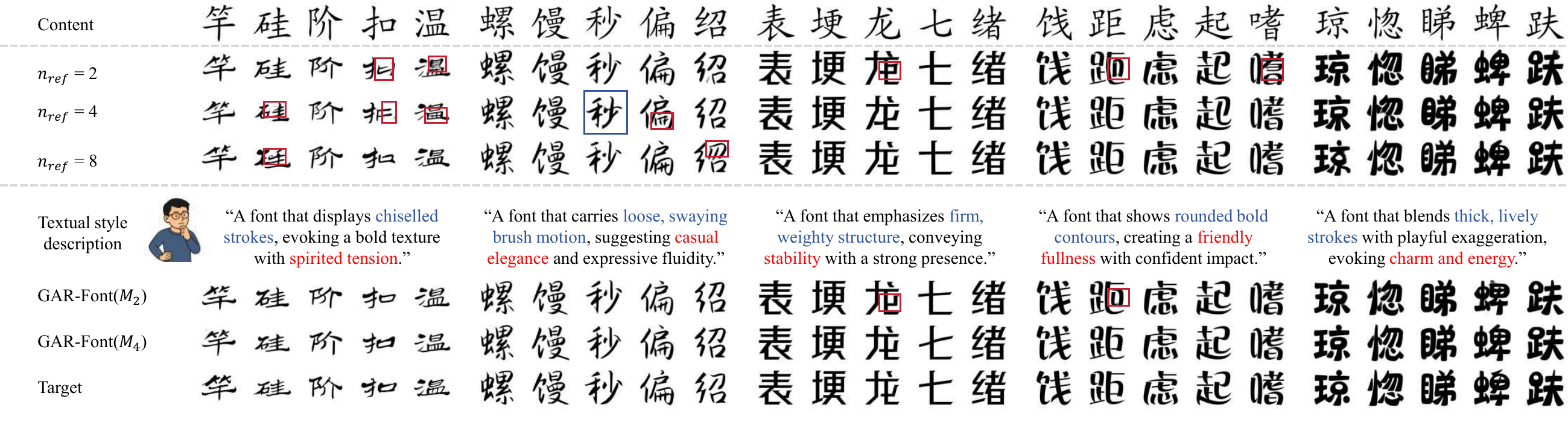}
        \caption{\textit{UFUC}, Large dataset}
        \label{fig: supple_main2_ufuc_large_post}
    \end{subfigure}

    \caption{
        Qualitative results of Post-Refine multimodal FFG under \textit{UFSC} and \textit{UFUC} protocols
        (Large dataset). 
        \textcolor{ex2figure_red}{\fbox{\rule{0pt}{4pt}\rule{4pt}{0pt}}} denotes local slight structural mistakes, 
        and \textcolor{ex2figure_blue}{\fbox{\rule{0pt}{4pt}\rule{4pt}{0pt}}} marks stylistic drift.
    }
    \label{fig: supple_main2_large_post}

\end{figure*}

\begin{figure*}[!tbp]  
    \centering
    \begin{subfigure}{\linewidth}
        \centering
        \includegraphics[width=\linewidth]{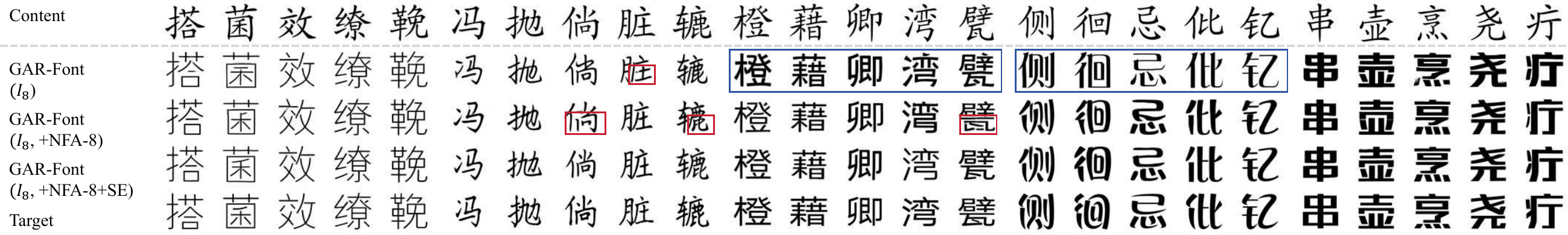}

        \caption{\textit{UFSC}, Small dataset}
        \label{fig: supple_postrefine_ufsc_small}

    \end{subfigure}

    \begin{subfigure}{\linewidth}
        \centering
        \includegraphics[width=\linewidth]{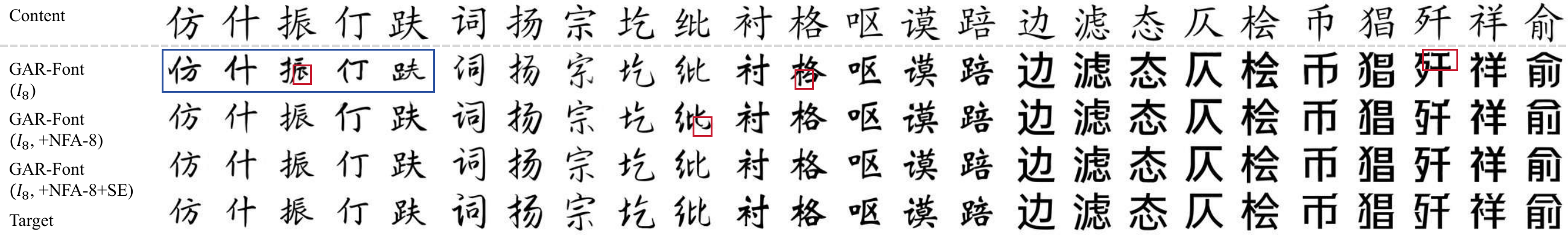}

        \caption{\textit{UFUC}, Small dataset}
        \label{fig: supple_postrefine_ufuc_small}

    \end{subfigure}

    \begin{subfigure}{\linewidth}
        \centering
        \includegraphics[width=\linewidth]{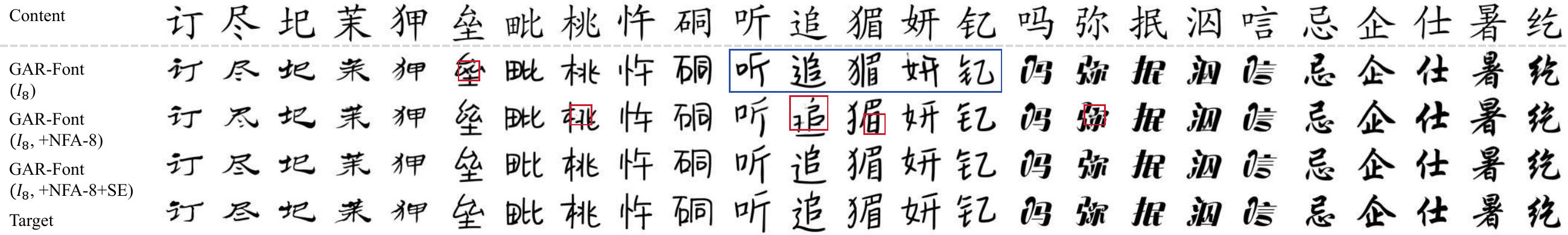}

        \caption{\textit{UFSC}, Large dataset}
        \label{fig: supple_postrefine_ufsc_large}

    \end{subfigure}

    \begin{subfigure}{\linewidth}
        \centering
        \includegraphics[width=\linewidth]{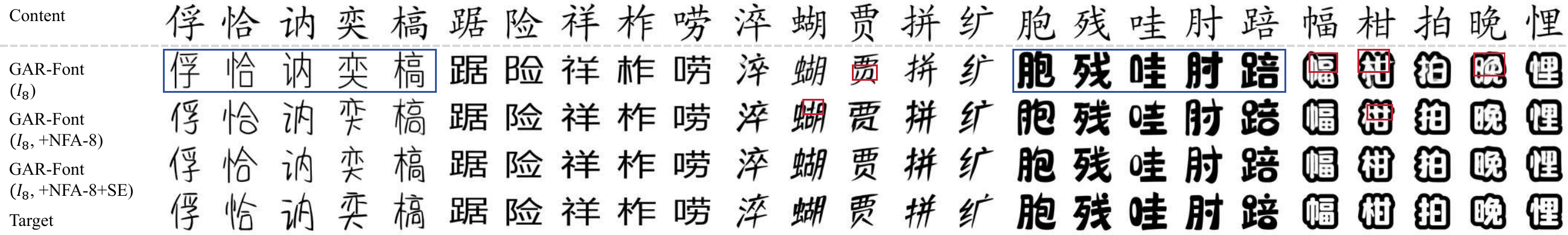}

        \caption{\textit{UFUC}, Large dataset}
        \label{fig: supple_postrefine_ufuc_large}

    \end{subfigure}

    \caption{
        Qualitative results on Post-refinement across \textit{UFSC/UFUC} protocols and Small/Large datasets.
        \textcolor{ex2figure_red}{\fbox{\rule{0pt}{4pt}\rule{4pt}{0pt}}} denotes local slight structural mistakes, 
        and \textcolor{ex2figure_blue}{\fbox{\rule{0pt}{4pt}\rule{4pt}{0pt}}} marks stylistic drift.
    }
    \label{fig: supple_pre_post_refinement}

    \includegraphics[width=\linewidth]{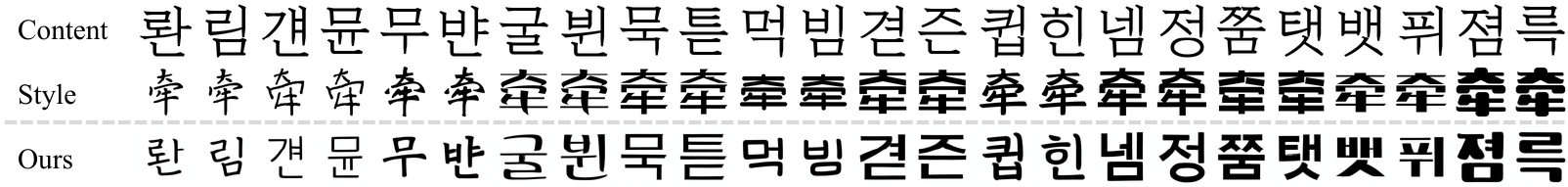}

    \caption{Qualitative results of cross-language font synthesis using GAR-Font($I_8$) trained on the Large dataset, demonstrating the strong generalization ability.}
    \label{fig:Korean}

    \includegraphics[width=\linewidth]{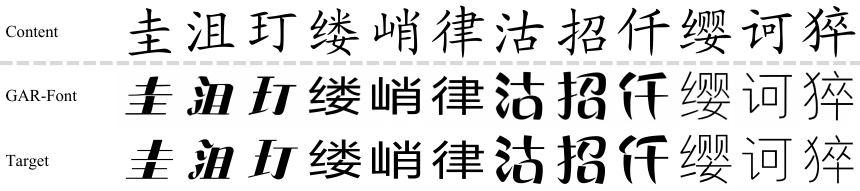}

    \caption{Qualitative results of $128\times128$ font generation using GAR-Font($I_8$, +NFA-$8$+SE) trained on the Large dataset, showing the scalability of GAR-Font to higher resolutions while preserving style faithfulness and structural fidelity.}
    \label{fig:supple_highres_128}

\end{figure*}

\begin{figure*}[t]
  \centering
  \includegraphics[width=0.95\linewidth]{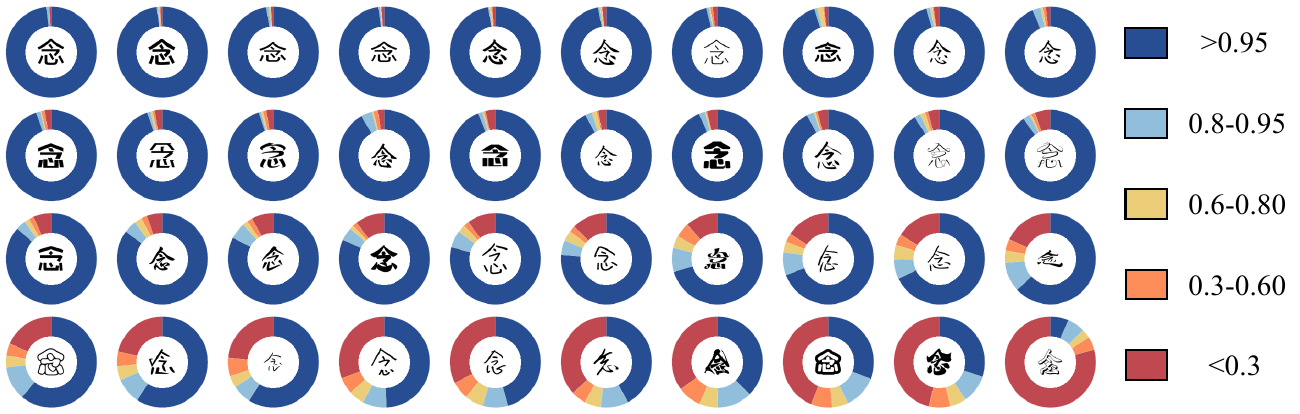}

  \caption{Content confidence distribution of GAR-Font generated characters across different font styles. Each pie chart corresponds to a specific font, indicated by the central character. The color segments represent the proportion of samples falling into different content confidence ranges, highlighting that more complex styles tend to have lower content confidence.}

  \label{fig:supple_analysis}
\end{figure*}

\begin{figure}[t]
  \centering
  \includegraphics[width=0.95\linewidth]{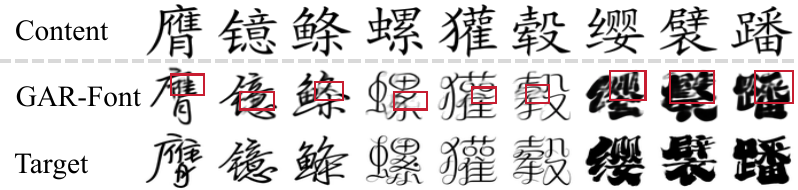}

  \caption{Failure cases. \textcolor{ex2figure_red}{\fbox{\rule{0pt}{4pt}\rule{4pt}{0pt}}} highlights regions with dense details where GAR-Font tends to produce distorted strokes.}

  \label{fig:supple_failure_cases}
\end{figure}

\section{Visualization Results}
\label{supplesec:sup_vis_results}
\subsection{Comparison on Few-shot Font Generation}
We provide complete visualizations for the experiment in Section 4.3. In \cref{fig: supple_main1_all}, we show the full qualitative comparisons of visual-only FFG models trained on Small and Large datasets, evaluated under both \textit{UFSC} and \textit{UFUC} protocols. These results indicate that methods such as LF--Font, VQ-Font, DG-Font, CF-Font and Diff-Font often fail to preserve structural fidelity in intricate fonts. IF-Font tends to produce incomplete characters, while Font-Diffuser generates with inaccurate stroke widths. In contrast, GAR-Font($I_8$,+NFA-$8$+SE) achieves the best style fidelity while maintaining structural consistency, effectively capturing fine stroke details of the target fonts.

\subsection{Efficient Vision-Language Adaptation}

\subsubsection{Pretrain}

We provide full qualitative results complementing the experiment in Section 4.4. 
In \cref{fig: supple_main2_large_pre_all}, we show multimodal FFG comparisons under both \textit{UFSC} and \textit{UFUC} settings on the Large dataset, illustrating the improvements introduced by incorporating textual style descriptions in GAR-Font($M_2$) and GAR-Font($M_4$) compared with their vision-only counterparts. With textual style guidance, GAR-Font($M_2$) and GAR-Font($M_4$) better align with the target style, generating glyphs with strokes closely matching the target and improved structural fidelity.

\subsubsection{Post-Refinement}
To further assess the potential of our efficient vision–language adaptation, we apply the complete post-refinement pipeline (NFA-$128$ and SE) to GAR-Font($M_2$) and GAR-Font($M_4$). 
\cref{fig: supple_main2_large_post} presents qualitative results under \textit{UFSC} and \textit{UFUC} on the Large dataset. Applying NFA and SE post-refinement significantly improves both structural and style fidelity for all models. Textual guidance further enables GAR-Font($M_2$) and GAR-Font($M_4$) to more accurately capture the target style, yielding glyphs with improved style fidelity.

\subsection{Effect of Post-Refinement}
To further analyze the effect of the post-refinement, we provide visual comparisons. As shown in \cref{fig: supple_pre_post_refinement}, the pretrained GAR-Font($I_8$) already produces characters with generally correct structures and styles, though minor font inconsistency exists. Applying NFA significantly improves style fidelity but may introduce slight distortions in fine strokes. The SE stage preserves style fidelity while further enhancing visual clarity and the accuracy of stroke details especially in complex fonts.

\subsection{Cross-Language Font Synthesis}
\label{ssec:cross_language}
To evaluate the generalizability of GAR-Font, we conduct a cross-language experiment in which the model synthesizes Korean characters using styles learned from Chinese fonts. As shown in \cref{fig:Korean}, GAR-Font accurately generates Korean characters while preserving the reference font style, demonstrating the effective generalization of our method.

\subsection{High-Resolution Font Generation}
\label{ssec:qual_highres}
To demonstrate the scalability of GAR-Font beyond the $64\times64$ resolution adopted in our main experiments, we modify the CNN encoder within G-Tok to discretize a $128\times128$ glyph into $64$ tokens, corresponding to a downsample ratio of $16$. As illustrated in \cref{fig:supple_highres_128}, GAR-Font maintains both style faithfulness and structural fidelity at this increased resolution, demonstrating its potential in high-resolution font generation tasks.

\subsection{More GAR-Font Generation Examples}
\begin{figure*}
    \centering
    \includegraphics[width=0.95\linewidth]{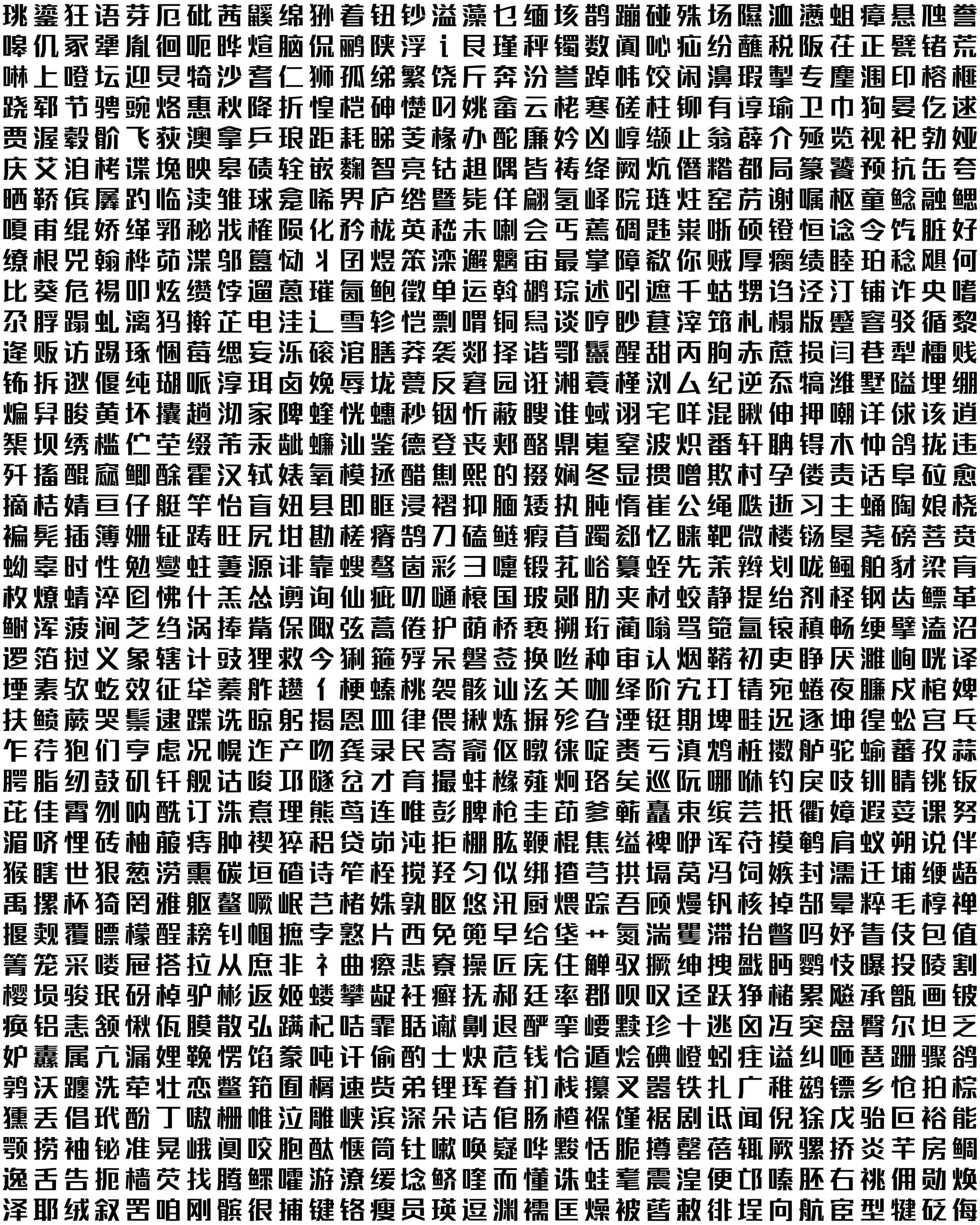}

    \caption{Generated glyphs from test fonts using GAR-Font($I_8$, +NFA-$8$+SE, Large dataset).}
    \label{fig: add_sheet_1}

\end{figure*}

\begin{figure*}
    \centering
    \includegraphics[width=0.95\linewidth]{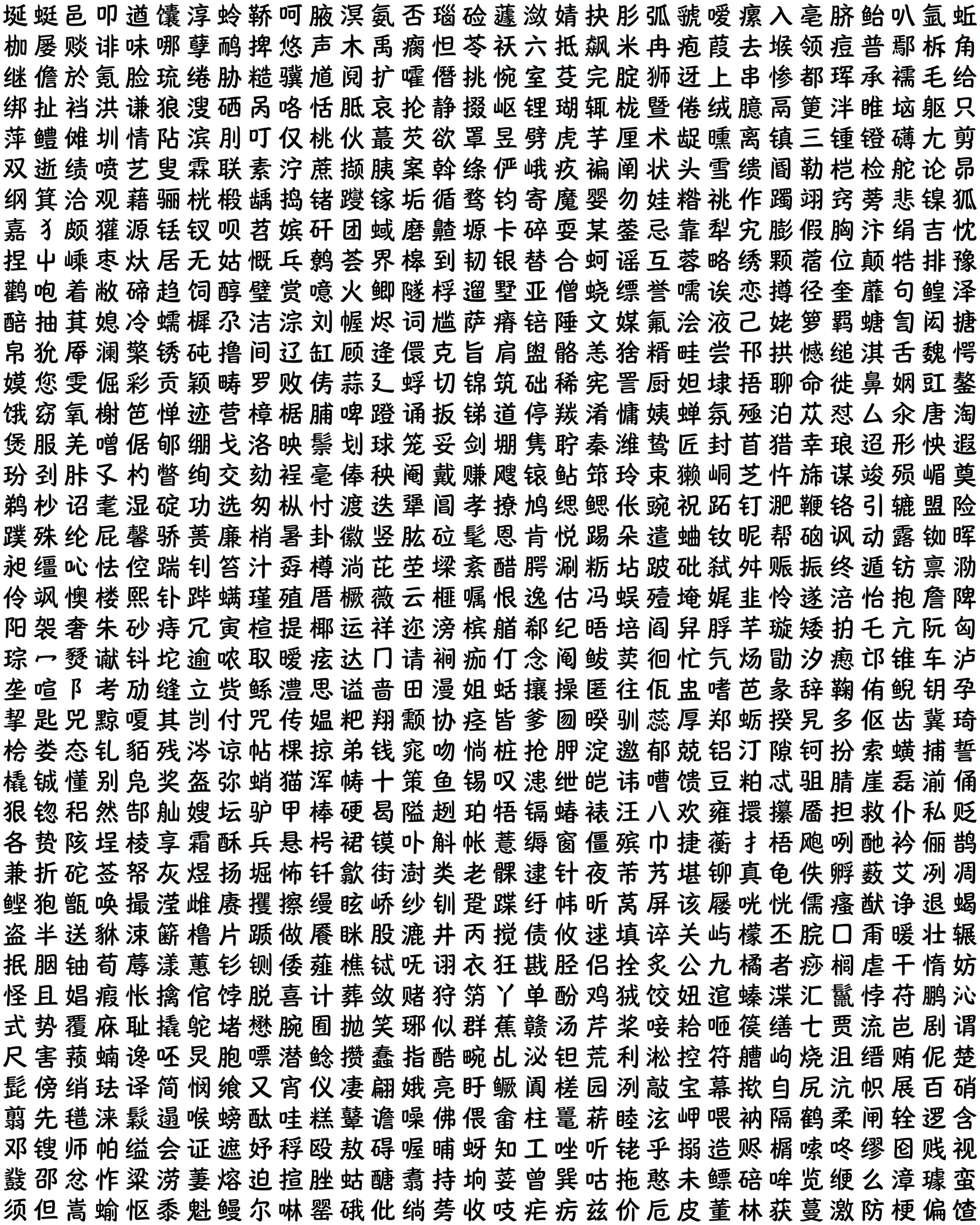}

    \caption{Generated glyphs from test fonts using GAR-Font($I_8$, +NFA-$8$+SE, Large dataset).}
    \label{fig: add_sheet_2}

\end{figure*}

\begin{figure*}
    \centering
    \includegraphics[width=0.95\linewidth]{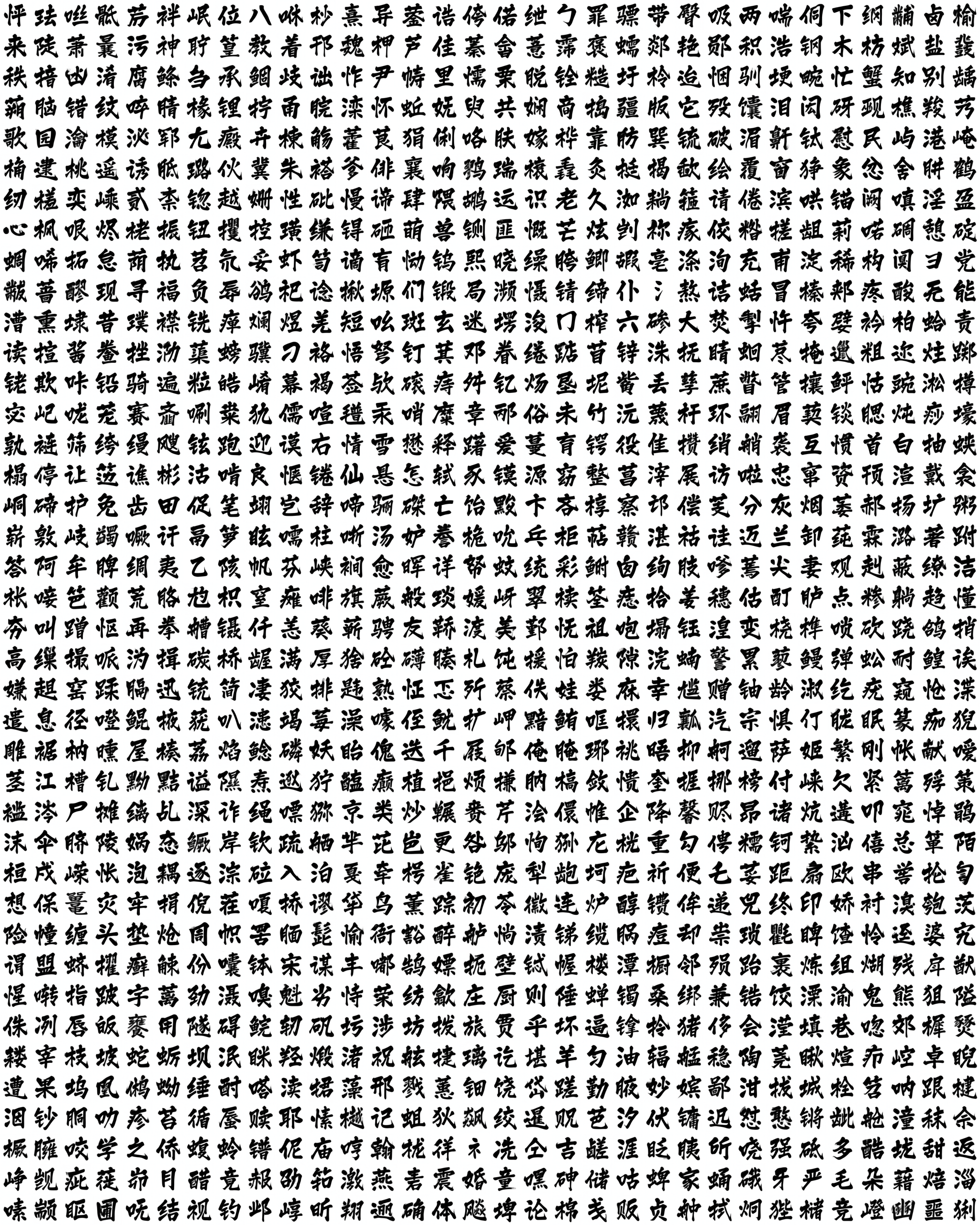}

    \caption{Generated glyphs from test fonts using GAR-Font($I_8$, +NFA-$8$+SE, Large dataset).}
    \label{fig: add_sheet_3}

\end{figure*}

\begin{figure*}
    \centering
    \includegraphics[width=0.95\linewidth]{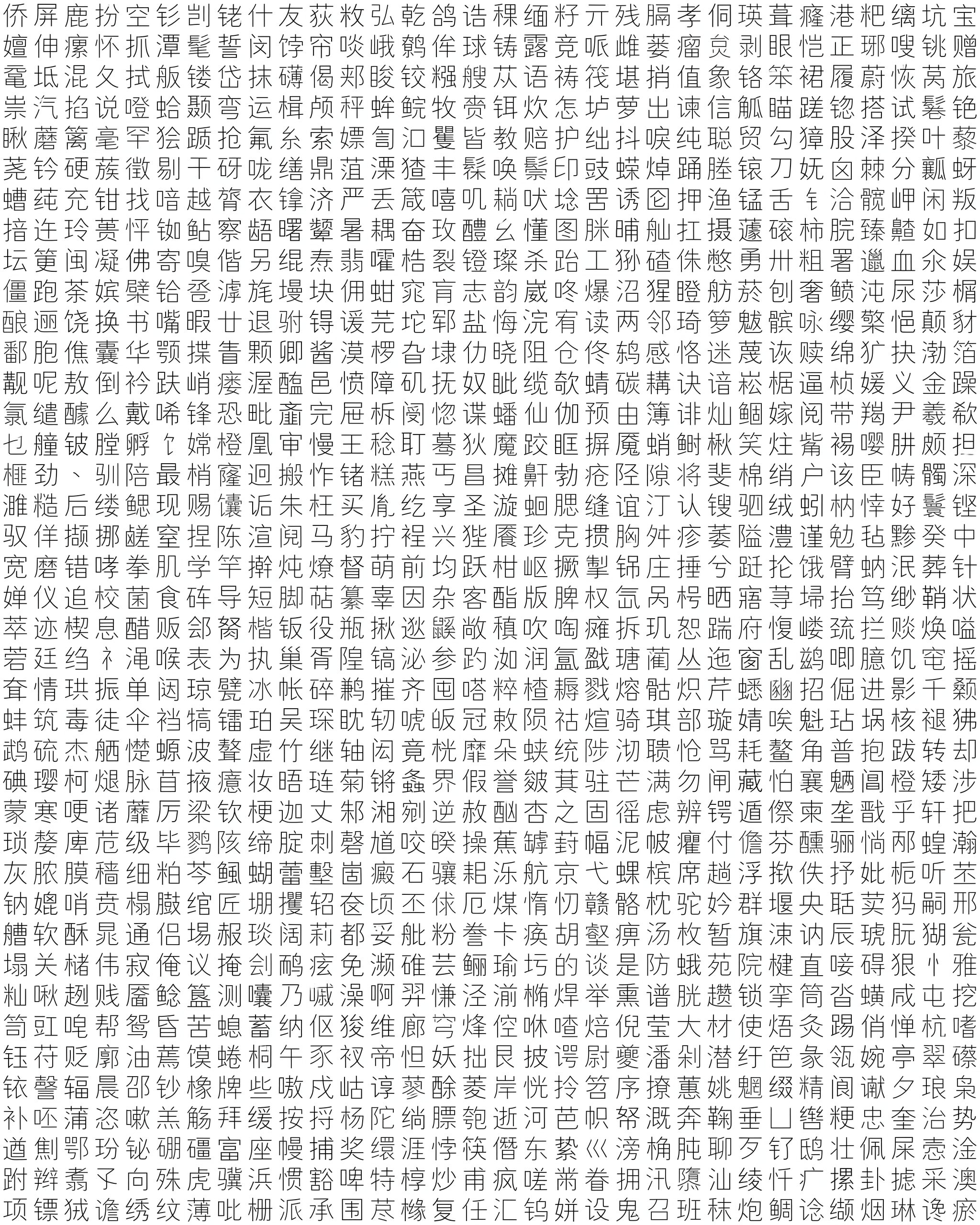}

    \caption{Generated glyphs from test fonts using GAR-Font($I_8$, +NFA-$8$+SE, Large dataset).}
    \label{fig: add_sheet_4}

\end{figure*}

\begin{figure*}
    \centering
    \includegraphics[width=0.95\linewidth]{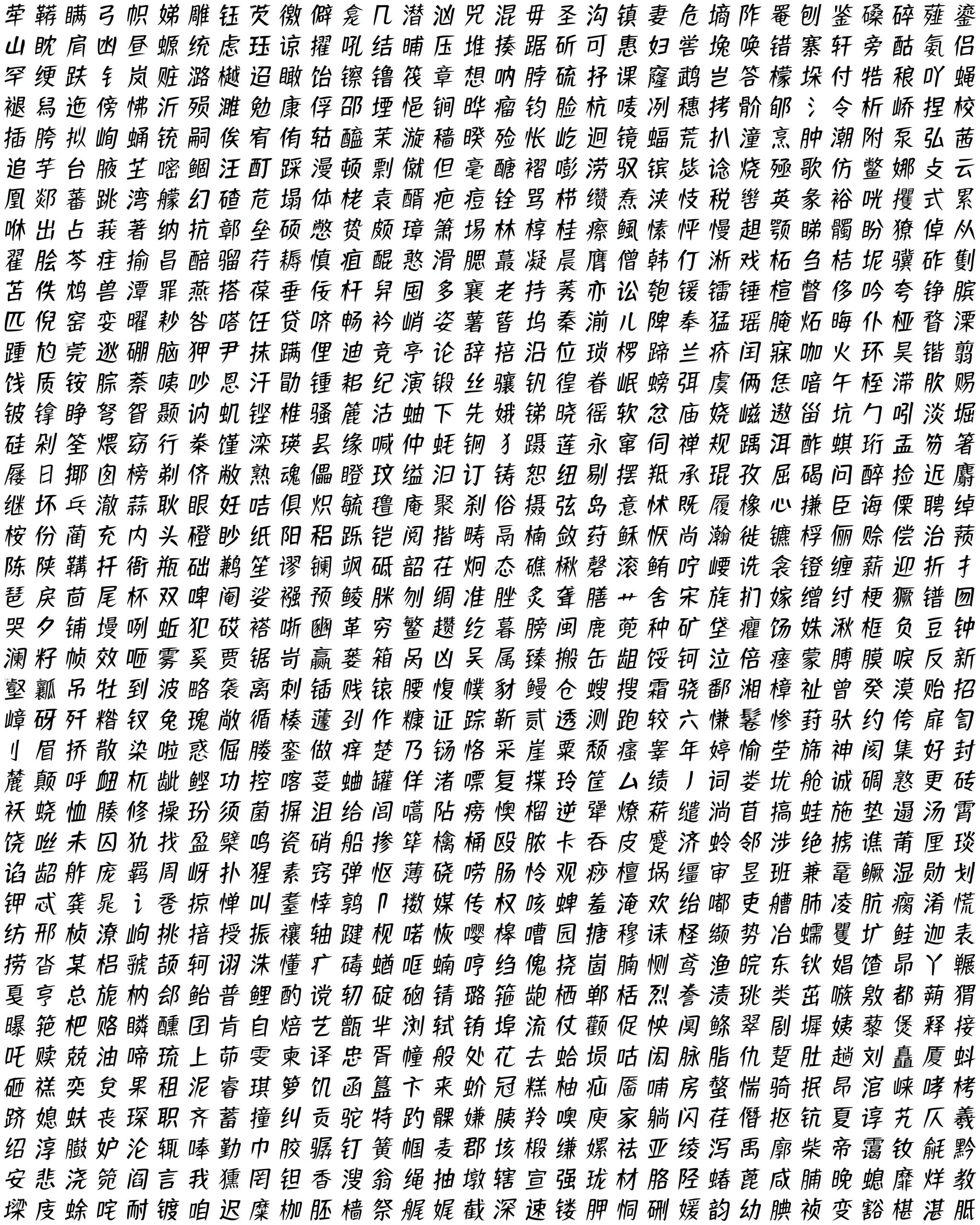}

    \caption{Generated glyphs from test fonts using GAR-Font($I_8$, +NFA-$8$+SE, Large dataset).}
    \label{fig: add_sheet_5}

\end{figure*}

To illustrate the capabilities of GAR-Font, we generate the full GB2312 character set for five test fonts with GAR-Font($I_8$, +NFA-$8$+SE, trained on Large dataset) and randomly select 1,280 samples per font. The generated glyphs are shown in \cref{fig: add_sheet_1}-\cref{fig: add_sheet_5} , demonstrating the model’s ability to produce large-scale character sets while faithfully preserving each font’s distinctive stylistic features.

\section{Failure Cases and Analysis}
\label{supplesec:sup_Failure_cases}

While GAR-Font generally performs well, distortions and blurring occasionally appear in dense-stroke regions of highly complex fonts (\cref{fig:supple_failure_cases}). To investigate this, we applied a content classifier to all \textit{UFUC} samples generated by GAR-Font($I_8$,+NFA-$128$+SE), using the softmax output as a measure of content confidence. The results reveal a clear trend: content confidence notably decreases as stylistic complexity increases (\cref{fig:supple_analysis}), suggesting the model sometimes sacrifices structural accuracy to better capture stylistic features.

We hypothesize that this structural degradation results from the error accumulation inherent in autoregressive modeling. Without explicit structural constraints, the model tends to drift when generating intricate stroke patterns. A promising direction for future work is to incorporate explicit structural priors, such as character skeletons or stroke sequences, to guide the generation process. This would help preserve structural fidelity in complex styles.

\clearpage
\clearpage
\clearpage
\clearpage
\clearpage
{
    \small
    \bibliographystyle{ieeenat_fullname}
    \bibliography{main}

\begin{thebibliography}{81}
\providecommand{\natexlab}[1]{#1}
\providecommand{\url}[1]{\texttt{#1}}
\expandafter\ifx\csname urlstyle\endcsname\relax
  \providecommand{\doi}[1]{doi: #1}\else
  \providecommand{\doi}{doi: \begingroup \urlstyle{rm}\Url}\fi

\bibitem[Alayrac et~al.(2022)Alayrac, Donahue, Luc, Miech, Barr, Hasson, Lenc, Mensch, Millican, Reynolds, et~al.]{Flamingo}
Jean-Baptiste Alayrac, Jeff Donahue, Pauline Luc, Antoine Miech, Iain Barr, Yana Hasson, Karel Lenc, Arthur Mensch, Katherine Millican, Malcolm Reynolds, et~al.
\newblock Flamingo: a visual language model for few-shot learning.
\newblock \emph{Advances in neural information processing systems}, 35:\penalty0 23716--23736, 2022.

\bibitem[Bachmann et~al.(2025)Bachmann, Allardice, Mizrahi, Fini, Kar, Amirloo, El-Nouby, Zamir, and Dehghan]{FlexTok}
Roman Bachmann, Jesse Allardice, David Mizrahi, Enrico Fini, O{\u{g}}uzhan~Fatih Kar, Elmira Amirloo, Alaaeldin El-Nouby, Amir Zamir, and Afshin Dehghan.
\newblock Flextok: Resampling images into 1d token sequences of flexible length.
\newblock In \emph{Forty-second International Conference on Machine Learning}, 2025.

\bibitem[Bai et~al.(2025)Bai, Chen, Liu, Wang, Ge, Song, Dang, Wang, Wang, Tang, et~al.]{Qwen25VL}
Shuai Bai, Keqin Chen, Xuejing Liu, Jialin Wang, Wenbin Ge, Sibo Song, Kai Dang, Peng Wang, Shijie Wang, Jun Tang, et~al.
\newblock Qwen2. 5-vl technical report.
\newblock \emph{arXiv preprint arXiv:2502.13923}, 2025.

\bibitem[Bhunia et~al.(2018)Bhunia, Bhunia, Banerjee, Konwer, Bhowmick, Roy, and Pal]{word2word}
Ankan~Kumar Bhunia, Ayan~Kumar Bhunia, Prithaj Banerjee, Aishik Konwer, Abir Bhowmick, Partha~Pratim Roy, and Umapada Pal.
\newblock Word level font-to-font image translation using convolutional recurrent generative adversarial networks.
\newblock In \emph{2018 24th International Conference on Pattern Recognition (ICPR)}, pages 3645--3650. IEEE, 2018.

\bibitem[Cao et~al.(2023)Cao, Yin, Huang, Liu, Zhao, Zhao, and Huang]{EfficientVQGAN}
Shiyue Cao, Yueqin Yin, Lianghua Huang, Yu Liu, Xin Zhao, Deli Zhao, and Kaigi Huang.
\newblock Efficient-vqgan: Towards high-resolution image generation with efficient vision transformers.
\newblock In \emph{Proceedings of the IEEE/CVF International Conference on Computer Vision}, pages 7368--7377, 2023.

\bibitem[Cao et~al.(2025)Cao, Chen, Chen, Cheng, Cui, Deng, Dong, Gong, Gu, Gu, et~al.]{Hunyuan-Image}
Siyu Cao, Hangting Chen, Peng Chen, Yiji Cheng, Yutao Cui, Xinchi Deng, Ying Dong, Kipper Gong, Tianpeng Gu, Xiusen Gu, et~al.
\newblock Hunyuanimage 3.0 technical report.
\newblock \emph{arXiv preprint arXiv:2509.23951}, 2025.

\bibitem[Cao et~al.(2024)Cao, Zhang, Frittoli, Cheng, Shen, and Boracchi]{AdaCLIP}
Yunkang Cao, Jiangning Zhang, Luca Frittoli, Yuqi Cheng, Weiming Shen, and Giacomo Boracchi.
\newblock Adaclip: Adapting clip with hybrid learnable prompts for zero-shot anomaly detection.
\newblock In \emph{European Conference on Computer Vision}, pages 55--72. Springer, 2024.

\bibitem[Cha et~al.(2020)Cha, Chun, Lee, Lee, Kim, and Lee]{fewshotFFG}
Junbum Cha, Sanghyuk Chun, Gayoung Lee, Bado Lee, Seonghyeon Kim, and Hwalsuk Lee.
\newblock Few-shot compositional font generation with dual memory.
\newblock In \emph{European conference on computer vision}, pages 735--751. Springer, 2020.

\bibitem[Chang et~al.(2018{\natexlab{a}})Chang, Zhang, Pan, and Meng]{zi2zi}
Bo Chang, Qiong Zhang, Shenyi Pan, and Lili Meng.
\newblock Generating handwritten chinese characters using cyclegan.
\newblock In \emph{2018 IEEE winter conference on applications of computer vision (WACV)}, pages 199--207. IEEE, 2018{\natexlab{a}}.

\bibitem[Chang et~al.(2022)Chang, Zhang, Jiang, Liu, and Freeman]{MaskGIT}
Huiwen Chang, Han Zhang, Lu Jiang, Ce Liu, and William~T Freeman.
\newblock Maskgit: Masked generative image transformer.
\newblock In \emph{Proceedings of the IEEE/CVF conference on computer vision and pattern recognition}, pages 11315--11325, 2022.

\bibitem[Chang et~al.(2018{\natexlab{b}})Chang, Gu, Zhang, Wang, and Innovation]{HGAN}
Jie Chang, Yujun Gu, Ya Zhang, YanFeng Wang, and CM Innovation.
\newblock Chinese handwriting imitation with hierarchical generative adversarial network.
\newblock In \emph{BMVC}, page 290, 2018{\natexlab{b}}.

\bibitem[Chen et~al.(2020)Chen, Radford, Child, Wu, Jun, Luan, and Sutskever]{ImageGPT}
Mark Chen, Alec Radford, Rewon Child, Jeffrey Wu, Heewoo Jun, David Luan, and Ilya Sutskever.
\newblock Generative pretraining from pixels.
\newblock In \emph{International conference on machine learning}, pages 1691--1703. PMLR, 2020.

\bibitem[Chen et~al.(2024)Chen, Ke, and Guo]{IF_Font}
Xinping Chen, Xiao Ke, and Wenzhong Guo.
\newblock If-font: Ideographic description sequence-following font generation.
\newblock In \emph{Advances in Neural Information Processing Systems}, pages 14177--14199. Curran Associates, Inc., 2024.

\bibitem[Chen et~al.(2025)Chen, Wu, Liu, Pan, Liu, Xie, Yu, and Ruan]{januspro}
Xiaokang Chen, Zhiyu Wu, Xingchao Liu, Zizheng Pan, Wen Liu, Zhenda Xie, Xingkai Yu, and Chong Ruan.
\newblock Janus-pro: Unified multimodal understanding and generation with data and model scaling.
\newblock \emph{arXiv preprint arXiv:2501.17811}, 2025.

\bibitem[Deng et~al.(2025)Deng, Zhu, Li, Gou, Li, Wang, Zhong, Yu, Nie, Song, et~al.]{BAGEL}
Chaorui Deng, Deyao Zhu, Kunchang Li, Chenhui Gou, Feng Li, Zeyu Wang, Shu Zhong, Weihao Yu, Xiaonan Nie, Ziang Song, et~al.
\newblock Emerging properties in unified multimodal pretraining.
\newblock \emph{arXiv preprint arXiv:2505.14683}, 2025.

\bibitem[Dosovitskiy et~al.(2020)Dosovitskiy, Beyer, Kolesnikov, Weissenborn, Zhai, Unterthiner, Dehghani, Minderer, Heigold, Gelly, et~al.]{ViT}
Alexey Dosovitskiy, Lucas Beyer, Alexander Kolesnikov, Dirk Weissenborn, Xiaohua Zhai, Thomas Unterthiner, Mostafa Dehghani, Matthias Minderer, Georg Heigold, Sylvain Gelly, et~al.
\newblock An image is worth 16x16 words: Transformers for image recognition at scale.
\newblock \emph{arXiv preprint arXiv:2010.11929}, 2020.

\bibitem[Esser et~al.(2021)Esser, Rombach, and Ommer]{VQ_GAN}
Patrick Esser, Robin Rombach, and Bjorn Ommer.
\newblock Taming transformers for high-resolution image synthesis.
\newblock In \emph{Proceedings of the IEEE/CVF conference on computer vision and pattern recognition}, pages 12873--12883, 2021.

\bibitem[Fu et~al.(2024)Fu, Yu, Liu, Wang, Wen, He, and Qiao]{GenerateExpert}
Bin Fu, Fanghua Yu, Anran Liu, Zixuan Wang, Jie Wen, Junjun He, and Yu Qiao.
\newblock Generate like experts: multi-stage font generation by incorporating font transfer process into diffusion models.
\newblock In \emph{Proceedings of the IEEE/CVF conference on computer vision and pattern recognition}, pages 6892--6901, 2024.

\bibitem[Gao et~al.(2019)Gao, Guo, Lian, Tang, and Xiao]{AGIS-Net}
Yue Gao, Yuan Guo, Zhouhui Lian, Yingmin Tang, and Jianguo Xiao.
\newblock Artistic glyph image synthesis via one-stage few-shot learning.
\newblock \emph{ACM Transactions on Graphics (ToG)}, 38\penalty0 (6):\penalty0 1--12, 2019.

\bibitem[Ge et~al.(2024)Ge, Zhao, Zhu, Ge, Yi, Song, Li, Ding, and Shan]{Seed-X}
Yuying Ge, Sijie Zhao, Jinguo Zhu, Yixiao Ge, Kun Yi, Lin Song, Chen Li, Xiaohan Ding, and Ying Shan.
\newblock Seed-x: Multimodal models with unified multi-granularity comprehension and generation.
\newblock \emph{arXiv preprint arXiv:2404.14396}, 2024.

\bibitem[He et~al.(2024{\natexlab{a}})He, Chen, Wang, Liu, Du, Tao, and Yu]{Diff_Font}
Haibin He, Xinyuan Chen, Chaoyue Wang, Juhua Liu, Bo Du, Dacheng Tao, and Qiao Yu.
\newblock Diff-font: Diffusion model for robust one-shot font generation.
\newblock \emph{International Journal of Computer Vision}, 132\penalty0 (11):\penalty0 5372--5386, 2024{\natexlab{a}}.

\bibitem[He et~al.(2024{\natexlab{b}})He, Zhu, Wang, and Gao]{DS-Font}
Xiao He, Mingrui Zhu, Nannan Wang, and Xinbo Gao.
\newblock Few-shot font generation by learning style difference and similarity.
\newblock \emph{IEEE Transactions on Circuits and Systems for Video Technology}, 34\penalty0 (9):\penalty0 8013--8025, 2024{\natexlab{b}}.

\bibitem[Hu et~al.(2022)Hu, Shen, Wallis, Allen-Zhu, Li, Wang, Wang, Chen, et~al.]{LoRA}
Edward~J Hu, Yelong Shen, Phillip Wallis, Zeyuan Allen-Zhu, Yuanzhi Li, Shean Wang, Lu Wang, Weizhu Chen, et~al.
\newblock Lora: Low-rank adaptation of large language models.
\newblock \emph{ICLR}, 1\penalty0 (2):\penalty0 3, 2022.

\bibitem[Huang et~al.(2023)Huang, Mao, Wang, and Zhang]{MVTQ}
Mengqi Huang, Zhendong Mao, Quan Wang, and Yongdong Zhang.
\newblock Not all image regions matter: Masked vector quantization for autoregressive image generation.
\newblock In \emph{Proceedings of the IEEE/CVF Conference on Computer Vision and Pattern Recognition}, pages 2002--2011, 2023.

\bibitem[Huang et~al.(2025{\natexlab{a}})Huang, Chen, Zheng, Duan, Zhou, and Lu]{SpectralAR}
Yuanhui Huang, Weiliang Chen, Wenzhao Zheng, Yueqi Duan, Jie Zhou, and Jiwen Lu.
\newblock Spectralar: Spectral autoregressive visual generation.
\newblock In \emph{Proceedings of the IEEE/CVF International Conference on Computer Vision}, pages 15842--15852, 2025{\natexlab{a}}.

\bibitem[Huang et~al.(2025{\natexlab{b}})Huang, Qiu, Ma, Zhou, Chen, Zhang, Zhang, and Li]{NFIG}
Zhihao Huang, Xi Qiu, Yukuo Ma, Yifu Zhou, Junjie Chen, Hongyuan Zhang, Chi Zhang, and Xuelong Li.
\newblock Nfig: Autoregressive image generation with next-frequency prediction.
\newblock \emph{arXiv preprint arXiv:2503.07076}, 2025{\natexlab{b}}.

\bibitem[Jiang et~al.(2019)Jiang, Lian, Tang, and Xiao]{SCFont}
Yue Jiang, Zhouhui Lian, Yingmin Tang, and Jianguo Xiao.
\newblock Scfont: Structure-guided chinese font generation via deep stacked networks.
\newblock In \emph{Proceedings of the AAAI conference on artificial intelligence}, pages 4015--4022, 2019.

\bibitem[Kim et~al.(2024)Kim, Jeong, and Sim]{LegacyLearning}
Younghwi Kim, Seok~Chan Jeong, and Sunghyun Sim.
\newblock Legacy learning using few-shot font generation models for automatic text design in metaverse content: Cases studies in korean and chinese.
\newblock \emph{arXiv preprint arXiv:2408.16900}, 2024.

\bibitem[Kong et~al.(2022)Kong, Luo, Ma, Zhu, Zhu, Yuan, and Jin]{CG_GAN}
Yuxin Kong, Canjie Luo, Weihong Ma, Qiyuan Zhu, Shenggao Zhu, Nicholas Yuan, and Lianwen Jin.
\newblock Look closer to supervise better: One-shot font generation via component-based discriminator.
\newblock In \emph{Proceedings of the IEEE/CVF conference on computer vision and pattern recognition}, pages 13482--13491, 2022.

\bibitem[Koo et~al.(2025)Koo, Kim, Kwak, Nam, Kim, and Shin]{FontAdapter}
Myungkyu Koo, Subin Kim, Sangkyung Kwak, Jaehyun Nam, Seojin Kim, and Jinwoo Shin.
\newblock Fontadapter: Instant font adaptation in visual text generation.
\newblock \emph{arXiv preprint arXiv:2506.05843}, 2025.

\bibitem[Lee et~al.(2022)Lee, Kim, Kim, Cho, and Han]{RQ_VAE_RQ_FORMER}
Doyup Lee, Chiheon Kim, Saehoon Kim, Minsu Cho, and Wook-Shin Han.
\newblock Autoregressive image generation using residual quantization.
\newblock In \emph{Proceedings of the IEEE/CVF conference on computer vision and pattern recognition}, pages 11523--11532, 2022.

\bibitem[Li et~al.(2023)Li, Wong, Zhang, Usuyama, Liu, Yang, Naumann, Poon, and Gao]{llavamed}
Chunyuan Li, Cliff Wong, Sheng Zhang, Naoto Usuyama, Haotian Liu, Jianwei Yang, Tristan Naumann, Hoifung Poon, and Jianfeng Gao.
\newblock Llava-med: Training a large language-and-vision assistant for biomedicine in one day.
\newblock \emph{Advances in Neural Information Processing Systems}, 36:\penalty0 28541--28564, 2023.

\bibitem[Li and Lian(2024)]{HFH_Font}
Hua Li and Zhouhui Lian.
\newblock Hfh-font: few-shot chinese font synthesis with higher quality, faster speed, and higher resolution.
\newblock \emph{ACM Transactions on Graphics (TOG)}, 43\penalty0 (6):\penalty0 1--16, 2024.

\bibitem[Li and Zhu(2025)]{FSTDIFF}
Shilin Li and Anna Zhu.
\newblock Fstdiff: One-shot font generation via cross-font style transformation learning.
\newblock In \emph{International Conference on Document Analysis and Recognition}, pages 167--182. Springer, 2025.

\bibitem[Li et~al.(2024)Li, Qiu, Chen, Kuen, Gu, Raj, and Lin]{ImageFolder}
Xiang Li, Kai Qiu, Hao Chen, Jason Kuen, Jiuxiang Gu, Bhiksha Raj, and Zhe Lin.
\newblock Imagefolder: Autoregressive image generation with folded tokens.
\newblock \emph{arXiv preprint arXiv:2410.01756}, 2024.

\bibitem[Lian and Gao(2022)]{CVFont}
Zhouhui Lian and Yichen Gao.
\newblock Cvfont: Synthesizing chinese vector fonts via deep layout inferring.
\newblock In \emph{Computer Graphics Forum}, pages 212--225. Wiley Online Library, 2022.

\bibitem[Lian et~al.(2018)Lian, Zhao, Chen, and Xiao]{EasyFont}
Zhouhui Lian, Bo Zhao, Xudong Chen, and Jianguo Xiao.
\newblock Easyfont: a style learning-based system to easily build your large-scale handwriting fonts.
\newblock \emph{ACM Transactions on Graphics (TOG)}, 38\penalty0 (1):\penalty0 1--18, 2018.

\bibitem[Lin et~al.(2024)Lin, Ye, Zhu, Cui, Ning, Jin, and Yuan]{Videollava}
Bin Lin, Yang Ye, Bin Zhu, Jiaxi Cui, Munan Ning, Peng Jin, and Li Yuan.
\newblock Video-llava: Learning united visual representation by alignment before projection.
\newblock In \emph{Proceedings of the 2024 conference on empirical methods in natural language processing}, pages 5971--5984, 2024.

\bibitem[Liu et~al.(2022)Liu, Liu, Ding, He, and Yi]{Xmpfont}
Wei Liu, Fangyue Liu, Fei Ding, Qian He, and Zili Yi.
\newblock Xmp-font: Self-supervised cross-modality pre-training for few-shot font generation.
\newblock In \emph{Proceedings of the IEEE/CVF conference on computer vision and pattern recognition}, pages 7905--7914, 2022.

\bibitem[Liu et~al.(2024)Liu, Zhang, Luo, Huang, and Xu]{TextAdapter}
Xiao-Qian Liu, Peng-Fei Zhang, Xin Luo, Zi Huang, and Xin-Shun Xu.
\newblock Textadapter: Self-supervised domain adaptation for cross-domain text recognition.
\newblock \emph{IEEE Transactions on Multimedia}, 26:\penalty0 9854--9865, 2024.

\bibitem[Liu et~al.(2023)Liu, Zhang, Guo, Fisher, Wang, and Zhang]{DualVector}
Ying-Tian Liu, Zhifei Zhang, Yuan-Chen Guo, Matthew Fisher, Zhaowen Wang, and Song-Hai Zhang.
\newblock Dualvector: Unsupervised vector font synthesis with dual-part representation.
\newblock In \emph{Proceedings of the IEEE/CVF Conference on Computer Vision and Pattern Recognition}, pages 14193--14202, 2023.

\bibitem[Lu et~al.(2024)Lu, Clark, Lee, Zhang, Khosla, Marten, Hoiem, and Kembhavi]{Unifiedio2}
Jiasen Lu, Christopher Clark, Sangho Lee, Zichen Zhang, Savya Khosla, Ryan Marten, Derek Hoiem, and Aniruddha Kembhavi.
\newblock Unified-io 2: Scaling autoregressive multimodal models with vision language audio and action.
\newblock In \emph{Proceedings of the IEEE/CVF Conference on Computer Vision and Pattern Recognition}, pages 26439--26455, 2024.

\bibitem[Luo et~al.(2025)Luo, Tang, Huang, Hao, and Lian]{CalliReader}
Yuxuan Luo, Jiaqi Tang, Chenyi Huang, Feiyang Hao, and Zhouhui Lian.
\newblock Callireader: contextualizing chinese calligraphy via an embedding-aligned vision-language model.
\newblock In \emph{Proceedings of the IEEE/CVF International Conference on Computer Vision}, pages 23030--23040, 2025.

\bibitem[Ma et~al.(2024)Ma, Zhou, Liang, Bai, Zhao, Li, Chen, and Jin]{star_scale_wise}
Xiaoxiao Ma, Mohan Zhou, Tao Liang, Yalong Bai, Tiejun Zhao, Biye Li, Huaian Chen, and Yi Jin.
\newblock Star: Scale-wise text-conditioned autoregressive image generation.
\newblock \emph{arXiv preprint arXiv:2406.10797}, 2024.

\bibitem[Marafioti et~al.(2025)Marafioti, Zohar, Farr{\'e}, Noyan, Bakouch, Cuenca, Zakka, Allal, Lozhkov, Tazi, et~al.]{smolvlm}
Andr{\'e}s Marafioti, Orr Zohar, Miquel Farr{\'e}, Merve Noyan, Elie Bakouch, Pedro Cuenca, Cyril Zakka, Loubna~Ben Allal, Anton Lozhkov, Nouamane Tazi, et~al.
\newblock Smolvlm: Redefining small and efficient multimodal models.
\newblock \emph{arXiv preprint arXiv:2504.05299}, 2025.

\bibitem[Pan et~al.(2023)Pan, Zhu, Zhou, Iwana, and Li]{FineGrainedFFG}
Wei Pan, Anna Zhu, Xinyu Zhou, Brian~Kenji Iwana, and Shilin Li.
\newblock Few shot font generation via transferring similarity guided global style and quantization local style.
\newblock In \emph{Proceedings of the IEEE/CVF International Conference on Computer Vision}, pages 19506--19516, 2023.

\bibitem[Park et~al.(2021{\natexlab{a}})Park, Chun, Cha, Lee, and Shim]{LF_Font}
Song Park, Sanghyuk Chun, Junbum Cha, Bado Lee, and Hyunjung Shim.
\newblock Few-shot font generation with localized style representations and factorization.
\newblock In \emph{Proceedings of the AAAI conference on artificial intelligence}, pages 2393--2402, 2021{\natexlab{a}}.

\bibitem[Park et~al.(2021{\natexlab{b}})Park, Chun, Cha, Lee, and Shim]{MX_Font}
Song Park, Sanghyuk Chun, Junbum Cha, Bado Lee, and Hyunjung Shim.
\newblock Multiple heads are better than one: Few-shot font generation with multiple localized experts.
\newblock In \emph{Proceedings of the IEEE/CVF international conference on computer vision}, pages 13900--13909, 2021{\natexlab{b}}.

\bibitem[Parmar et~al.(2018)Parmar, Vaswani, Uszkoreit, Kaiser, Shazeer, Ku, and Tran]{ImageTransformer}
Niki Parmar, Ashish Vaswani, Jakob Uszkoreit, Lukasz Kaiser, Noam Shazeer, Alexander Ku, and Dustin Tran.
\newblock Image transformer.
\newblock In \emph{International conference on machine learning}, pages 4055--4064. PMLR, 2018.

\bibitem[Razavi et~al.(2019)Razavi, Van~den Oord, and Vinyals]{VQ_VAE2}
Ali Razavi, Aaron Van~den Oord, and Oriol Vinyals.
\newblock Generating diverse high-fidelity images with vq-vae-2.
\newblock \emph{Advances in neural information processing systems}, 32, 2019.

\bibitem[Shao et~al.(2024)Shao, Wang, Zhu, Xu, Song, Bi, Zhang, Zhang, Li, Wu, et~al.]{DeepseekMath_GRPO}
Zhihong Shao, Peiyi Wang, Qihao Zhu, Runxin Xu, Junxiao Song, Xiao Bi, Haowei Zhang, Mingchuan Zhang, YK Li, Yang Wu, et~al.
\newblock Deepseekmath: Pushing the limits of mathematical reasoning in open language models.
\newblock \emph{arXiv preprint arXiv:2402.03300}, 2024.

\bibitem[Shi et~al.(2025)Shi, Song, Zhang, Liu, and Zou]{FonTS}
Wenda Shi, Yiren Song, Dengming Zhang, Jiaming Liu, and Xingxing Zou.
\newblock Fonts: Text rendering with typography and style controls.
\newblock In \emph{Proceedings of the IEEE/CVF International Conference on Computer Vision}, pages 18463--18474, 2025.

\bibitem[Sun et~al.(2024)Sun, Jiang, Chen, Zhang, Peng, Luo, and Yuan]{llamagen}
Peize Sun, Yi Jiang, Shoufa Chen, Shilong Zhang, Bingyue Peng, Ping Luo, and Zehuan Yuan.
\newblock Autoregressive model beats diffusion: Llama for scalable image generation.
\newblock \emph{arXiv preprint arXiv:2406.06525}, 2024.

\bibitem[Sun et~al.(2023)Sun, Yu, Cui, Zhang, Zhang, Wang, Gao, Liu, Huang, and Wang]{Emu}
Quan Sun, Qiying Yu, Yufeng Cui, Fan Zhang, Xiaosong Zhang, Yueze Wang, Hongcheng Gao, Jingjing Liu, Tiejun Huang, and Xinlong Wang.
\newblock Emu: Generative pretraining in multimodality.
\newblock \emph{arXiv preprint arXiv:2307.05222}, 2023.

\bibitem[Tang et~al.(2022)Tang, Cai, Liu, Hong, Gong, Fan, Han, Liu, Ding, and Wang]{Fs_Font}
Licheng Tang, Yiyang Cai, Jiaming Liu, Zhibin Hong, Mingming Gong, Minhu Fan, Junyu Han, Jingtuo Liu, Errui Ding, and Jingdong Wang.
\newblock Few-shot font generation by learning fine-grained local styles.
\newblock In \emph{Proceedings of the IEEE/CVF conference on computer vision and pattern recognition}, pages 7895--7904, 2022.

\bibitem[Tang et~al.(2019)Tang, Xia, Lian, Tang, and Xiao]{FontRNN}
Shusen Tang, Zeqing Xia, Zhouhui Lian, Yingmin Tang, and Jianguo Xiao.
\newblock Fontrnn: Generating large-scale chinese fonts via recurrent neural network.
\newblock In \emph{Computer Graphics Forum}, pages 567--577. Wiley Online Library, 2019.

\bibitem[Thamizharasan et~al.(2024)Thamizharasan, Liu, Agarwal, Fisher, Gharbi, Wang, Jacobson, and Kalogerakis]{Vecfusion}
Vikas Thamizharasan, Difan Liu, Shantanu Agarwal, Matthew Fisher, Micha{\"e}l Gharbi, Oliver Wang, Alec Jacobson, and Evangelos Kalogerakis.
\newblock Vecfusion: Vector font generation with diffusion.
\newblock In \emph{Proceedings of the IEEE/CVF conference on computer vision and pattern recognition}, pages 7943--7952, 2024.

\bibitem[Tian et~al.(2024)Tian, Jiang, Yuan, Peng, and Wang]{VAR}
Keyu Tian, Yi Jiang, Zehuan Yuan, Bingyue Peng, and Liwei Wang.
\newblock Visual autoregressive modeling: Scalable image generation via next-scale prediction.
\newblock \emph{Advances in neural information processing systems}, 37:\penalty0 84839--84865, 2024.

\bibitem[Van Den~Oord et~al.(2017)Van Den~Oord, Vinyals, et~al.]{VQ_VAE}
Aaron Van Den~Oord, Oriol Vinyals, et~al.
\newblock Neural discrete representation learning.
\newblock \emph{Advances in neural information processing systems}, 30, 2017.

\bibitem[Wang et~al.(2023{\natexlab{a}})Wang, Zhou, Ge, Jiang, Bao, and Xu]{CF_Font}
Chi Wang, Min Zhou, Tiezheng Ge, Yuning Jiang, Hujun Bao, and Weiwei Xu.
\newblock Cf-font: Content fusion for few-shot font generation.
\newblock In \emph{Proceedings of the IEEE/CVF conference on computer vision and pattern recognition}, pages 1858--1867, 2023{\natexlab{a}}.

\bibitem[Wang et~al.(2025{\natexlab{a}})Wang, Tian, Wang, Zhang, Huang, Wu, and Jiang]{simpleAR}
Junke Wang, Zhi Tian, Xun Wang, Xinyu Zhang, Weilin Huang, Zuxuan Wu, and Yu-Gang Jiang.
\newblock Simplear: Pushing the frontier of autoregressive visual generation through pretraining, sft, and rl.
\newblock \emph{arXiv preprint arXiv:2504.11455}, 2025{\natexlab{a}}.

\bibitem[Wang et~al.(2024)Wang, Zhang, Luo, Sun, Cui, Wang, Zhang, Wang, Li, Yu, et~al.]{Emu3}
Xinlong Wang, Xiaosong Zhang, Zhengxiong Luo, Quan Sun, Yufeng Cui, Jinsheng Wang, Fan Zhang, Yueze Wang, Zhen Li, Qiying Yu, et~al.
\newblock Emu3: Next-token prediction is all you need.
\newblock \emph{arXiv preprint arXiv:2409.18869}, 2024.

\bibitem[Wang and Lian(2021)]{Deepvecfont}
Yizhi Wang and Zhouhui Lian.
\newblock Deepvecfont: synthesizing high-quality vector fonts via dual-modality learning.
\newblock \emph{ACM Transactions on Graphics (TOG)}, 40\penalty0 (6):\penalty0 1--15, 2021.

\bibitem[Wang et~al.(2023{\natexlab{b}})Wang, Wang, Yu, Zhu, and Lian]{DeepVecFontV2}
Yuqing Wang, Yizhi Wang, Longhui Yu, Yuesheng Zhu, and Zhouhui Lian.
\newblock Deepvecfont-v2: Exploiting transformers to synthesize vector fonts with higher quality.
\newblock In \emph{Proceedings of the IEEE/CVF conference on computer vision and pattern recognition}, pages 18320--18328, 2023{\natexlab{b}}.

\bibitem[Wang et~al.(2025{\natexlab{b}})Wang, Ren, Lin, Han, Guo, Yang, Zou, Feng, and Liu]{ARPG_LocalTokens}
Yuqing Wang, Shuhuai Ren, Zhijie Lin, Yujin Han, Haoyuan Guo, Zhenheng Yang, Difan Zou, Jiashi Feng, and Xihui Liu.
\newblock Parallelized autoregressive visual generation.
\newblock In \emph{Proceedings of the IEEE/CVF Conference on Computer Vision and Pattern Recognition}, pages 12955--12965, 2025{\natexlab{b}}.

\bibitem[Wang et~al.(2025{\natexlab{c}})Wang, Chen, Hu, Liu, Sun, Wu, Su, Yu, Barsoum, and Liu]{InstellaT2I}
Ze Wang, Hao Chen, Benran Hu, Jiang Liu, Ximeng Sun, Jialian Wu, Yusheng Su, Xiaodong Yu, Emad Barsoum, and Zicheng Liu.
\newblock Instella-t2i: Pushing the limits of 1d discrete latent space image generation.
\newblock \emph{arXiv preprint arXiv:2506.21022}, 2025{\natexlab{c}}.

\bibitem[Wen et~al.(2021)Wen, Li, Han, and Yuan]{zigan}
Qi Wen, Shuang Li, Bingfeng Han, and Yi Yuan.
\newblock Zigan: Fine-grained chinese calligraphy font generation via a few-shot style transfer approach.
\newblock In \emph{Proceedings of the 29th ACM international conference on multimedia}, pages 621--629, 2021.

\bibitem[Wu et~al.(2025)Wu, Li, Zhou, Lin, Gao, Yan, Yin, Bai, Xu, Chen, et~al.]{Qwen-Image}
Chenfei Wu, Jiahao Li, Jingren Zhou, Junyang Lin, Kaiyuan Gao, Kun Yan, Sheng-ming Yin, Shuai Bai, Xiao Xu, Yilei Chen, et~al.
\newblock Qwen-image technical report.
\newblock \emph{arXiv preprint arXiv:2508.02324}, 2025.

\bibitem[Xia et~al.(2023)Xia, Xiong, and Lian]{Vecfontsdf}
Zeqing Xia, Bojun Xiong, and Zhouhui Lian.
\newblock Vecfontsdf: Learning to reconstruct and synthesize high-quality vector fonts via signed distance functions.
\newblock In \emph{Proceedings of the IEEE/CVF conference on computer vision and pattern recognition}, pages 1848--1857, 2023.

\bibitem[Xiao et~al.(2024)Xiao, Xu, Yuille, Yan, and Wang]{PaLM2-VAdapter}
Junfei Xiao, Zheng Xu, Alan Yuille, Shen Yan, and Boyu Wang.
\newblock Palm2-vadapter: Progressively aligned language model makes a strong vision-language adapter.
\newblock \emph{arXiv preprint arXiv:2402.10896}, 2024.

\bibitem[Xie et~al.(2021)Xie, Chen, Sun, and Lu]{DG_Font}
Yangchen Xie, Xinyuan Chen, Li Sun, and Yue Lu.
\newblock Dg-font: Deformable generative networks for unsupervised font generation.
\newblock In \emph{Proceedings of the IEEE/CVF conference on computer vision and pattern recognition}, pages 5130--5140, 2021.

\bibitem[Yang et~al.(2024)Yang, Peng, Kong, Zhang, Yao, and Jin]{Font_Diffuser}
Zhenhua Yang, Dezhi Peng, Yuxin Kong, Yuyi Zhang, Cong Yao, and Lianwen Jin.
\newblock Fontdiffuser: One-shot font generation via denoising diffusion with multi-scale content aggregation and style contrastive learning.
\newblock In \emph{Proceedings of the AAAI conference on artificial intelligence}, pages 6603--6611, 2024.

\bibitem[Yao et~al.(2024)Yao, Zhang, Lin, Li, and Zuo]{VQ_Font}
Mingshuai Yao, Yabo Zhang, Xianhui Lin, Xiaoming Li, and Wangmeng Zuo.
\newblock Vq-font: Few-shot font generation with structure-aware enhancement and quantization.
\newblock In \emph{Proceedings of the AAAI Conference on Artificial Intelligence}, pages 16407--16415, 2024.

\bibitem[Yu et~al.(2021)Yu, Li, Koh, Zhang, Pang, Qin, Ku, Xu, Baldridge, and Wu]{VIT_VQGAN}
Jiahui Yu, Xin Li, Jing~Yu Koh, Han Zhang, Ruoming Pang, James Qin, Alexander Ku, Yuanzhong Xu, Jason Baldridge, and Yonghui Wu.
\newblock Vector-quantized image modeling with improved vqgan.
\newblock \emph{arXiv preprint arXiv:2110.04627}, 2021.

\bibitem[Yu et~al.(2023)Yu, Shi, Pasunuru, Muller, Golovneva, Wang, Babu, Tang, Karrer, Sheynin, et~al.]{Chameleon}
Lili Yu, Bowen Shi, Ramakanth Pasunuru, Benjamin Muller, Olga Golovneva, Tianlu Wang, Arun Babu, Binh Tang, Brian Karrer, Shelly Sheynin, et~al.
\newblock Scaling autoregressive multi-modal models: Pretraining and instruction tuning.
\newblock \emph{arXiv preprint arXiv:2309.02591}, 2023.

\bibitem[Yu et~al.(2024)Yu, Weber, Deng, Shen, Cremers, and Chen]{Titok}
Qihang Yu, Mark Weber, Xueqing Deng, Xiaohui Shen, Daniel Cremers, and Liang-Chieh Chen.
\newblock An image is worth 32 tokens for reconstruction and generation.
\newblock \emph{Advances in Neural Information Processing Systems}, 37:\penalty0 128940--128966, 2024.

\bibitem[Yu et~al.(2025)Yu, He, Deng, Shen, and Chen]{RAR}
Qihang Yu, Ju He, Xueqing Deng, Xiaohui Shen, and Liang-Chieh Chen.
\newblock Randomized autoregressive visual generation.
\newblock In \emph{Proceedings of the IEEE/CVF International Conference on Computer Vision}, pages 18431--18441, 2025.

\bibitem[Zha et~al.(2025)Zha, Yu, Fathi, Ross, Schmid, Katabi, and Gu]{TexTok}
Kaiwen Zha, Lijun Yu, Alireza Fathi, David~A Ross, Cordelia Schmid, Dina Katabi, and Xiuye Gu.
\newblock Language-guided image tokenization for generation.
\newblock In \emph{Proceedings of the Computer Vision and Pattern Recognition Conference}, pages 15713--15722, 2025.

\bibitem[Zhang et~al.(2018{\natexlab{a}})Zhang, Zhang, and Cai]{EMD}
Yexun Zhang, Ya Zhang, and Wenbin Cai.
\newblock Separating style and content for generalized style transfer.
\newblock In \emph{Proceedings of the IEEE conference on computer vision and pattern recognition}, pages 8447--8455, 2018{\natexlab{a}}.

\bibitem[Zhang et~al.(2018{\natexlab{b}})Zhang, Zhang, and Cai]{zhang2018separating}
Yexun Zhang, Ya Zhang, and Wenbin Cai.
\newblock Separating style and content for generalized style transfer.
\newblock In \emph{Proceedings of the IEEE conference on computer vision and pattern recognition}, pages 8447--8455, 2018{\natexlab{b}}.

\bibitem[Zheng et~al.(2025)Zheng, Wang, Zhao, Deng, Wang, Zhang, and Qi]{Hita}
Anlin Zheng, Haochen Wang, Yucheng Zhao, Weipeng Deng, Tiancai Wang, Xiangyu Zhang, and Xiaojuan Qi.
\newblock Holistic tokenizer for autoregressive image generation.
\newblock In \emph{Proceedings of the IEEE/CVF International Conference on Computer Vision}, pages 16916--16926, 2025.

\end{thebibliography}
}


\end{document}